\documentclass{article} 
\usepackage{iclr2027_conference,times} 
\usepackage{amsmath,amsfonts,bm}

\def\eqref#1{equation~\ref{#1}}

\def\1{\bm{1}}

\DeclareMathAlphabet{\mathsfit}{\encodingdefault}{\sfdefault}{m}{sl}
\SetMathAlphabet{\mathsfit}{bold}{\encodingdefault}{\sfdefault}{bx}{n}

\usepackage{hyperref}
\usepackage{url}
\usepackage{graphicx}
\usepackage{amsmath,amssymb,amsthm,mathtools,bm}
\usepackage{algorithm}
\usepackage{algorithmic}
\usepackage{booktabs}

\usepackage{tcolorbox}
\usepackage{enumitem}

\newtheorem{assumption}{Assumption}

\newtheorem{remark}{Remark}
\newtheorem{theorem}{Theorem}

\newtheorem{lemma}{Lemma}
\newtheorem{corollary}{Corollary}
\newtheorem{proposition}{Proposition}

\usepackage{booktabs}

\usepackage{amsmath,amssymb}
\usepackage{fix-cm}
\usepackage{tikz}
\usetikzlibrary{arrows.meta,calc,decorations.pathreplacing}

\providecommand{\lcpPilotRef}{Sec.~\ref{sec:pilot_policy_and_nonconformity_score}}
\providecommand{\lcpCalibrationRef}{Sec.~\ref{sec:conformal_calibration}}
\providecommand{\lcpPredictionRef}{Eq.~(\ref{eq:conformal_prediction_set_for_observable_preference_outcome})}
\providecommand{\lcpPessimisticRef}{Eq.~(\ref{eq:unit_level_pessimistic_dpo_objective})}
\providecommand{\lcpOptimizationRef}{Sec.~\ref{sec:from_latent_cp_to_pessimistic_dpo}}
\providecommand{\lcpCoverageRef}{Theorem~\ref{thm:latent_coverage}}
\providecommand{\lcpLatentCPSectionRef}{Sec.~\ref{sec:conformal_calibration} and \ref{sec:uncertainty_sets_for_latent_relative_rewards}}
\providecommand{\lcpPessimisticSectionRef}{Sec.~\ref{sec:from_latent_cp_to_pessimistic_dpo}}

\definecolor{lcpBlue}{HTML}{2586BA}
\definecolor{lcpBlueLight}{HTML}{E8F5FC}
\definecolor{lcpPilot}{HTML}{DDF2FA}
\definecolor{lcpPeach}{HTML}{FCF0E8}
\definecolor{lcpGreen}{HTML}{589A39}
\definecolor{lcpGreenLight}{HTML}{E3F3D8}
\definecolor{lcpGreenInner}{HTML}{A9D989}
\definecolor{lcpRed}{HTML}{C82C2C}
\definecolor{lcpOrange}{HTML}{C96C35}
\definecolor{lcpInk}{HTML}{202327}
\definecolor{lcpMuted}{HTML}{81878B}
\definecolor{lcpBorder}{HTML}{A6AAAC}
\definecolor{lcpGray}{HTML}{F0F1F2}

\newcommand{\lcpMainFont}{\fontsize{11}{12.5}\selectfont}
\newcommand{\lcpAuxFont}{\fontsize{9.5}{11}\selectfont}
\newcommand{\lcpFineStroke}{.60pt}
\newcommand{\lcpOutlineStroke}{1.15pt}
\newcommand{\lcpFlowStroke}{1.20pt}
\tikzset{
  lcp primary/.style={font=\lcpMainFont},
  lcp auxiliary/.style={font=\lcpAuxFont},
  lcp outline/.style={line width=\lcpOutlineStroke},
  lcp rule/.style={line width=\lcpFineStroke},
  lcp emphasis/.style={line width=\lcpFlowStroke},
  lcp panel/.style={draw=lcpInk,lcp outline,rounded corners=10pt},
  lcp flow/.style={-{Latex[length=2.0mm,width=1.45mm]},
    draw=lcpInk,line width=\lcpFlowStroke,line join=round},
  lcp box/.style={draw=lcpInk,lcp outline,rounded corners=4pt,
    align=center,inner sep=3pt},
  lcp title/.style={lcp primary},
  lcp small/.style={lcp auxiliary},
  lcp reference/.style={lcp auxiliary,
    text=lcpMuted,inner sep=1.5pt},
  lcp muted/.style={text=lcpMuted},
  pics/lcp response/.style args={#1}{code={
    \path[fill=lcpGray,draw=lcpBorder,lcp outline,rounded corners=3pt]
      (0,0) rectangle (.77,1.02);
    \node[lcp primary,inner sep=0] at (.385,.73) {$a^{#1}$};
    \foreach \yy in {.20,.32,.44}{
      \draw[lcpBorder!65,lcp rule] (.12,\yy)--(.65,\yy);
    }
  }},
  pics/lcp prompt/.style={code={
    \path[fill=lcpGray,draw=lcpBorder,lcp outline]
      (.18,.24)--(.28,.24)--(.33,.02)--(.71,.24)--(1.65,.24)
      .. controls (1.76,.24) and (1.83,.31) .. (1.83,.42)
      --(1.83,1.41)
      .. controls (1.83,1.52) and (1.76,1.59) .. (1.65,1.59)
      --(.18,1.59)
      .. controls (.07,1.59) and (0,1.52) .. (0,1.41)
      --(0,.42)
      .. controls (0,.31) and (.07,.24) .. (.18,.24)--cycle;
    \node[lcp primary,inner sep=0] at (.915,1.21) {Prompt $s$};
    \foreach \yy in {.61,.78,.95}{
      \draw[lcpBorder!65,lcp rule] (.16,\yy)--(1.66,\yy);
    }
  }},
  pics/lcp thumb/.style args={#1}{code={
    \path[fill=#1,draw=black,lcp rule,rounded corners=.4pt]
      (0,.10) rectangle (.12,.40);
    \path[fill=#1,draw=black,lcp rule,line join=round]
      (.18,.10)
      .. controls (.22,.08) and (.25,.07) .. (.29,.07)
      --(.49,.07)
      .. controls (.54,.07) and (.55,.12) .. (.53,.15)
      .. controls (.60,.17) and (.60,.22) .. (.56,.24)
      .. controls (.63,.27) and (.63,.32) .. (.58,.34)
      .. controls (.66,.37) and (.64,.44) .. (.58,.44)
      --(.38,.44)
      .. controls (.41,.51) and (.43,.60) .. (.40,.66)
      .. controls (.37,.70) and (.32,.68) .. (.32,.62)
      .. controls (.32,.54) and (.27,.47) .. (.18,.43)
      --cycle;
  }},
  pics/lcp annotator/.style={code={
    \draw[lcpInk,lcp outline] (0,.35) circle[radius=.14];
    \draw[lcpInk,lcp outline] (-.29,-.12)--(-.29,.03)
      .. controls (-.29,.22) and (.29,.22) .. (.29,.03)--(.29,-.12);
  }}
}

\title{LatentCP-DPO: Conformal Pessimism for DPO with Finite-Sample Guarantees}

\author{Hongwei Xu\thanks{Equal Contribution.}\\
Xi'an Jiaotong University\\
National University of Singapore\\
\texttt{bracy.xu@gmail.com} \\
\And
Jingyi Liu\footnotemark[1]\\
King Abdullah University of Science and Technology \\
\texttt{jingyi.liu2657@gmail.com} \\
\And
Vincent Y. F. Tan\\
National University of Singapore\\
\texttt{vtan@nus.edu.sg}\\
\And
Zai Yang\\
Xi'an Jiaotong University\\
\texttt{yangzai@xjtu.edu.cn}\\
\And
Yulian Wu\thanks{Corresponding to:yulian.wu@nus.edu.sg} \\
National University of Singapore\\
\texttt{yulian.wu@nus.edu.sg}
}

\iclrfinalcopy 
\begin{document}
\maketitle
\lhead{Preprint}      


\begin{abstract}
Direct Preference Optimization (DPO) learns a response policy from human preference data, but with a limited dataset it can overfit to noisy or weakly informative preference labels.
We introduce LatentCP-DPO, a framework that combines conformal prediction with pessimistic preference optimization for Bradley–Terry (BT) pairwise feedback and Plackett–Luce (PL) ranking feedback.
Building on LatentCP’s construction of uncertainty sets for unobserved model parameters, we address how to incorporate this uncertainty into preference optimization and characterize its effect on policy performance.
Under exchangeability and a correctly specified preference model, we construct uncertainty sets for latent relative rewards with finite-sample marginal coverage from conformal prediction sets for observable preferences.
We then learn a response policy by maximizing the worst-case expected preference log-likelihood over these uncertainty sets, using an independent collection of prompts and candidate responses without preference labels.
Under suitable assumptions, we establish finite-sample upper bounds on regularized policy suboptimality that separate policy-class approximation error, conformal uncertainty, and statistical estimation error.
For BT feedback, we further establish a worst-case lower bound for LatentCP-DPO that matches the upper bound’s reciprocal dependence on the number of unlabeled optimization examples.
The lower bound also shows that the uncertainty sets can induce a bias that persists as the optimization sample grows.
Together, these results characterize both the statistical accuracy and potential limitations of preference optimization using conformal uncertainty sets.
Experiments support the benefits of the proposed framework in finite preference datasets.
\end{abstract}

\newpage
\setcounter{tocdepth}{2}   
\tableofcontents
\newpage 

\section{Introduction}
Direct Preference Optimization (DPO) simplifies language-model alignment with human preferences by replacing separate reward-model training and reinforcement-learning optimization with a single stage of policy training \citep{rafailov2023direct}.
However, DPO can overfit to finite preference datasets \citep{azar2024general, rafailov2024scaling}, while noisy and inconsistent labels further complicate reliable preference learning \citep{chen2026conformal}.
The standard DPO objective fits a policy-induced preference model to observed labels without explicitly quantifying statistical uncertainty in the inferred preferences.
Policy updates may therefore place undue weight on comparisons that are weakly supported by the available data.
This motivates incorporating statistically grounded uncertainty quantification into preference optimization.

Under a preference model, the probabilities of observed comparisons or rankings depend on unobserved reward differences between candidate responses.
Although DPO avoids a separately trained reward model, its preference objective still depends on policy-induced reward differences.
The uncertainty of interest therefore concerns the latent relative rewards governing the observed preferences.
Conformal prediction (CP) provides a distribution-free framework for constructing prediction sets for observable outcomes with finite-sample marginal coverage under exchangeability \citep{shafer2008tutorial, angelopoulos2023conformal}.
Under suitable sampling assumptions, LatentCP extends this construction to uncertainty sets for unobserved distributional parameters through a correctly specified forward model \citep{zheng2026beyond}.
Using these sets in preference optimization raises two questions: 
\emph{How should latent reward uncertainty be incorporated into the policy objective}, and 
\emph{what performance guarantees can be established for the resulting policy}?

\begin{figure}[t]
    \centering
    \resizebox{\linewidth}{!}{%
\begin{tikzpicture}[
  x=1cm,y=1cm,font=\lcpMainFont,
  text=lcpInk,every node/.style={inner sep=1.5pt},
  line cap=round,line join=round
]
  \path[use as bounding box] (0,.36) rectangle (20.8,12.05);
  \fill[white] (0,.36) rectangle (20.8,12.05);

  \begin{scope}[yshift=.40cm]
  \draw[lcp panel]
    (.08,8.32) rectangle (8.60,11.55);
  \pic[scale=1.10] at (.30,9.71) {lcp prompt};
  \node[lcp title] (responseheading) at (4.30,11.25) {Candidate responses};
  \pic[scale=1.10] at (2.65,9.85) {lcp response=1};
  \pic[scale=1.10] at (3.84,9.85) {lcp response=2};
  \pic[scale=1.10] at (5.03,9.85) {lcp response=3};
  \pic[scale=.65] at (2.46,10.53) {lcp thumb=black};
  \pic[scale=.65] at (3.65,10.53) {lcp thumb=white};
  \draw[lcpInk,lcp outline,decorate,decoration={brace,amplitude=3pt}]
    (6.40,11.46)--(6.40,9.86);
  \node[lcp primary] (contextlabel)
    at (7.48,10.66) {Context $D$};
  \pic at (.60,9.20) {lcp annotator};
  \node[fill=lcpGreenLight,minimum width=4.78cm,minimum height=.58cm,
    lcp title] (preference) at (3.85,9.39)
    {Preference: $a^1\succ a^2\succ a^3$};
  \draw[lcpInk,lcp outline,decorate,decoration={brace,amplitude=3pt}]
    (6.40,9.75)--(6.40,9.04);
  \node[lcp primary] (outcomelabel)
    at (7.48,9.39) {Outcome $y$};
  \draw[lcpBorder,lcp rule] (.39,8.99)--(8.29,8.99);
  \node[lcp title] (labeledfooter) at (4.34,8.65) {Labeled preference data};

  \node[lcp box,fill=lcpPilot,minimum width=3.68cm,minimum height=.91cm,
    lcp title] (pilot) at (12.04,11.05)
    {\textbf{Pilot split}\\[2pt]
      {\lcpAuxFont\color{lcpMuted}\lcpPilotRef}};
  \node[lcp box,fill=lcpPeach,minimum width=3.68cm,minimum height=.91cm,
    lcp title] (cal) at (12.04,9.80)
    {\textbf{Calibration split}\\[2pt]
      {\lcpAuxFont\color{lcpMuted}\lcpCalibrationRef}};
  \draw[lcpInk,lcp emphasis] (8.60,10.43)--(9.53,10.43);
  \draw[lcp flow] (9.53,10.43)|-(pilot.west);
  \draw[lcp flow] (9.53,10.43)|-(cal.west);
  \node[lcp small] (splitlabel) at (9.06,10.70) {Split};
  \node[lcp box,fill=lcpPilot,minimum width=5.72cm,minimum height=.91cm,
    lcp title] (fit) at (17.85,11.05)
    {Fit pilot policy $\widehat\pi_{\mathrm{pil}}$};
  \draw[lcp flow] (pilot.east)--(fit.west);
  \node[lcp box,fill=lcpBlueLight,minimum width=5.72cm,minimum height=1.02cm,
    lcp title] (score) at (17.85,9.73)
    {Define nonconformity score\\[2pt]$\widehat s(D,y)$};
  \draw[lcp flow] (fit.south)--(score.north);
  \end{scope}

  \path[lcp panel,rounded corners=4pt,fill=lcpGreenLight!65!white]
    (10.20,7.94) rectangle (20.08,9.10);
  \node[lcp primary] (quantile) at (15.14,8.79)
    {Conformal $(1-\gamma)$-quantile of
      $\{\widehat s(D_i^{\mathrm{cal}},y_i^{\mathrm{cal}})\}$:
      $\widehat q^{\mathrm{cal}}_{1-\gamma}$};
  \coordinate (caljoin) at (12.04,9.42);
  \draw[lcpInk,lcp emphasis] (cal.south)--(caljoin);
  \draw[lcpInk,lcp emphasis] (score.south)--(17.85,9.42)--(caljoin);
  \draw[lcp flow] (caljoin)--(12.04,9.10);
  \fill[lcpInk] (caljoin) circle[radius=.026];
  \node[lcp reference] (gammanote) at (15.14,8.22)
    {Any $\gamma\in(0,\alpha)$ may be specified before calibration.};

  \draw[lcp flow] (12.04,7.94)--(12.04,7.55);
  \coordinate (scorestart) at (20.40,0 |- score.south);
  \coordinate (scoreend) at (20.40,7.55);
  \draw[lcp flow] (scorestart)--(scoreend);
  \node[lcp auxiliary,anchor=west,xshift=.07cm] (scoreinput)
    at ($(scorestart)!.5!(scoreend)$) {$\widehat s$};

  \begin{scope}[yshift=-1.08cm]
  \draw[lcp panel]
    (.08,3.55) rectangle (3.90,8.55);
  \begin{scope}[xshift=.225cm]
  \pic at (.85,6.70) {lcp prompt};
  \pic at (.29,5.48) {lcp response=1};
  \pic at (1.38,5.48) {lcp response=2};
  \pic at (2.47,5.48) {lcp response=3};
  \node[lcp title] (optcontext)
    at (1.765,5.16) {Context $D$};
  \draw[lcpBorder,lcp rule]
    (.135,4.77)--(3.395,4.77);
  \node[align=center,lcp primary] (optfooter)
    at (1.765,4.42) {\textbf{Optimization split}};
  \node[lcp auxiliary] (optqualifier)
    at (1.765,3.89) {Label-free contexts};
  \end{scope}
  \draw[lcp flow, line width=\lcpOutlineStroke, dash pattern=on 5pt off 3pt on .7pt off 3pt] (3.90,5.55)--(5.08,5.55);
  \node[align=center,lcp small] (contextarrow)
    at (4.49,6.04) {Each\\[-1pt]context};
  \node[lcp small] (contextsymbol)
    at (4.49,5.24) {$D$};

  \draw[lcp panel,
    dash pattern=on 5pt off 3pt on .7pt off 3pt]
    (5.15,2.97) rectangle (20.72,8.55);

  \begin{scope}[xshift=-.50cm]
  \node[lcp primary] (predictiontitle)
    at (8.07,8.18) {Conformal prediction set};
  \node[lcp reference] (predictionref) at (8.07,7.63) {\lcpPredictionRef};
  \draw[lcpBlue,lcp outline,rounded corners=6pt,
    dash pattern=on 4pt off 3pt] (5.95,4.54) rectangle (10.15,7.28);
  \path[fill=lcpBlueLight,draw=lcpBlue,lcp outline,
    rounded corners=4pt,dash pattern=on 3pt off 2pt]
    (6.12,5.94) rectangle (9.98,7.18);
  \node[lcp primary,anchor=west] (rankone)
    at (6.36,6.84)
    {\scalebox{1.12}{${\color{lcpRed}a^1}$}$\succ a^2\succ$\scalebox{1.12}{${\color{lcpRed}a^3}$}};
  \node[lcp primary,anchor=west] (ranktwo)
    at (6.36,6.28)
    {$a^2\succ$ \scalebox{1.12}{${\color{lcpRed}a^1}$}$\succ$\scalebox{1.12}{${\color{lcpRed}a^3}$}};
  \draw[lcpInk,lcp rule] (9.09,6.10)--(9.09,7.02);
  \node[lcp primary] (csymbol)
    at (9.57,6.56) {$\widehat{\mathcal C}_{\gamma}$};
  \node[lcp primary,text=lcpMuted,anchor=west]
    (excludedrank) at (6.36,5.52) {$a^3\succ a^1\succ a^2\quad\cdots$};
  \draw[lcpBlue,lcp rule] (6.17,5.22)--(9.93,5.22);
  \node[lcp auxiliary] (outcomespace)
    at (8.05,4.87) {Outcome space $\mathcal Y(D)$};
  \end{scope}

  \begin{scope}[xshift=-.40cm]
  \draw[lcp flow] (10.24,6.48)--(10.81,6.48);
  \node[lcp primary] (latenttitle)
    at (12.79,8.18) {Latent reward uncertainty set};
  \node[lcp auxiliary] (compatibility)
    at (12.79,7.58)
    {$P_{\mathrm M, \boldsymbol{u}}(\widehat{\mathcal C}_{\gamma}(D)\mid D)
      \geq 1-\gamma/\alpha$};
  \path[fill=lcpGreenLight,draw=lcpGreen!85,lcp outline,
    dash pattern=on 4pt off 3pt]
    (10.93,7.02)
      .. controls (11.09,7.36) and (11.42,7.36) .. (11.99,7.18)
      .. controls (12.90,7.00) and (13.75,7.04) .. (14.36,7.11)
      .. controls (14.77,7.14) and (14.80,6.55) .. (14.88,5.70)
      .. controls (14.98,5.12) and (14.17,4.83) .. (13.28,4.79)
      .. controls (12.00,4.68) and (11.10,4.76) .. (10.91,5.18)
      .. controls (10.77,5.45) and (11.83,5.75) .. (11.58,6.09)
      .. controls (11.33,6.44) and (10.70,6.66) .. (10.93,7.02)
      --cycle;
  \node[lcp auxiliary] (latentspace)
    at (12.87,6.78) {Relative reward space};
  \path[fill=lcpGreenInner,draw=lcpGreen!70!black,lcp outline,
    dash pattern=on 4pt off 3pt]
    (12.40,5.96)
      .. controls (12.80,6.23) and (13.67,6.30) .. (14.16,6.05)
      .. controls (14.66,5.87) and (14.61,5.43) .. (14.14,5.20)
      .. controls (13.67,4.99) and (12.82,4.90) .. (12.59,5.08)
      .. controls (12.28,5.30) and (12.19,5.70) .. (12.40,5.96)
      --cycle;
  \node[lcp primary] (usymbol)
    at (13.19,5.53) {$\widehat{\mathcal U}_{\alpha}$};
  \node[lcp auxiliary,text=black] (thetasymbol)
    at (11.59,5.20) {$\Theta(D)$};
  \fill[lcpGreen!80!black] (13.86,5.62) circle[radius=.060];
  \node[lcp small,text=lcpGreen!65!black] (truelatent)
    at (14.19,5.62) {$\boldsymbol{u}^\star$};
  \node[lcp auxiliary] (coverage)
    at (10.54,4.19)
    {{\color{lcpMuted}\lcpCoverageRef:}\enspace
    $\mathbb P\{\boldsymbol{u}^\star(D)\in\widehat{\mathcal U}_{\alpha}(D;\gamma)\}
      \geq 1-\alpha(1+\Delta/\gamma)$};
  \end{scope}

  \draw[lcp flow] (14.56,6.48)--(15.18,6.48);
  \node[lcp primary] (worsttitle)
    at (17.96,8.18) {Worst-case loss};
  \draw[lcpBlue!80!black,lcp outline] (15.68,7.56)--(20.36,7.56);
  \foreach \xx/\yy in {16.09/7.00,16.75/7.18,17.41/7.28,
      18.82/7.02,19.54/7.13}{
    \draw[lcpBlue!75,lcp emphasis] (\xx,7.56)--(\xx,\yy);
    \fill[lcpBlue!85] (\xx,\yy) circle[radius=.060];
  }
  \draw[lcpOrange,lcp emphasis] (18.14,7.56)--(18.14,6.60);
  \fill[lcpOrange] (18.14,6.60) circle[radius=.060];
  \node[lcp small,text=lcpOrange,anchor=west] (worstlabel)
    at (18.25,6.62) {Worst case};
  \path[fill=lcpPeach,rounded corners=3pt]
    (15.32,3.85) rectangle (20.58,6.34);
  \node[lcp reference] (worstref) at (17.95,6.07) {\lcpPessimisticRef};
  \node[lcp primary] (pessformula)
    at (17.95,4.845)
    {$\begin{gathered}
      \mathcal L^{\mathrm{pess}}_{\alpha,\gamma}(\pi;D)
      =\inf_{\boldsymbol{u}\in\widehat{\mathcal U}_{\alpha}(D;\gamma)}\\[2pt]
      \mathop{\mathbb E}\limits_{y\sim P_{\mathrm M,u}(\cdot\mid D)}
      \!\left[\log p_{\mathrm M,\pi}(y\mid D)\right]
    \end{gathered}$};

  \draw[lcpBlue,lcp emphasis] (5.46,3.78)--(14.30,3.78);
  \draw[lcpGreen,lcp emphasis] (14.60,3.78)--(20.48,3.78);
  \node[lcp primary] (latentfooter)
    at (9.91,3.40)
    {\textbf{(a) LatentCP}\enspace
      {\lcpAuxFont\color{lcpMuted}(\lcpLatentCPSectionRef)}};
  \node[lcp primary] (pessfooter)
    at (17.54,3.40)
    {\textbf{(b) Pessimistic evaluation}\enspace
      {\lcpAuxFont\color{lcpMuted}(\lcpPessimisticSectionRef)}};
  \end{scope}

  \draw[lcp flow]
  (5.15,2.15) -| (2.39,1.63);
  \begin{scope}[yshift=.42cm]
  \path[fill=lcpGray!65!white,lcp panel,rounded corners=4pt]
    (.08,0) rectangle (20.72,1.16);
  \node[font=\lcpMainFont\bfseries,anchor=west]
    (opttitle) at (.40,.58) {Policy optimization};
  \node[lcp reference,anchor=east] (optref)
    at (20.39,.58) {\lcpOptimizationRef};
  \node[lcp primary] (optformula) at (11.4,.58)
    {$\displaystyle\widehat\pi_{\mathrm{LCP}}
      \in\operatorname*{arg\,max}_{\pi\in\Pi}
      \frac{1}{n_{\mathrm{opt}}}
      \sum\nolimits_{i=1}^{n_{\mathrm{opt}}}
      \mathcal L^{\mathrm{pess}}_{\alpha,\gamma}(\pi;D_i^{\mathrm{opt}})$};
  \end{scope}

\end{tikzpicture}
    }
    \caption{Overview of LatentCP-DPO framework. Conformal prediction sets for preference outcomes are inverted into uncertainty sets for latent relative rewards, which define the pessimistic policy objective. PL ranking feedback is shown for illustration.}
    \label{fig:overview}
\end{figure}
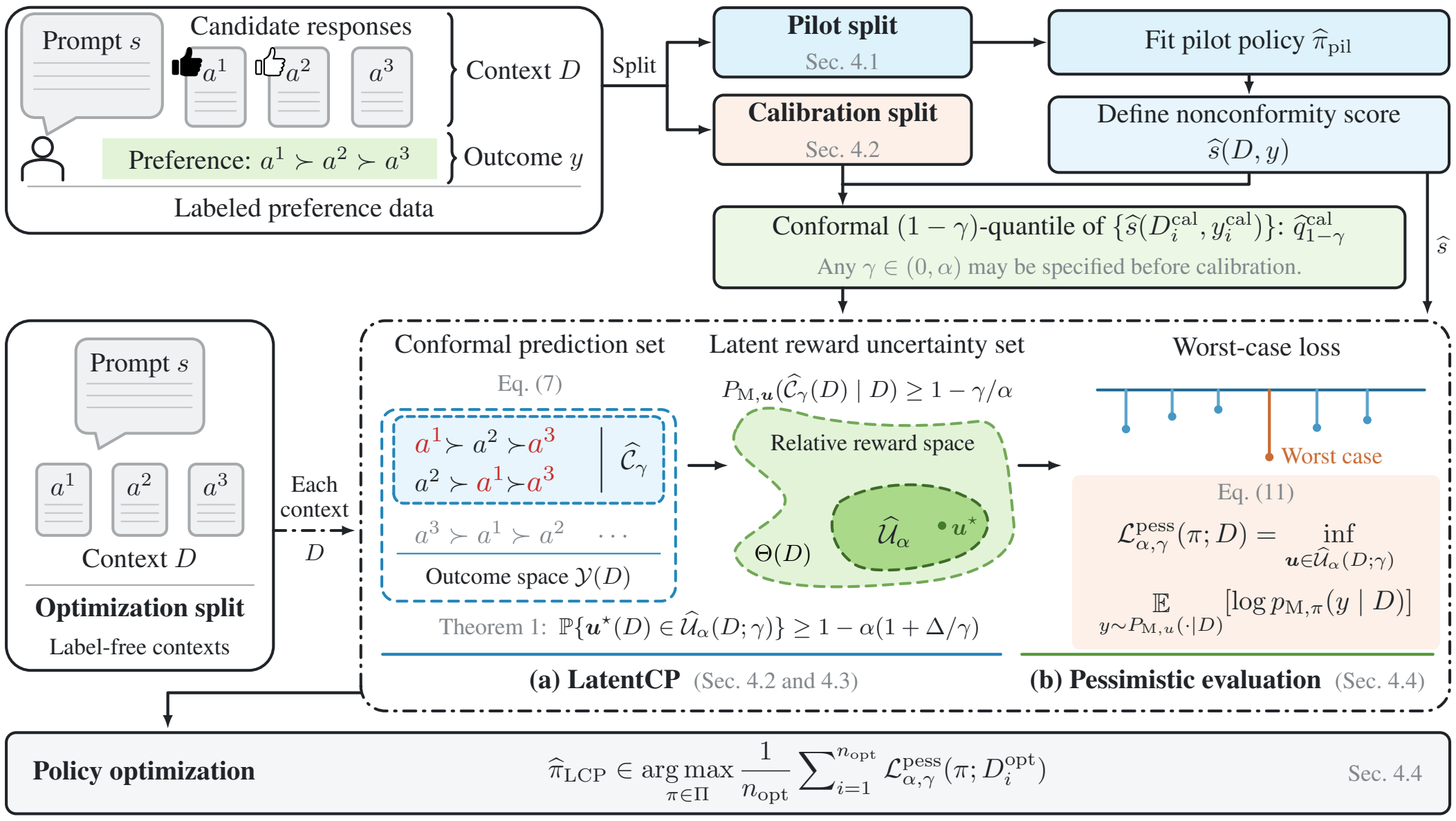

To address these questions, we propose \emph{LatentCP-DPO}, a framework that combines conformal prediction with pessimistic preference optimization for Bradley--Terry (BT) pairwise feedback and Plackett--Luce (PL) ranking feedback; see Fig.~\ref{fig:overview}.
Building on the latent-inference construction of \citet{zheng2026beyond}, our contributions are as follows.

\begin{itemize}[leftmargin=*]
    \item \textbf{Preference optimization using latent reward uncertainty.}
    We first fit an initial (pilot) policy on a labeled pilot sample and use a separate labeled calibration sample to construct conformal prediction sets for observable preferences.
    Specializing the LatentCP construction to preference models, we convert these prediction sets into uncertainty sets for latent relative rewards with finite-sample marginal coverage under suitable assumptions.
    These sets induce a family of preference distributions.
    We then learn a response policy by maximizing the worst-case expected preference log-likelihood over this family, using an independent collection of prompts and candidate responses without preference labels.

    \item \textbf{Concrete objectives and algorithms for pairwise and ranking feedback.}
    For BT feedback, we derive a closed-form pessimistic objective that makes explicit how conformal uncertainty moderates policy-induced preference margins.
    For PL feedback, we develop a feasibility-preserving minorization--maximization (MM) algorithm to approximately solve the inner worst-case optimization problem.

    \item \textbf{A unified finite-sample upper bound.}
    Under stated assumptions, we establish an upper bound on KL-regularized policy suboptimality for the exact empirical optimizer.
    The bound separates policy-class approximation error, conformal uncertainty and calibration effects, and statistical estimation error from the independent unlabeled optimization sample.

    \item \textbf{A complementary lower bound and persistent bias.}
    For BT feedback, we establish a worst-case lower bound for LatentCP-DPO, conditional on suitable pilot realizations.
    At fixed confidence, this bound matches the upper bound's $d_\Pi/n_{\mathrm{opt}}$ dependence.
    It also shows that, with calibration sample size and coverage levels fixed, bias induced by the uncertainty sets can persist as the optimization sample grows.

    \item \textbf{Empirical improvements in prediction-set efficiency.}
    Evaluated over three matched policy seeds and 3,000 test contexts on Nectar with Qwen3-8B ($K = 6$), LatentCP-DPO reduces the average split-conformal ranking prediction-set size at a common requested coverage level by $2.78\%$ relative to Pure DPO and $10.09\%$ relative to the ROPO adaptation, while maintaining comparable pairwise accuracy ($66.6\%$ versus $66.5\%$ for Pure DPO).
    Experiments on Llama-3.1-8B-Instruct and Llama-2-7B-Chat show similar improvements in prediction-set efficiency.
    An ablation comparing the pilot and final policies further shows an improved coverage--set-size trade-off after preference optimization.
\end{itemize}

\section{Related Work}
We discuss the most relevant work here and defer additional discussion to Appendix~\ref{app:additional_related_work}.

\textbf{Provable and Pessimistic Preference Optimization.}
Provable preference learning has been studied under BT/PL feedback \citep{zhu2023principled} and with general function approximation in offline RL \citep{zhan2024provable}, with sharp guarantees for KL-regularized objectives and extensions to private preference labels \citep{zhao2026sharp, wu2025offline}.
Pessimistic methods use adversarial rewards or critics \citep{liu2024provably, ji2024self}, regularization and uncertainty penalties \citep{huang2025correcting, houliston2024uncertainty}, or policy ensembles \citep{barla2026provably}.
Distributionally robust preference optimization addresses noisy preference data, distribution shift, and PL ranking-label uncertainty \citep{wu2025towards, xu2026robust, wu2026distributionally}.
Closely related, \citet{kim2025lightweight} optimize over $\chi^2$-divergence ambiguity sets of conditional preference distributions, obtaining closed-form inner solutions.
Our ambiguity sets are induced by conformally calibrated uncertainty sets for latent relative rewards, and our analysis connects these sets to policy suboptimality and the size of an independent unlabeled optimization sample.

\textbf{Conformal Inference for Latent Parameters.}
Related methods infer latent quantities using simulated parameter--observation pairs \citep{patel2024variational} or pseudo-outcomes for treatment effects \citep{alaa2023conformal}.
LatentCP constructs uncertainty sets for unobserved distributional parameters without latent calibration labels, under suitable sampling assumptions and a correctly specified forward model \citep{zheng2026beyond}.
We adapt this construction to relative rewards under BT and PL feedback and establish guarantees for the resulting preference-optimization procedure.

\textbf{Conformal Alignment.}
Conformal calibration has been used in robust optimization \citep{yeh2024end} and policy risk control \citep{prinster2026conformal}.
For alignment, conformal reward-interval widths have been used to reweight GRPO advantages \citep{pan2026uncertainty}, while response-level conformal reliability has been used to weight DPO preference pairs \citep{chen2026conformal}.
In our framework, calibrated uncertainty sets define the preference distributions considered by the inner worst-case optimization, directly determining the pessimistic training objective.

\section{Preliminaries}
\textbf{Direct Preference Optimization (DPO).}
Let $s \sim \rho$ be a prompt, let $a \in \mathcal{A}(s)$ be a response, and let $\pi_{\mathrm{ref}}$ be a fixed reference policy.
Following \citet{rafailov2023direct}, given a latent reward $r^\star$ and a regularization parameter $\beta > 0$, consider the KL-regularized problem
\begin{equation} \label{eq:kl_regularized_objective}
    \pi^\star \in \arg\max_{\pi} J_{\beta}(\pi),
    \qquad
    J_{\beta}(\pi) = J(\pi) - \beta H(\pi)
\end{equation}
where $J(\pi) = \mathbb{E}_{\substack{s \sim \rho,\ a \sim \pi(\cdot | s)}} [r^\star(s, a)]$ and $H(\pi) = \mathbb{E}_{\substack{s \sim \rho}}[D_{\mathrm{KL}}(\pi(\cdot | s) \Vert \pi_{\mathrm{ref}}(\cdot | s))]$.
Assuming that $\pi_{\mathrm{ref}}$ has full support and $0 < Z_\beta^\star(s) < \infty$ for $\rho$-almost every $s$, an optimizer of (\ref{eq:kl_regularized_objective}) is given by
$ \pi^\star(a | s)= \frac{1}{Z_\beta^\star(s)} \pi_{\mathrm{ref}}(a | s) \exp\left({r^\star(s, a)}/{\beta}\right),$
where $Z_\beta^\star(s) = \sum_{a'\in\mathcal{A}(s)} \pi_{\mathrm{ref}}(a' | s) \exp(r^\star(s, a') / \beta)$.
Equivalently,
\begin{equation} \label{eq:kl_regularized_solution_r_star}
    r^\star(s, a)
    = \beta \log \frac{\pi^\star(a | s)}{\pi_{\mathrm{ref}}(a | s)} + \beta \log Z_\beta^\star(s).
\end{equation}
The final term depends on the prompt but not on the response and therefore cancels from within-prompt reward comparisons.
Accordingly, for any policy $\pi$, define its implicit reward score as $r_\pi(s,a) = \beta\log\frac{\pi(a | s)}{\pi_{\mathrm{ref}}(a | s)}$.

\textbf{Preference Models and the DPO Objective.}
Let $\{(D_i, y_i)\}_{i=1}^n$ be observed preference units, where $D_i = (s_i, a_i)$ denotes the label-free context and $a_i$ is an ordered tuple of candidate responses.
For pairwise feedback, $a_i = (a_i^0, a_i^1)$ and $y_i \in \{0,1\}$, where the response indices are fixed before the preference is observed and $y_i = 1$ indicates $a_i^1 \succ a_i^0$.
Under the Bradley--Terry (BT) model \citep{bradley1952rank}, the policy-induced likelihood is
\begin{equation} \label{eq:bt_model}
    p_{\mathrm{BT}, \pi}(y_i \mid D_i)
    = \sigma \bigl(r_{\pi}(s_i, a_i^1) - r_{\pi}(s_i, a_i^0)\bigr)^{y_i} \cdot \sigma\bigl(r_{\pi}(s_i, a_i^0) - r_{\pi}(s_i, a_i^1)\bigr)^{1 - y_i},
\end{equation}
where $\sigma(t) = 1 / (1+e^{-t})$ is the logistic sigmoid function. 
For complete-ranking feedback, write $a_i = (a_i^1, \ldots, a_i^K)$, where $K \geq 2$, and let $y_i = (y_{i,1}, \ldots, y_{i,K})$ be a permutation of $\{1, \ldots, K\}$ that orders the candidate responses from most to least preferred.
Under the Plackett--Luce (PL) model \citep{plackett1975analysis, luce2005individual}, the policy-induced likelihood is
\begin{equation}
    \label{eq:pl_model}
    p_{\mathrm{PL}, \pi}(y_i | D_i)
    = \prod_{k=1}^{K-1} \frac{\exp\left(r_{\pi}(s_i, a_i^{y_{i, k}})\right)}{\sum_{\ell=k}^{K} \exp\left(r_{\pi}(s_i, a_i^{y_{i, \ell}})\right)}.
\end{equation}
Throughout, $\mathrm{M} \in \{\mathrm{BT}, \mathrm{PL}\}$ denotes the feedback model, which is fixed in advance.

Let $\Pi$ be a pre-specified policy class for which all relevant policy log-ratios and preference log-likelihoods are finite.
Define the unit-level negative log-likelihood loss as $\ell_i(\pi) = -\log p_{\mathrm{M},\pi}(y_i \mid D_i)$.
Assuming that the empirical minimum is attained, the empirical DPO estimator is defined by
\begin{equation} \label{eq:empirical_dpo_estimator}
    \widehat{\pi}_{\mathrm{DPO}} \in \operatorname*{argmin}_{\substack{\pi \in \Pi}} 
    \frac{1}{n}\sum_{i=1}^n  \ell_i(\pi).
\end{equation}

\textbf{KL-Regularized Suboptimality Gap.}
For a learned policy $\widehat{\pi}$ and a comparator policy $\pi^\dagger$, define the KL-regularized performance gap as $\mathrm{SubOpt}_{\beta}(\widehat{\pi}, \pi^\dagger) = J_{\beta}(\pi^\dagger) - J_{\beta}(\widehat{\pi})$.
The comparator $\pi^\dagger$ need not be optimal, so these relative performance gaps are not necessarily nonnegative.

\textbf{Conformal Prediction.}
Conformal prediction (CP), in its split form, uses a fitted predictor and a separate calibration sample to construct prediction sets for observable outcomes, with distribution-free finite-sample marginal coverage under exchangeability \citep{angelopoulos2024theoretical}.
Here, \emph{exchangeability} means that, conditional on the data used to fit the score, the joint distribution of the calibration units and a future unit is invariant under permutations.
In our setting, preference outcomes are observable, whereas relative rewards are latent; uncertainty sets for the latter are developed in Section~\ref{sec:uncertainty_sets_for_latent_relative_rewards}.

\section{The LatentCP-DPO Framework} \label{sec:methods_and_algorithms}
We partition the $n$ labeled preference units into disjoint pilot and calibration subsets, $\{1, \ldots, n\} = \mathcal{I}_{\mathrm{pil}} \mathbin{\dot\cup} \mathcal{I}_{\mathrm{cal}}$.
We use the pilot subset to fit a pilot DPO policy and construct the nonconformity score, and the calibration subset to calibrate the conformal prediction sets.
Let $D_1^{\mathrm{opt}}, \ldots, D_{n_{\mathrm{opt}}}^{\mathrm{opt}}$ be label-free preference contexts drawn from the target distribution independently of the pilot and calibration data.
We use these contexts to form the sample-average pessimistic DPO objective.

\subsection{Pilot Policy and Nonconformity Score} \label{sec:pilot_policy_and_nonconformity_score}
Let $n_{\mathrm{pil}} = |\mathcal{I}_{\mathrm{pil}}|$. 
Applying the empirical DPO estimator defined in (\ref{eq:empirical_dpo_estimator}) to the pilot subset, we obtain the pilot policy $\widehat{\pi}_{\mathrm{pil}}$.
Using this pilot policy, we define the nonconformity score as
\begin{equation} \label{eq:nonconformity_score}
    \widehat{s}(D, y) = -\log p_{\mathrm{M}, \widehat{\pi}_{\mathrm{pil}}}(y \mid D).
\end{equation}
A larger value of $\widehat{s}(D, y)$ corresponds to a lower probability assigned to $y$ by the pilot-induced preference distribution and, therefore, to greater nonconformity.
After fitting, both $\widehat{\pi}_{\mathrm{pil}}$ and the score function $\widehat{s}$ are  fixed throughout the subsequent calibration and pessimistic optimization stages.

\subsection{Conformal Calibration for Preference Outcomes} \label{sec:conformal_calibration}
Fix a target miscoverage level $\alpha \in (0,1)$ and an auxiliary preference-outcome miscoverage level $\gamma \in (0, \alpha)$, chosen independently of the calibration data and the future preference unit.
Let $n_{\mathrm{cal}} = |\mathcal{I}_{\mathrm{cal}}|$.
For each $i \in \mathcal{I}_{\mathrm{cal}}$, define the calibration score $S_i^{\mathrm{cal}} = \widehat{s}(D_i,y_i)$ and let $S_{(1)}^{\mathrm{cal}} \leq \cdots \leq S_{(n_{\mathrm{cal}})}^{\mathrm{cal}}$ denote the corresponding order statistics.
Set $k_\gamma = \lceil(n_{\mathrm{cal}}+1)(1-\gamma)\rceil$ and define the conformal threshold $\widehat{q}_{1-\gamma}^{\mathrm{cal}} = S_{(k_\gamma)}^{\mathrm{cal}}$ for $k_\gamma \leq n_{\mathrm{cal}}$, and $\widehat{q}_{1-\gamma}^{\mathrm{cal}} = +\infty$ for $k_\gamma = n_{\mathrm{cal}} + 1$.
For any label-free context $D$, let $\mathcal{Y}(D)$ denote the set of admissible preference outcomes.
The resulting conformal prediction set is
\begin{equation} \label{eq:conformal_prediction_set_for_observable_preference_outcome}
    \widehat{\mathcal{C}}_{\gamma}(D)
    = \left\{y \in \mathcal{Y}(D): \widehat{s}(D, y) \leq \widehat{q}_{1-\gamma}^{\mathrm{cal}}\right\}.
\end{equation}
If the calibration preference units and a future unit $(D_{n+1}, y_{n+1})$ are exchangeable conditional on the pilot data, then $\mathbb{P}\{y_{n+1} \in \widehat{\mathcal{C}}_{\gamma}(D_{n+1})\} \geq 1 - \gamma$.

\subsection{Uncertainty Sets for Latent Relative Rewards} \label{sec:uncertainty_sets_for_latent_relative_rewards}
Define the candidate-level latent reward vector\footnote{Here $K$ is the number of candidate responses. For BT feedback, we use the indexing $(a^0, a^1)$, so $K = 2$ and $\boldsymbol{r}^\star(D) = (r^\star(s,a^0), r^\star(s,a^1))^\top$.} by $\boldsymbol{r}^\star(D) = (r^\star(s,a^1), \ldots, r^\star(s,a^K))^\top \in \mathbb{R}^K$.
Because both the BT and PL models are invariant to a common additive shift applied to all candidate rewards, define the relative-reward vector $\boldsymbol{u}^\star(D)$ by centering $\boldsymbol{r}^\star(D)$.
By construction, $\boldsymbol{1}_K^\top \boldsymbol{u}^\star(D) = 0$.
Accordingly, the latent parameter space is $\Theta(D) = \{\boldsymbol{u} \in \mathbb{R}^K: \boldsymbol{1}_K^\top \boldsymbol{u} = 0\}$.
For the fixed feedback model $\mathrm{M} \in \{\mathrm{BT},\mathrm{PL}\}$ and any $\boldsymbol{u} \in \Theta(D)$, let $P_{\mathrm{M}, \boldsymbol{u}}(\cdot \mid D)$ denote the corresponding preference distribution.
Its probability mass function $p_{\mathrm{M}, \boldsymbol{u}}(y \mid D)$ is obtained from (\ref{eq:bt_model}) or (\ref{eq:pl_model}) by replacing each candidate-level implicit reward $r_\pi(s, a^k)$ with the corresponding coordinate of $\boldsymbol{u}$.

For the $i$-th preference unit $(D_i, y_i)$, let $\boldsymbol{u}_i^\star = \boldsymbol{u}^\star(D_i) \in \Theta(D_i)$ denote its unobserved latent relative reward vector.
If the feedback model $\mathrm{M}$ is correctly specified, the observed preference outcome $y_i$ satisfies
\begin{equation} \label{eq:latent_relative_reward_generation}
    y_i \mid (D_i, \boldsymbol{u}_i^\star) \sim P_{\mathrm{M}, \boldsymbol{u}_i^\star}(\cdot \mid D_i).
\end{equation}
Using this forward model, we define the forward $\gamma$-compatibility of any relative-reward vector $\boldsymbol{u} \in \Theta(D)$ with the conformal prediction set $\widehat{\mathcal{C}}_\gamma(D)$ as
\begin{equation} \label{eq:forward_gamma_compatibility}
    p_\gamma(\boldsymbol{u}, D) 
    = \mathbb{P}_{y \sim P_{\mathrm{M},\boldsymbol{u}}(\cdot \mid D)}\left\{y \in \widehat{\mathcal{C}}_{\gamma}(D)\right\}.
\end{equation}
Thus, $p_{\gamma}(\boldsymbol{u}, D)$ is the probability mass that the preference-outcome distribution induced by $\boldsymbol{u}$ assigns to $\widehat{\mathcal{C}}_{\gamma}(D)$.
The latent relative-reward uncertainty set is defined by 
\begin{equation} \label{eq:latent_relative_reward_uncertainty_set}
    \widehat{\mathcal{U}}_\alpha(D; \gamma) = \left\{\boldsymbol{u} \in \Theta(D): p_{\gamma}(\boldsymbol{u}, D) \geq 1 - \varepsilon_{\alpha, \gamma} \right\},
\end{equation}
where $\varepsilon_{\alpha, \gamma} = \gamma / \alpha$.
To avoid optimizing over an empty set, whenever the set $\widehat{\mathcal{U}}_\alpha(D; \gamma)$ defined above is empty, we adopt the conservative \emph{empty-set convention} $\widehat{\mathcal{U}}_\alpha(D; \gamma) = \Theta(D)$.
We have the following coverage guarantee, with the proof provided in Appendix~\ref{app:latent_coverage_proof}.

\begin{theorem}[Finite-sample marginal coverage guarantee] \label{thm:latent_coverage}
    Fix $\alpha \in (0, 1)$ and $\gamma \in (0, \alpha)$ independently of the calibration data and the future preference unit.
    Suppose that, conditional on the pilot data, the pairs $\left\{(D_i, \boldsymbol{u}_i^\star): i \in \mathcal{I}_{\mathrm{cal}} \cup \{n+1\}\right\}$ are exchangeable.
    Suppose further that, conditional on both the pilot data and these pairs, the preference outcomes $y_i$ are mutually independent with respective distributions $Q(\cdot \mid D_i)$, where $Q$ denotes the true conditional preference distribution.
    Define $\Delta = \mathbb{E}[\operatorname{TV} (Q(\cdot \mid D_{n+1}), P_{\mathrm{M}, \boldsymbol{u}_{n+1}^\star}(\cdot \mid D_{n+1}))]$. 
    Then $\mathbb{P}\{\boldsymbol{u}_{n+1}^\star \notin \widehat{\mathcal{U}}_\alpha(D_{n+1}; \gamma)\} \leq \min \{1, \alpha (1 + \Delta / \gamma) \}$.
\end{theorem}

\subsection{From Latent Uncertainty Sets to Pessimistic DPO} \label{sec:from_latent_cp_to_pessimistic_dpo}
For each $D$, the latent uncertainty set $\widehat{\mathcal{U}}_\alpha(D; \gamma)$ induces the ambiguity set $\{P_{\mathrm{M}, \boldsymbol{u}}(\cdot \mid D): \boldsymbol{u} \in \widehat{\mathcal U}_\alpha(D; \gamma)\}$, which collects the plausible conditional distributions of the preference outcome given $D$.
For any candidate policy $\pi \in \Pi$, we define its unit-level pessimistic DPO objective as the worst-case expected preference log-likelihood over this ambiguity set, namely,
\begin{equation} \label{eq:unit_level_pessimistic_dpo_objective}
    \mathcal{L}_{\alpha, \gamma}^{\mathrm{pess}}(\pi; D)
    = \inf_{\boldsymbol{u} \in \widehat{\mathcal{U}}_\alpha(D; \gamma)} \mathbb{E}_{y \sim P_{\mathrm{M}, \boldsymbol{u}}(\cdot \mid D)}\left[\log p_{\mathrm{M}, \pi}(y \mid D)\right].
\end{equation}
Here, $\boldsymbol{u}$ determines the preference distribution under which the candidate policy is evaluated, while $\pi$ determines the preference log-likelihood being evaluated.
We then define the LatentCP-DPO policy by maximizing the sample-average pessimistic DPO objective over the label-free optimization sample, namely,
\begin{equation*}
    \widehat{\pi}_{\mathrm{LCP}} \in \operatorname*{argmax}_{\pi \in \Pi} \frac{1}{n_{\mathrm{opt}}} \sum_{i=1}^{n_{\mathrm{opt}}} \mathcal{L}_{\alpha,\gamma}^{\mathrm{pess}}(\pi; D_i^{\mathrm{opt}}).
\end{equation*}
Algorithm~\ref{alg:latentcp_dpo} in Appendix~\ref{app:overall_procedure} gives a model-agnostic specification of LatentCP-DPO.

\section{Instantiations for BT and PL Models} \label{sec:instantiations}
We next specialize the uncertainty sets $\widehat{\mathcal{U}}_\alpha$ and inner optimization problems (\ref{eq:unit_level_pessimistic_dpo_objective}) to the BT and PL models.
The BT case admits closed-form characterizations of the latent uncertainty sets and the pessimistic objective, while the PL case is handled by a feasibility-preserving MM algorithm for approximate inner optimization.

\subsection{Bradley--Terry Pairwise Feedback}
For pairwise feedback, the latent parameter space is $\Theta(D) = \{(u^0, u^1)^\top \in \mathbb{R}^2: u^0 + u^1 = 0\}$.
Under the BT model, the preference outcome is binary, so that $\mathcal{Y}(D) = \{0, 1\}$ and $\widehat{\mathcal{C}}_\gamma(D) \in \{\varnothing, \{0\}, \{1\}, \{0, 1\}\}$.
Combining (\ref{eq:latent_relative_reward_uncertainty_set}) with the empty-set convention then yields
\begin{equation} \label{eq:bt_latent_uncertainty_set}
    \widehat{\mathcal{U}}_\alpha(D; \gamma)
    = \begin{cases}
        \{\boldsymbol{u} \in \Theta(D): u^1 - u^0 \geq \log \frac{\alpha - \gamma}{\gamma}\}, & \widehat{\mathcal{C}}_{\gamma}(D) = \{1\}, \\
        \{\boldsymbol{u} \in \Theta(D): u^1 - u^0 \leq -\log \frac{\alpha - \gamma}{\gamma}\}, & \widehat{\mathcal{C}}_{\gamma}(D) = \{0\}, \\
        \Theta(D), & \widehat{\mathcal{C}}_{\gamma}(D) = \{0, 1\} \text{ or } \varnothing.
    \end{cases}
\end{equation}
For any policy $\pi \in \Pi$, define the policy-induced preference margin as $z_\pi(D) = r_\pi(s, a^1) - r_\pi(s, a^0)$.
Taking the infimum of the expected BT log-likelihood over this uncertainty set gives
\begin{equation} \label{eq:bt_unit_level_pessimistic_objective}
    \mathcal{L}_{\alpha, \gamma}^{\mathrm{pess}}(\pi; D)
    = \begin{cases}
        \log\sigma\left(z_\pi(D)\right) - \varepsilon_{\alpha,\gamma} [z_\pi(D)]_+, & \widehat{\mathcal{C}}_\gamma(D) = \{1\},\\
        \log\sigma\left(-z_\pi(D)\right) - \varepsilon_{\alpha,\gamma} [-z_\pi(D)]_+, & \widehat{\mathcal{C}}_\gamma(D) = \{0\},\\
        \log\sigma\left(-|z_\pi(D)|\right), & \widehat{\mathcal{C}}_\gamma(D) = \{0,1\} \text{ or } \varnothing,
    \end{cases}
\end{equation}
where $[t]_+ = \max\{t, 0\}$.
The detailed derivation is provided in Appendix~\ref{app:bt_instantiation}.

The BT instantiation illustrates how conformal uncertainty moderates preference optimization. 
For an observed label $y = 1$, the standard DPO log-likelihood $\log\sigma(z_\pi(D))$ increases as the policy-induced preference probability $\sigma(z_\pi(D))$ approaches one. 
In contrast, when $\widehat{\mathcal{C}}_\gamma(D) = \{1\}$, our uncertainty set allows the opposite outcome to have probability up to $\varepsilon_{\alpha, \gamma}$.
The resulting one-sided penalty in (\ref{eq:bt_unit_level_pessimistic_objective}) accounts for this possibility.
For $\varepsilon_{\alpha,\gamma} < 1/2$\footnote{Both instantiations are defined for $0 < \gamma < \alpha$. The stronger condition $\gamma<\alpha/2$ is used to obtain a nonzero unit-level optimum for a singleton BT prediction set and in the BT lower-bound construction.}, the objective, viewed as a function of an unrestricted scalar margin, is maximized when $\sigma(z_\pi(D)) = 1 - \varepsilon_{\alpha,\gamma}$: \emph{the supported preference direction is retained, but its strength is moderated}.
The case $\widehat{\mathcal{C}}_\gamma(D) = \{0\}$ is symmetric.
When both outcomes are included in $\widehat{\mathcal{C}}_\gamma(D)$, or the empty-set convention applies, the unit-level objective is instead maximized at zero margin, corresponding to equal preference probabilities.
These characterizations explain how the conformal uncertainty set controls the direction and strength favored by the unit-level objective.
Derivations and further discussion are also provided in Appendix~\ref{app:bt_instantiation}.

\subsection{Plackett--Luce Ranking Feedback}
Let $\mathfrak{S}_K = \{y^{(1)}, \ldots, y^{(N_K)}\}$ denote the set of all permutations of $\{1, \ldots, K\}$, where $N_K = K!$, so that $\mathcal{Y}(D) = \mathfrak{S}_K$.
The latent parameter space is $\Theta(D) = \{\boldsymbol{u} = (u^1, \ldots, u^K)^\top \in \mathbb{R}^K : \boldsymbol{1}_K^\top \boldsymbol{u} = 0\}$.
To eliminate the zero-sum constraint, let $B_K \in \mathbb{R}^{K \times (K-1)}$ have columns forming an orthonormal basis of $\boldsymbol{1}_K^\perp$.
Then $\Theta(D) = \{B_K \boldsymbol{v} : \boldsymbol{v} \in \mathbb{R}^{K-1}\}$.

For fixed $D$ and $\pi$, define $q_m(\boldsymbol{v}) = p_{\mathrm{PL}, B_K \boldsymbol{v}}(y^{(m)} \mid D)$, $h_m(\pi) = \log p_{\mathrm{PL}, \pi}(y^{(m)} \mid D)$, and $c_m = \mathbb{I}\{y^{(m)} \in \widehat{\mathcal{C}}_\gamma(D)\}$ for $m = 1, \ldots, N_K$.
Let $\boldsymbol{q}(\boldsymbol{v})$, $\boldsymbol{h}(\pi)$, and $\boldsymbol{c}$ collect these quantities in the same enumeration order.
For a nonempty conformal prediction set, the inner problem (\ref{eq:unit_level_pessimistic_dpo_objective}) admits the exact finite-dimensional reformulation
\begin{equation} \label{eq:pl_pessimistic_objective_computational_formulation}
    \mathcal{L}_{\alpha, \gamma}^{\mathrm{pess}}(\pi; D) 
    = \inf_{\boldsymbol{v} \in \mathbb{R}^{K-1}} \boldsymbol{q}(\boldsymbol{v})^\top \boldsymbol{h}(\pi)
    \quad
    \text{s.t.}
    \quad
    \boldsymbol{c}^\top \boldsymbol{q}(\boldsymbol{v}) \geq 1 - \varepsilon_{\alpha, \gamma}.
\end{equation}
When $\widehat{\mathcal{C}}_\gamma(D) = \varnothing$, the compatibility constraint is omitted under the conservative empty-set convention.
Here, $\boldsymbol{q}(\boldsymbol{v})$ ranges over the ranking distributions induced by the PL family.
The detailed derivation is provided in Appendix~\ref{app:pl_instantiation}.

Although (\ref{eq:pl_pessimistic_objective_computational_formulation}) is exact, the resulting constrained problem need not be convex in $\boldsymbol{v}$ and does not generally admit a closed-form solution.
We therefore develop a feasibility-preserving minorization--maximization (MM) procedure based on an equivalent maximization formulation.
The derivation is provided in Appendix~\ref{app:pl_approximation_of_the_pessimistic_pl_objective}, and the procedure is summarized in Algorithm~\ref{alg:pl_mm_for_pessimistic_pl_inner_problem}.
For fixed $\pi$ and $D$, let $\widetilde{\boldsymbol{v}} = \widetilde{\boldsymbol{v}}(\pi; D)$ denote the feasible approximate solution returned by the algorithm.
We define the resulting approximation to the unit-level pessimistic PL objective as $\widetilde{\mathcal{L}}_{\alpha, \gamma}^{\mathrm{pess}}(\pi; D) = \boldsymbol{q}(\widetilde{\boldsymbol{v}})^\top \boldsymbol{h}(\pi)$.

In practice, we may alternate between updating each $\widetilde{\boldsymbol{v}}_i$ using the MM procedure at the current policy and updating the policy by a standard first-order method while holding the resulting ranking distributions fixed.
This alternating procedure provides a practical approximation to the exact LatentCP-DPO estimator $\widehat{\pi}_{\mathrm{LCP}}$.

\section{Theoretical Results}
To establish finite-sample guarantees for the exact LatentCP-DPO estimator $\widehat{\pi}_{\mathrm{LCP}}$, we introduce the additional assumptions required for the analysis.

\begin{assumption}[Sampling for suboptimality analysis] \label{ass:subopt_sampling}
    Suppose that the sampling conditions of Theorem~\ref{thm:latent_coverage} hold and that the feedback model $\mathrm{M}$ is correctly specified as in~(\ref{eq:latent_relative_reward_generation}).
    Suppose further that, conditional on the pilot sample, the calibration preference units are i.i.d. draws from the target preference-unit distribution.
    The label-free optimization contexts $D_1^{\mathrm{opt}},\ldots,D_{n_{\mathrm{opt}}}^{\mathrm{opt}}$ are i.i.d. draws from its context marginal, independently of the pilot and calibration samples.
\end{assumption}
    
To treat BT and PL feedback uniformly, define the policy-induced comparison score by
\begin{equation*}
    x_\pi(D)
    = \begin{cases}
        r_\pi(s, a^1) - r_\pi(s, a^0), & \mathrm{M} = \mathrm{BT},\\
        \left(r_\pi(s, a^k) - \frac{1}{K}\sum_{\ell=1}^K r_\pi(s, a^\ell)\right)_{k=1}^K, & \mathrm{M} = \mathrm{PL},
    \end{cases}
\end{equation*}
where $K\geq2$ is fixed across all PL preference units.
Define $x^\star(D)$ analogously using $r^\star$.
Thus, $x_\pi(D)=z_\pi(D)$ under BT feedback, whereas $x^\star(D) = \boldsymbol{u}^\star(D)$ under PL feedback.
For BT scores, norms are interpreted as absolute values.

\begin{assumption}[Bounded Comparison Scores] \label{ass:bounded_comparison_scores}
    There exists a constant $0 < B < \infty$ such that, for almost every target context $D$, $\sup_{\pi\in\Pi}\|x_\pi(D)\|_\infty \leq B$ and $ \|x^\star(D)\|_\infty \leq B$.
\end{assumption}
A sufficient condition is that, for $\rho$-almost every prompt $s$, $\sup_{\substack{\pi \in \Pi,\ a \in \mathcal{A}(s)}} |r_\pi(s, a)| \leq B/2$ and $\sup_{a \in \mathcal{A}(s)} |r^\star(s, a)| \leq B/2$.

Define the policy-class approximation error $\mathcal{E}_\Pi = \inf_{\pi \in \Pi} \mathbb{E}_D [\|x_\pi(D) - x^\star(D)\|^2]$, which vanishes when $\pi^\star \in \Pi$.
Let $\eta_\gamma = \mathbb{P}\{|\widehat{\mathcal{C}}_\gamma(D)| \neq 1\}$ denote the probability of an empty or non-singleton conformal set.
We use the refined measure $\widetilde{\eta}_{\alpha, \gamma}$ defined in (\ref{eq:subopt_structural_set_disagreement}), which equals $\eta_\gamma$ for BT feedback.
For PL feedback, it accounts for partial ranking information retained by non-singleton prediction sets.
Here and below, expectations and probabilities over a fresh target context $D$ are conditional on the pilot and calibration samples.

For a fixed evaluation response distribution $\nu(\cdot\mid s)$, define the comparison coefficient
\begin{equation} \label{eq:subopt_cross_pair_comparison_coverage}
    C_{\mathrm{comp}}(\nu; \Pi)
    = \sup_{\pi \in \Pi} \frac{\mathbb{E}_{\substack{s \sim \rho\\ a \sim \nu(\cdot \mid s), a' \sim \pi(\cdot \mid s)}}
        \left[\left([r_{\pi}(s,a) - r^\star(s,a)] - [r_{\pi}(s,a') - r^\star(s,a')]\right)^2\right]}{\mathbb{E}_D\left[\|x_{\pi}(D) - x^\star(D)\|^2\right]}
\end{equation}
with the convention $0/0=0$, where $a$ and $a'$ are conditionally independent given $s$.
This coefficient measures how effectively the preference-data distribution controls reward-error differences between responses drawn from $\nu$ and a candidate policy.
Larger values indicate greater error amplification and hence weaker comparison coverage.
Its role is analogous to that of relative condition numbers in log-linear alignment analyses \citep{zhou2025unified}.

For $\delta \in (0,1)$, define the conformal uncertainty envelope
\begin{equation*}
    \Psi_{\alpha,\gamma}(\delta)
    = \min\left\{1,\,
        \varepsilon_{\alpha,\gamma}^2
        + \colorbox{red!6}{$\displaystyle (1-\varepsilon_{\alpha,\gamma}^2)
        \left(\widetilde{\eta}_{\alpha,\gamma}+\gamma+
        \frac{\log(2/\delta)}{n_{\mathrm{cal}}}\right)$}
    \right\},
\end{equation*}
and the common error term
\begin{equation} \label{eq:subopt_main_error_term}
    \mathfrak{E}_{\alpha,\gamma}(\delta)
    = \underbrace{\mathcal{E}_\Pi}_{\text{policy approximation error}}
    + \underbrace{\colorbox{red!6}{$\displaystyle
        \Psi_{\alpha,\gamma}(\delta)
        $}}_{\text{conformal uncertainty}}
    + \underbrace{\colorbox{blue!6}{$\displaystyle
        \frac{d_\Pi + \log(2/\delta)}{n_{\mathrm{opt}}}
        $}}_{\text{statistical estimation error}}.
\end{equation}
Here, $d_\Pi$ is the complexity parameter in the localized Rademacher condition (\ref{eq:subopt_localized_complexity_condition}).
Sufficient conditions for (\ref{eq:subopt_localized_complexity_condition}) are given in Lemma~\ref{lem:subopt_common_policy_classes}.

\begin{theorem}[Finite-Sample Suboptimality Guarantee] \label{thm:subopt_main_guarantees}
    Suppose that Assumptions~\ref{ass:subopt_sampling} and~\ref{ass:bounded_comparison_scores} and the localized-complexity condition (\ref{eq:subopt_localized_complexity_condition}) hold.
    Assume further that $C_{\mathrm{comp}}(\pi^\star;\Pi) < \infty$.
    Fix $0< \gamma < \alpha < 1$ independently of the calibration sample, and let $\delta \in (0, 1)$.
    Conditional on the pilot sample, with probability at least $1 - \delta$ over the calibration and optimization samples, every comparator policy $\pi^\dagger$ such that $J(\pi)$ and $H(\pi)$ are finite for $\pi \in \{\pi^\star, \widehat\pi_{\mathrm{LCP}}, \pi^\dagger\}$ satisfies
    \begin{equation} \label{eq:subopt_main_regularized}
        \mathrm{SubOpt}_\beta(\widehat{\pi}_{\mathrm{LCP}}, \pi^\dagger)
        \leq \frac{C_{\mathrm{M}} C_{\mathrm{comp}}(\pi^\star; \Pi)}{2\beta} \mathfrak{E}_{\alpha,\gamma}(\delta)
            - \beta\mathbb{E}_{s \sim \rho} \left[D_{\mathrm{KL}}\left(\pi^\dagger(\cdot \mid s) \middle\| \pi^\star(\cdot \mid s)\right)\right],
    \end{equation}
    where $C_{\mathrm{M}} > 0$ depends only on $B$, the model $\mathrm{M}$, the localized-complexity constant $C_{\mathrm{loc}}$, and, for PL feedback, the fixed number of candidates $K$.
    In particular, setting $\pi^\dagger = \pi^\star$ removes the comparator-dependent KL correction.
\end{theorem}

The conformal component of the bound depends on the compatibility tolerance $\varepsilon_{\alpha,\gamma}$, the information retained by the prediction sets, and finite-sample calibration.
Decreasing $\gamma$ lowers $\varepsilon_{\alpha,\gamma}$ but can enlarge the prediction sets, so its effect on the bound need not be monotone.
For fixed confidence and a fixed policy class, the optimization-sample term scales as \colorbox{blue!6}{$d_\Pi/n_{\mathrm{opt}}$} and decreases as more label-free optimization contexts are added.
Corollary~\ref{cor:subopt_approximate_optimization} extends this bound to any approximate policy $\widetilde{\pi}$, including a policy returned by the MM-based PL procedure, by adding a term proportional to $\Delta_{\mathrm{opt}}$, defined as the empirical pessimistic loss at $\widetilde{\pi}$ minus the infimum of that loss over $\Pi$.
The full theorem statement and proof are provided in Appendix~\ref{app:suboptimality_upper_bound}.
Explicit choices of $C_{\mathrm{M}}$ are given in (\ref{eq:model_specific_constant_bt}) for BT feedback and (\ref{eq:model_specific_constant_pl}) for PL feedback; further discussion appears in Appendix~\ref{app:suboptimality_theorem_and_discussion}.

\begin{theorem}[Worst-case Lower Bound for LatentCP-DPO] \label{thm:subopt_joint_worst_case_lower}
    Fix $\alpha \in (0, 1)$, $B, \beta > 0$, and $\gamma \in (0, \alpha/2)$ independently of the calibration sample.
    For every integer $d_\Pi\geq1$ and integer sample sizes satisfying $n_{\mathrm{pil}} \geq 2d_\Pi + 1$, $n_{\mathrm{cal}} \geq 1 / \gamma - 1$, and $n_{\mathrm{opt}} \geq 8 \max\left\{d_\Pi, 1 / \gamma\right\}$, there exist a BT instance and a log-linear policy class satisfying Assumptions~\ref{ass:subopt_sampling} and~\ref{ass:bounded_comparison_scores} and the localized-complexity condition, with $\pi^\star \in \Pi$ and $C_{\mathrm{comp}}(\pi^\star; \Pi) \leq 1$.
    Moreover, there exists a pilot-sample event $\mathcal{H}_{\mathrm{pil}}$ of positive probability such that, conditional on $\mathcal{H}_{\mathrm{pil}}$, with probability at least $1/16$ over the calibration and optimization samples, every exact LatentCP-DPO empirical optimizer satisfies
    \begin{equation} \label{eq:subopt_joint_worst_case_lower}
        \mathrm{SubOpt}_{\beta}(\widehat{\pi}_{\mathrm{LCP}}, \pi^\star)
        \geq c_{B,\beta}\left[
        \colorbox{blue!6}{$\displaystyle
            \frac{d_\Pi}{n_{\mathrm{opt}}}
            $}
        + \colorbox{red!6}{$\displaystyle
            \left(1-2\varepsilon_{\alpha,\gamma}\right)^2 \left(\gamma+\frac{1}{n_{\mathrm{cal}}}\right)
            $}
        \right],
    \end{equation}
    where $c_{B, \beta} > 0$ depends only on $B$ and $\beta$.
\end{theorem}

The construction and proof are provided in Appendix~\ref{app:suboptimality_lower_bound_joint}.
This lower bound matches the upper bound's \colorbox{blue!6}{$d_\Pi/n_{\mathrm{opt}}$} dependence for the exact estimator $\widehat{\pi}_{\mathrm{LCP}}$.
It further shows that, with the calibration sample size and coverage levels fixed, \colorbox{red!6}{conformal bias} can persist as the optimization sample grows; see also Proposition~\ref{prop:persistent_conformal_bias} for a BT example without conditioning on a pilot event.
Together, these results distinguish the estimation error reduced by additional label-free optimization contexts from the residual effects of conformal uncertainty on policy performance.

\section{Experiments}

We first use synthetic experiments with known rewards to examine how the observed policy suboptimality relates to our theoretical predictions. We then evaluate LatentCP-DPO with LLM policies on real preference datasets.

\subsection{ Synthetic Experiments}

\begin{figure}[t]
    \centering
    \includegraphics[width=\textwidth]{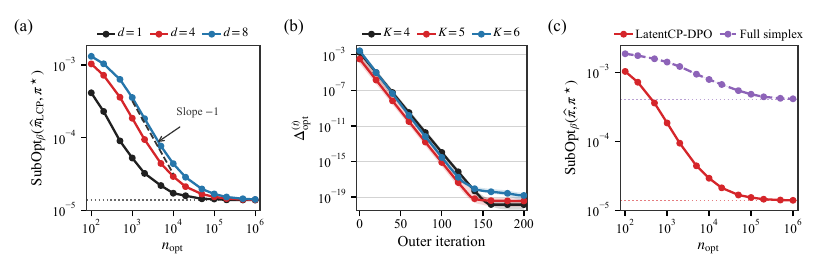}
    \caption{
    Synthetic PL experiments for LatentCP-DPO.
    (a) Mean KL-regularized policy suboptimality $\mathrm{SubOpt}_{\beta}(\widehat{\pi}_{\mathrm{LCP}}, \pi^\star)$ versus $n_{\mathrm{opt}}$ for $d \in \{1, 4, 8\}$ at $K = 4$.
    (b) Median empirical optimization gap $\Delta_{\mathrm{opt}}$ versus outer iteration $t$ for $K \in \{4, 5, 6\}$ at $d = 4$ and $n_{\mathrm{opt}} = 5000$. Shading shows the range across five runs.
    (c) Mean KL-regularized policy suboptimality of LatentCP-DPO and full-simplex pessimism versus $n_{\mathrm{opt}}$ at $K = d = 4$.
    In (a) and (c), each point averages policy suboptimality over $5{,}000$ independent draws of the unlabeled optimization sample, with the pilot and calibration data held fixed.
    }
    \label{fig:synthetic}
\end{figure}

Inspired by the synthetic experiments of \citet{zhu2023principled}, we use a structured PL construction with known rewards to evaluate KL-regularized policy suboptimality directly.
Appendix~\ref{app:exp_synthetic} details the synthetic construction.
With the pilot and calibration data held fixed, the curves in Figure~\ref{fig:synthetic}(a) decline with $n_{\mathrm{opt}}$ and approach a nonzero plateau.
The decline over intermediate sample sizes is consistent with the predicted $n_{\mathrm{opt}}^{-1}$ trend, and the plateau illustrates persistent pessimistic bias.
Figure~\ref{fig:synthetic}(b) tracks the empirical optimization gap $\Delta_{\mathrm{opt}}^{(t)}$ over outer iterations.
At iteration $50$, the empirical objective value attained by the MM and outer updates differs from the global optimum by less than $10^{-8}$ in all $15$ runs.
In Figure~\ref{fig:synthetic}(c), we ablate the conformal compatibility constraint by replacing the induced ambiguity set in (\ref{eq:unit_level_pessimistic_dpo_objective}) with the full simplex over rankings.
Full-simplex pessimism yields larger policy suboptimality at every tested $n_{\mathrm{opt}}$ and approaches a higher plateau.

\subsection{LLM Experiments}
We now evaluate LatentCP-DPO under PL complete-ranking feedback, the setting with a nontrivial inner optimization problem, to assess prediction-set efficiency and pairwise preference accuracy. 

\textbf{Datasets.} 
We use \texttt{berkeley-nest/Nectar}\footnote{\url{https://huggingface.co/datasets/berkeley-nest/Nectar}}~\citep{starling2023}, retaining prompts whose seven candidate responses have distinct ranks.
For each prompt, we randomly permute the seven responses and keep the first $K = 6$ in permuted order, preventing candidate position from revealing the original rank.
Each resulting context admits $6! = 720$ complete rankings.
Using a fixed seed, we create the following disjoint splits: pilot ($3,000$), $\gamma$-calibration ($1,500$), validation ($2,000$), optimization ($8,000$, with $4,000$ used per final policy run), final calibration ($1,500$), and test ($3,000$).

\textbf{Experiment setting and compute resources.}
We use Qwen3-8B for the main experiments and Llama-3.1-8B-Instruct and Llama-2-7B-Chat for the cross-architecture study in Appendix~\ref{app:exp_cross_architecture}.
All policies use LoRA on attention and MLP projections ($r = 32$, $\alpha_{\mathrm{LoRA}} = 64$, dropout $0.05$) and train for one epoch with batch size $64$, AdamW (learning rate $10^{-5}$), a cosine schedule with $5\%$ warmup, gradient clipping at $1.0$, bf16, and gradient checkpointing.
We fix $\beta = 0.1$, $\alpha = 0.30$, and $\gamma = 0.20$. All methods are evaluated across three matched seeds using $24\times$NVIDIA A100 GPUs.

\textbf{Metric.}
For each frozen policy $\pi$, let $p_\pi(y\mid D)$ be the PL likelihood induced by its implicit rewards $\beta\log(\pi/\pi_{\mathrm{ref}})$, and score rankings by $s(D,y) = -\log p_\pi(y\mid D)$.
For 50 nominal coverage levels $\tau \in [0.75, 0.95]$, we compute the split-conformal threshold $\widehat q_\tau$ from that policy's scores on the shared 1,500-context final-calibration split and form
$\mathcal{C}_\tau(D) = \{y \in \mathcal{Y}(D): s(D,y) \leq \widehat{q}_\tau\}$.
On the $3,000$ test contexts, we report average prediction-set size against realized coverage. Each policy calibrates its own threshold at the same requested level $\tau$, so methods are compared at \emph{matched target coverage}, the standard coverage--informativeness trade-off in conformal prediction. 
All $50$-level results are tabulated in Appendix~\ref{app:exp_pl_on_nectar}.

\begin{figure}[t]
    \centering
    \includegraphics[width=\textwidth]{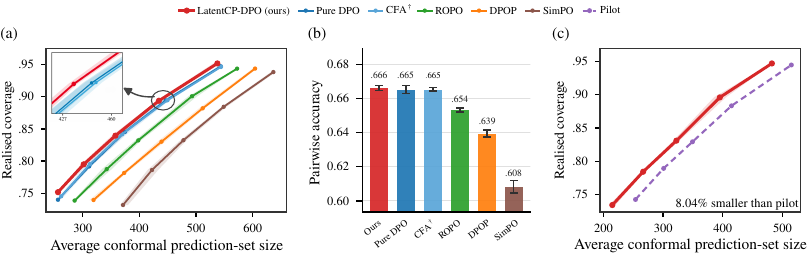}
    \caption{Results on Nectar with Qwen3-8B ($K = 6$).
    (a) Realized coverage versus average prediction-set size at five representative levels among the 50 evaluated nominal coverage levels. Curves show means over three matched policy seeds, and shaded bands show the seed min--max ranges.
    (b) Pairwise accuracy for the same policy runs and test split as in (a). Bars show seed means, and error bars show the seed ranges.
    (c) Descriptive coverage--set-size curves for the pilot policy and a final LatentCP-DPO policy, evaluated on different calibration and test splits from (a) and (b).
    }
    \label{fig:coverage}
\end{figure}

\textbf{Results.}
Figure~\ref{fig:coverage} summarizes the results on Nectar with Qwen3-8B.
At comparable realized coverage, LatentCP-DPO yields smaller average ranking prediction sets than all five baselines at the five displayed levels (Figure~\ref{fig:coverage}(a)).
Across all 50 evaluated levels, the average set-size reductions at common requested coverage level are $2.78\%$, $2.82\%$, and $10.09\%$ relative to Pure DPO~\citep{rafailov2023direct}, the CFA adaptation~\citep{chen2026conformal}, and the ROPO adaptation~\citep{liang2024ropo}, respectively.
Its pairwise accuracy is $66.6\%$, comparable to Pure DPO ($66.5\%$) and CFA ($66.5\%$), and numerically higher than ROPO, DPOP~\citep{pal2024smaug}, and SimPO~\citep{meng2024simpo} (Figure~\ref{fig:coverage}(b)).
The final policy also yields a smaller average set than the pilot at common requested coverage levels (Figure~\ref{fig:coverage}(c)).

Appendix~\ref{app:exp} reports additional results on sensitivity to PL misspecification, performance under BT feedback, downstream response quality, and late-stage overfitting.

\section{Conclusion}
We developed LatentCP-DPO, which incorporates conformally calibrated uncertainty sets for latent relative rewards into a pessimistic preference objective under BT and PL feedback.
Our finite-sample upper bound on KL-regularized policy suboptimality separates policy-class approximation, conformal uncertainty, and optimization-sample estimation error, while a BT worst-case lower bound shows that conformal bias can persist as the label-free optimization sample grows.
Extending this offline framework to online preference optimization remains future work.

\subsection*{AI use statement}
ChatGPT (GPT-5.6 and GPT-6) was used to assist with polishing the manuscript, checking references, and refining mathematical notation.
It also provided suggestions for constructing the BT instance used in the lower-bound proof.
The authors reviewed all AI-assisted material and take responsibility for the final content of this work.



\bibliography{iclr2027_conference}
\bibliographystyle{iclr2027_conference}

\appendix
\section{Additional Related Work} \label{app:additional_related_work}
\paragraph{Additional Background on Provable and Pessimistic Preference Optimization}
\citet{chowdhury2024provably} develop a noise-corrected DPO objective with guarantees under random preference-label flips at a known rate, and \citet{zhou2025unified} analyze the interplay between privacy and adversarial label corruption in offline RLHF and DPO under linear modeling assumptions.
Further pessimistic approaches address general pairwise preference models \citep{gupta2025mitigating} and combine pessimism with corruption modeling and length regularization in offline alignment \citep{chen2025provably}.

\paragraph{Additional background on conformal prediction.}
Conformal prediction constructs distribution-free prediction sets with finite-sample marginal coverage under exchangeability \citep{vovk2005algorithmic, shafer2008tutorial, angelopoulos2023conformal, angelopoulos2024theoretical}.
Applications include regression \citep{lei2018distribution}, classification \citep{ding2023class, dey2023conformal, huang2024conformal}, ranking inference \citep{fermanian2026transductive, liao2026distribution}, and risk control for ranked retrieval \citep{xu2025two}.
Conformal methods have also been developed for off-policy outcome prediction in contextual bandits \citep{taufiq2022conformal} and sequential decision-making settings \citep{zhang2023conformal}, as well as return prediction for infinite-horizon policy evaluation \citep{gan2026conformal}.
Other applications include online prediction with semi-bandit feedback \citep{ge2025variational} and cost-sensitive LLM-agent orchestration under coverage constraints \citep{si2026conformal}.

\section{Validity of Latent Conformal Uncertainty Sets} \label{app:latent_coverage_proof}
\begin{proof}[\bf Proof of Theorem \ref{thm:latent_coverage}]
    Let $\mathcal{A}_\gamma = \mathcal{Y}(D_{n + 1}) \setminus \widehat{\mathcal{C}}_\gamma(D_{n+1})$ and
    \begin{equation*}
        R_\gamma = P_{\mathrm{M}, \boldsymbol{u}_{n+1}^\star}\{\mathcal{A}_\gamma \mid D_{n+1}\} 
        = 1 - p_\gamma(u_{n+1}^\star, D_{n+1}).
    \end{equation*}
    By the definition of TV distance,
    \begin{equation*}
        R_\gamma \leq Q(\mathcal{A}_\gamma \mid D_{n+1}) + \operatorname{TV}\left(Q(\cdot \mid D_{n+1}), P_{\mathrm{M}, \boldsymbol{u}_{n+1}^\star}(\cdot \mid D_{n+1})\right).
    \end{equation*}
    Since $y_i | (D_i, \boldsymbol{u}_i^\star) \sim Q(\cdot\mid D_i)$, by the conditional independence and exchangeability assumptions in the theorem,
    \begin{equation*}
        \mathbb{E}[R_\gamma] \leq \mathbb{E}\left[Q(\mathcal{A}_\gamma \mid D_{n+1})\right] + \Delta 
        = \mathbb{P}\{y_{n+1} \notin \widehat{\mathcal{C}}_\gamma(D_{n+1})\} + \Delta 
        \leq \gamma + \Delta.
    \end{equation*}
    The definition of the uncertainty set, together with the empty-set convention, implies
    \begin{equation*}
        \left\{\boldsymbol{u}_{n+1}^\star \notin \widehat{\mathcal{U}}_\alpha(D_{n+1}; \gamma)\right\}
        \subseteq
        \left\{R_\gamma > \frac{\gamma}{\alpha}\right\}.
    \end{equation*}
    Markov's inequality therefore yields
    \begin{equation*}
        \mathbb{P}\left\{\boldsymbol{u}_{n+1}^\star \notin \widehat{\mathcal{U}}_\alpha(D_{n+1}; \gamma)\right\}
        \leq \mathbb{P}\left\{R_\gamma > \frac{\gamma}{\alpha}\right\}
        \leq \frac{\alpha}{\gamma} \mathbb{E}[R_\gamma]
        \leq \alpha \left(1 + \frac{\Delta}{\gamma}\right).
    \end{equation*}
    Taking the minimum with $1$ gives the stated bound.
\end{proof}

\section{Additional Details for LatentCP-DPO} \label{app:instantiations_for_bt_and_pl_models}

\subsection{Overall procedure.} \label{app:overall_procedure}
The LatentCP-DPO procedure is summarized in Algorithm~\ref{alg:latentcp_dpo}.

\begin{algorithm}[htb]
    \caption{LatentCP-DPO}
    \label{alg:latentcp_dpo}
    \begin{algorithmic}[1]
        \REQUIRE Labeled preference units $\mathcal{D}_{\mathrm{lab}} = \{(D_i, y_i)\}_{i=1}^n$;
        label-free contexts $\{D_j^{\mathrm{opt}}\}_{j=1}^{n_{\mathrm{opt}}}$;
        feedback model $\mathrm{M} \in \{\mathrm{BT}, \mathrm{PL}\}$;
        levels $0 < \gamma < \alpha < 1$.

        \ENSURE LatentCP-DPO policy $\widehat{\pi}_{\mathrm{LCP}}$.

        \STATE Split $\{1, \ldots, n\} = \mathcal{I}_{\mathrm{pil}} \mathbin{\dot\cup} \mathcal{I}_{\mathrm{cal}}$.

        \STATE Fit the pilot policy $\widehat{\pi}_{\mathrm{pil}}$ by solving (\ref{eq:empirical_dpo_estimator}) on the pilot subset.

        \STATE Compute $S_i^{\mathrm{cal}} \gets \widehat{s}(D_i,y_i)$ for all $i \in \mathcal{I}_{\mathrm{cal}}$ using (\ref{eq:nonconformity_score}), and sort them as $S_{(1)}^{\mathrm{cal}} \leq \cdots \leq S_{(n_{\mathrm{cal}})}^{\mathrm{cal}}$.

        \STATE Set $k_\gamma \gets \lceil(n_{\mathrm{cal}} + 1)(1 - \gamma)\rceil$.

        \STATE Compute $\widehat{q}_{1-\gamma}^{\mathrm{cal}} \gets S_{(k_\gamma)}^{\mathrm{cal}}$, with $S_{(n_{\mathrm{cal}}+1)}^{\mathrm{cal}} = +\infty$.

        \FOR{$i=1,\ldots,n_{\mathrm{opt}}$}
        \STATE Construct $\widehat{\mathcal{C}}_{\gamma,i} \gets \left\{y \in \mathcal{Y}(D_i^{\mathrm{opt}}): \widehat{s}(D_i^{\mathrm{opt}}, y) \leq \widehat{q}_{1-\gamma}^{\mathrm{cal}}\right\}$.
        \STATE Construct $\widehat{\mathcal{U}}_{\alpha,i} \gets \left\{\boldsymbol{u} \in \Theta(D_i^{\mathrm{opt}}): p_\gamma(\boldsymbol{u}, D_i^{\mathrm{opt}}) \geq 1 - \varepsilon_{\alpha, \gamma}\right\}$.
        \IF{$\widehat{\mathcal{U}}_{\alpha, i} = \varnothing$}
        \STATE $\widehat{\mathcal U}_{\alpha, i} \gets \Theta(D_i^{\mathrm{opt}})$.
        \ENDIF
        \STATE For $\pi \in \Pi$, define $\mathcal{L}_{\alpha,\gamma}^{\mathrm{pess}}(\pi; D_i^{\mathrm{opt}})$ according to (\ref{eq:unit_level_pessimistic_dpo_objective}).
        \ENDFOR

        \STATE Obtain $\widehat{\pi}_{\mathrm{LCP}} \in \operatorname*{argmax}_{\pi \in \Pi} \frac{1}{n_{\mathrm{opt}}} \sum_{i=1}^{n_{\mathrm{opt}}} \mathcal{L}_{\alpha,\gamma}^{\mathrm{pess}}(\pi; D_i^{\mathrm{opt}})$.

        \RETURN $\widehat{\pi}_{\mathrm{LCP}}$.
    \end{algorithmic}
\end{algorithm}

\subsection{BT Instantiation} \label{app:bt_instantiation}
\paragraph{Derivation of the Latent Uncertainty Set (\ref{eq:bt_latent_uncertainty_set}).}
The BT likelihood induced by $\boldsymbol{u}$ can be written as
\begin{equation*}
    p_{\mathrm{BT}, \boldsymbol{u}}(y \mid D)
    = \sigma\left(u^1 - u^0\right)^y \cdot \sigma\left(u^0 - u^1\right)^{1-y},
    \qquad y \in \{0,1\}.
\end{equation*}
For any $\boldsymbol{u} \in \Theta(D)$, let $q_{\boldsymbol{u}} = \sigma\left(u^1 - u^0\right)$.
The forward $\gamma$-compatibility (\ref{eq:forward_gamma_compatibility}) can be written as
\begin{equation} \label{eq:bt_gamma_compatibility}
    p_\gamma(\boldsymbol{u}, D) 
    = q_{\boldsymbol{u}} \mathbb{I}\{1 \in \widehat{\mathcal{C}}_{\gamma}(D)\} + (1 - q_{\boldsymbol{u}}) \mathbb{I}\{0 \in \widehat{\mathcal{C}}_{\gamma}(D)\},
\end{equation}
where $\mathbb{I}\{\cdot\}$ denotes the indicator function.
Since $\sigma(t) \geq 1 - \frac{\gamma}{\alpha}$ if and only if $t \geq \log\frac{\alpha - \gamma}{\gamma}$, the definition of the latent uncertainty set in (\ref{eq:latent_relative_reward_uncertainty_set}) yields (\ref{eq:bt_latent_uncertainty_set}), using the fact that $p_\gamma(\boldsymbol{u},D) = 1$ when $\widehat{\mathcal{C}}_\gamma(D) = \{0,1\}$ and the empty-set convention when $\widehat{\mathcal{C}}_\gamma(D) = \varnothing$.

\paragraph{Derivation of the Pessimistic Objective (\ref{eq:bt_unit_level_pessimistic_objective}).}
With the notation $z_\pi(D) = r_\pi(s, a^1) - r_\pi(s, a^0)$, we write
\begin{equation*}
    p_{\mathrm{BT}, \pi}(y \mid D)
    = \sigma(z_\pi(D))^{y} \cdot \sigma(-z_\pi(D))^{1 - y}.
\end{equation*}
The expected BT log-likelihood is thus
\begin{equation*}
    \begin{aligned}
        G_{\mathrm{BT}}(\boldsymbol{u}, \pi; D)
        &= \mathbb{E}_{y \sim P_{\mathrm{BT}, \boldsymbol{u}}(\cdot \mid D)}\left[\log p_{\mathrm{BT}, \pi}(y \mid D)\right]\\
        &= q_{\boldsymbol{u}} \log \sigma(z_\pi(D)) + \left(1 - q_{\boldsymbol{u}}\right) \log \sigma(-z_\pi(D))\\
        &= \log \sigma(-z_\pi(D)) + q_{\boldsymbol{u}} z_\pi(D),
    \end{aligned}
\end{equation*}
where the last equality follows from $\log\sigma(z) - \log\sigma(-z) = z$.
Thus, $G_{\mathrm{BT}}(\boldsymbol{u}, \pi; D)$ is affine in $q_{\boldsymbol{u}}$.

Combining (\ref{eq:latent_relative_reward_uncertainty_set}) and (\ref{eq:bt_gamma_compatibility}) with the empty-set convention shows that the closure of the feasible preference-probability set is
\begin{equation*}
    \operatorname{cl}\{q_{\boldsymbol{u}}: \boldsymbol{u} \in \widehat{\mathcal{U}}_\alpha(D; \gamma)\}
    = \begin{cases}
        [1 - \varepsilon_{\alpha, \gamma}, 1], & \widehat{\mathcal{C}}_\gamma(D) = \{1\},\\
        [0, \varepsilon_{\alpha,\gamma}], & \widehat{\mathcal{C}}_\gamma(D) = \{0\},\\
        [0,1], & \widehat{\mathcal{C}}_\gamma(D) = \{0, 1\} \text{ or } \varnothing.
    \end{cases}
\end{equation*}
For fixed $\pi$ and $D$, the affine expression above is continuous on $q \in [0, 1]$, so taking the closure does not change the infimum.
For any $z \in \mathbb{R}$ and $0 \leq \underline{q} \leq \overline{q} \leq 1$,
\begin{equation*}
    \min_{q \in [\underline{q}, \overline{q}]} \left\{\log\sigma(-z) + qz\right\}
    = \log\sigma(-z) + \underline{q}[z]_+ - \overline{q}[-z]_+.
\end{equation*}
Applying this identity to the three intervals above yields (\ref{eq:bt_unit_level_pessimistic_objective}).

\paragraph{Unit-Level Optima of (\ref{eq:bt_unit_level_pessimistic_objective}).}
Fix a context $D$ and regard $z = z_\pi(D)$ as an unrestricted scalar margin.
For the singleton conformal prediction sets, it suffices to consider $\widehat{\mathcal{C}}_\gamma(D) = \{1\}$, since the case $\widehat{\mathcal{C}}\gamma(D) = \{0\}$ follows by symmetry under $z \mapsto -z$.
Define
\begin{equation*}
    \mathcal{L}_{\mathrm{BT}, 1}(z)
    = \log\sigma\left(z\right) - \varepsilon_{\alpha,\gamma} [z]_+
    = \begin{cases}
        \log\sigma\left(z\right) - \varepsilon_{\alpha,\gamma} z, & z > 0,\\
        \log\sigma\left(z\right), & z \leq 0.
    \end{cases}
\end{equation*}
The function $\mathcal{L}_{\mathrm{BT}, 1}$ is continuous and strictly concave on $\mathbb{R}$, but nondifferentiable at $z = 0$.
For $z \neq 0$, its derivative is
\begin{equation*}
    \mathcal{L}'_{\mathrm{BT}, 1}(z)
    = \begin{cases}
        \sigma(-z), & z < 0,\\
        \sigma(-z) - \varepsilon_{\alpha, \gamma}, & z > 0.
    \end{cases}
\end{equation*}
In particular, its one-sided derivatives at the origin are
\begin{equation*}
    \mathcal{L}'_{\mathrm{BT}, 1}(0^-) = \frac{1}{2},
    \qquad
    \mathcal{L}'_{\mathrm{BT}, 1}(0^+) = \frac{1}{2} - \varepsilon_{\alpha, \gamma}.
\end{equation*}
If $\varepsilon_{\alpha, \gamma} < 1/2$, the unique stationary point on $(0,\infty)$ is the unique global maximizer by strict concavity, with
\begin{equation*}
    z^\star
    = \log\frac{1 - \varepsilon_{\alpha, \gamma}}{\varepsilon_{\alpha, \gamma}}
    > 0,
    \qquad
    \sigma(z^\star)
    = 1 - \varepsilon_{\alpha, \gamma}.
\end{equation*}
If $\varepsilon_{\alpha, \gamma} \geq 1/2$, then $\mathcal{L}_{\mathrm{BT}, 1}$ is strictly increasing on $(-\infty, 0)$ and decreasing on $(0, \infty)$. 
Its unique global maximizer is therefore
\begin{equation*}
    z^\star = 0,
    \qquad
    \sigma(z^\star) = \frac{1}{2}.
\end{equation*}

When $\widehat{\mathcal{C}}_\gamma(D) = \{0, 1\}$, or when $\widehat{\mathcal{C}}_\gamma(D) = \varnothing$ and the empty-set convention is invoked, the unit-level objective reduces to $\log \sigma(-|z|)$.
Since $\log \sigma(t)$ is strictly increasing in $t$, 
\begin{equation*}
    \log\sigma(-|z|) \leq \log\sigma(0),
\end{equation*}
with equality if and only if $z = 0$.
Hence, the unique global maximizer is again $z^\star = 0$.

\paragraph{Additional Comparison with Standard DPO.}
For a binary preference label $y \in \{0,1\}$, the standard unit-level DPO log-likelihood under the BT preference model is
\begin{equation*}
    \mathcal{L}^{\mathrm{DPO}}(\pi;D,y)
    = y\log\sigma\bigl(z_\pi(D)\bigr) + (1 - y)\log\sigma\bigl(-z_\pi(D)\bigr).
\end{equation*}
When viewed in isolation as an unconstrained scalar objective, it drives $\sigma(z_\pi(D))$ toward one for $y = 1$ and toward zero for $y = 0$.

When $\widehat{\mathcal{C}}_\gamma(D)$ is a singleton, the pessimistic objective (\ref{eq:bt_unit_level_pessimistic_objective}) can be written as
\begin{equation*}
    \mathcal{L}_{\alpha,\gamma}^{\mathrm{pess}}(\pi; D)
    = \begin{cases}
        \mathcal{L}^{\mathrm{DPO}}(\pi; D, 1) - \varepsilon_{\alpha,\gamma}[z_\pi(D)]_+, & \widehat{\mathcal{C}}_\gamma(D) = \{1\},\\
        \mathcal{L}^{\mathrm{DPO}}(\pi; D, 0) - \varepsilon_{\alpha,\gamma}[-z_\pi(D)]_+, & \widehat{\mathcal{C}}_\gamma(D) = \{0\}.
    \end{cases}
\end{equation*}
For $\widehat{\mathcal{C}}_\gamma(D) = \{1\}$, the one-sided penalty vanishes when $z_\pi(D) < 0$, leaving the standard DPO log-likelihood for $y = 1$ and its margin derivative unchanged.
When $z_\pi(D) > 0$, the penalty reduces this derivative by $\varepsilon_{\alpha, \gamma}$.
Thus, for $\varepsilon_{\alpha,\gamma} < 1/2$, the pessimistic objective retains the supported preference direction while favoring a finite margin determined by the uncertainty tolerance.
The case $\widehat{\mathcal{C}}_\gamma(D) = \{0\}$ is symmetric.
The additional penalty follows exactly from worst-case optimization over the compatible preference distributions.

\subsection{PL Instantiation} \label{app:pl_instantiation}
For any $\boldsymbol{u} \in \Theta(D)$, the induced PL likelihood of a complete ranking $y \in \mathfrak{S}_K$ is
\begin{equation*} \label{eq:pl_parameterization_pl_model}
    p_{\mathrm{PL}, \boldsymbol{u}}(y \mid D)
    = \prod_{k=1}^{K-1} \frac{\exp\left(u^{y_k}\right)}{\sum_{\ell=k}^{K}\exp\left(u^{y_\ell}\right)}.
\end{equation*}
The forward $\gamma$-compatibility under the PL model is
\begin{equation*}
    p_\gamma(\boldsymbol{u}, D)
    = \sum_{y \in \widehat{\mathcal{C}}_\gamma(D)} p_{\mathrm{PL}, \boldsymbol{u}}(y \mid D).
\end{equation*}
This quantity is the total probability mass that the PL distribution induced by $\boldsymbol{u}$ assigns to the conformal prediction set $\widehat{\mathcal{C}}_\gamma(D)$.

Suppose first that $\widehat{\mathcal{C}}_\gamma(D) \neq \varnothing$.
The PL latent uncertainty set is
\begin{equation} \label{eq:latent_uncertainty_set}
    \widehat{\mathcal{U}}_\alpha(D; \gamma)
    = \left\{\boldsymbol{u} \in \Theta(D): p_\gamma(\boldsymbol{u}, D) \geq 1 - \varepsilon_{\alpha, \gamma}\right\}.
\end{equation}
This set is nonempty because a PL distribution can assign probability arbitrarily close to one to any prescribed ranking by taking sufficiently large successive score gaps.
For $K > 2$, (\ref{eq:latent_uncertainty_set}) generally defines an implicit set in the $(K-1)$-dimensional relative-reward space.
Using the enumeration $\mathfrak{S}_K = \{y^{(1)}, \ldots, y^{(N_K)}\}$, define $q'_m(\boldsymbol{u}) = p_{\mathrm{PL}, \boldsymbol{u}}(y^{(m)} \mid D)$ and recall that $c_m = \mathbb{I}\{y^{(m)} \in \widehat{\mathcal{C}}_\gamma(D)\}$.
Let $\boldsymbol{q}'(\boldsymbol{u})$ collect the probabilities $q'_m(\boldsymbol{u})$ in the same enumeration order.
For $\boldsymbol{u} \in \Theta(D)$, membership in $\widehat{\mathcal{U}}_\alpha(D; \gamma)$ is equivalent to $\boldsymbol{c}^\top \boldsymbol{q}'(\boldsymbol{u}) \geq 1 - \varepsilon_{\alpha, \gamma}$.

For any $\pi \in \Pi$ and $\boldsymbol{u} \in \Theta(D)$, define
\begin{equation*}
    G_{\mathrm{PL}}(\boldsymbol{u}, \pi; D)
    = \mathbb{E}_{y \sim P_{\mathrm{PL}, \boldsymbol{u}}(\cdot \mid D)} \left[\log p_{\mathrm{PL}, \pi}(y \mid D)\right]
    = \sum_{y \in \mathfrak{S}_K} p_{\mathrm{PL}, \boldsymbol{u}}(y \mid D) \log p_{\mathrm{PL}, \pi}(y \mid D).
\end{equation*}
Recalling that $h_m(\pi) = \log p_{\mathrm{PL}, \pi}(y^{(m)} \mid D)$, we obtain
\begin{equation*}
    G_{\mathrm{PL}}(\boldsymbol{u}, \pi; D) = \boldsymbol{q}'(\boldsymbol{u})^\top \boldsymbol{h}(\pi).
\end{equation*}
Consequently, (\ref{eq:unit_level_pessimistic_dpo_objective}) can be rewritten as
\begin{equation} \label{eq:pl_unit_level_pessimistic_objective_temp}
    \mathcal{L}_{\alpha, \gamma}^{\mathrm{pess}}(\pi; D) = \inf_{\boldsymbol{u} \in \Theta(D)} \boldsymbol{q}'(\boldsymbol{u})^\top \boldsymbol{h}(\pi)
    \quad
    \text{s.t.}
    \quad
    \boldsymbol{c}^\top \boldsymbol{q}'(\boldsymbol{u}) \geq 1 - \varepsilon_{\alpha, \gamma}
\end{equation}
Every $\boldsymbol{u} \in \Theta(D)$ admits the unique representation $\boldsymbol{u} = B_K \boldsymbol{v}$, where $\boldsymbol{v} = B_K^\top \boldsymbol{u} \in \mathbb{R}^{K-1}$.
Since $\boldsymbol{q}'(B_K\boldsymbol{v}) = \boldsymbol{q}(\boldsymbol{v})$, this change of coordinates transforms (\ref{eq:pl_unit_level_pessimistic_objective_temp}) into (\ref{eq:pl_pessimistic_objective_computational_formulation}).

When $\widehat{\mathcal{C}}_\gamma(D) = \varnothing$, we have $p_\gamma(\boldsymbol{u}, D) = 0$ for every $\boldsymbol{u} \in \Theta(D)$.
The empty-set convention therefore sets $\widehat{\mathcal{U}}_\alpha(D; \gamma) = \Theta(D)$, and the compatibility constraint is omitted from both formulations.

\subsection{Approximating the Pessimistic PL Objective} \label{app:pl_approximation_of_the_pessimistic_pl_objective}
The exact reformulation in (\ref{eq:pl_pessimistic_objective_computational_formulation}) makes explicit the computational challenge of the PL instantiation.
For fixed context $D$ and policy $\pi \in \Pi$, evaluating the exact pessimistic objective amounts to solving a generally nonconvex constrained optimization problem over the latent-score coordinates $\boldsymbol{v} \in \mathbb{R}^{K-1}$.
We therefore develop a feasibility-preserving minorization--maximization (MM) procedure for obtaining an approximate solution to this inner problem.

Recall that $N_K = K!$.
Write $h_m = h_m(\pi) = \log p_{\mathrm{PL},\pi}(y^{(m)} \mid D)$.
Since all ranking log-likelihoods are finite by the definition of $\Pi$ and $K \geq 2$, we have $h_m < 0$ for every $m = 1, \ldots, N_K$.
Define the index set
\begin{equation*}
    \mathcal{I}_{\mathcal{C}} 
    = \left\{m: c_m = 1\right\}
    = \left\{m: y^{(m)} \in \widehat{\mathcal{C}}_{\gamma}(D)\right\},
\end{equation*}
and let
\begin{equation*}
    Q_{\mathcal{C}}(\boldsymbol{v}) 
    = \boldsymbol{c}^\top \boldsymbol{q}(\boldsymbol{v})
    = \sum_{m \in \mathcal{I}_{\mathcal{C}}} q_m(\boldsymbol{v}).
\end{equation*}
Thus, $Q_{\mathcal{C}}(\boldsymbol{v}) = p_{\gamma}(B_K\boldsymbol{v},D)$ is the forward compatibility expressed in the coordinates $\boldsymbol{v}$.
Under the conservative empty-set convention, define the feasible region by
\begin{equation*}
    \mathcal{V} = \begin{cases}
        \{\boldsymbol{v} \in \mathbb{R}^{K-1}: Q_{\mathcal{C}}(\boldsymbol{v}) \geq 1 - \varepsilon_{\alpha, \gamma}\}, & \mathcal{I}_{\mathcal{C}} \neq \varnothing,\\
        \mathbb{R}^{K-1}, & \mathcal{I}_{\mathcal{C}} = \varnothing.
    \end{cases}
\end{equation*}
Define
\begin{equation*}
    F(\boldsymbol{v}) = \sum_{m = 1}^{N_K} q_m(\boldsymbol{v}) h_m < 0,
    \qquad
    A(\boldsymbol{v}) = \log (-F(\boldsymbol{v})).
\end{equation*}
Since $t \mapsto \log(-t)$ is strictly decreasing on $(-\infty,0)$,
\begin{equation*}
    \mathcal{L}_{\alpha, \gamma}^{\mathrm{pess}}(\pi; D) 
    = \inf_{\boldsymbol{v} \in \mathcal{V}} F(\boldsymbol{v})
    = -\exp\left(\sup_{\boldsymbol{v} \in \mathcal{V}} A(\boldsymbol{v})\right).
\end{equation*}
Thus, minimizing $F$ over $\mathcal{V}$ is equivalent to maximizing $A$ over the same feasible region.
We apply MM to this equivalent maximization problem to obtain a feasible approximate solution.

\paragraph{Concavity of the PL Log-Likelihood.}
Under the parameterization $\boldsymbol{u} = B_K\boldsymbol{v}$, the PL log-likelihood of a ranking $y^{(m)} = (y_1^{(m)},\ldots,y_K^{(m)})$ is
\begin{equation*}
    \log q_{m}(\boldsymbol{v}) 
    = \sum_{k=1}^{K-1} \left[(B_K\boldsymbol{v})_{y_k^{(m)}} - \log\sum_{\ell=k}^{K} \exp\left((B_K\boldsymbol{v})_{y_\ell^{(m)}}\right)\right].
\end{equation*}
Each summand is a linear function minus a log-sum-exp of linear functions.
Therefore, $\log q_m(\boldsymbol{v})$ is smooth and concave in $\boldsymbol{v}$.
This property underlies the construction of the MM surrogate objective and constraint below.

\paragraph{Minorization Step.}
At a feasible iterate $\boldsymbol{v}^{(t)} \in \mathcal{V}$, define
\begin{equation} \label{eq:pl_mm_objective_weights_eta}
    \eta_m^{(t)} = \frac{q_m(\boldsymbol{v}^{(t)}) h_m}{\sum_{m' = 1}^{N_K} q_{m'}(\boldsymbol{v}^{(t)}) h_{m'}},
    \qquad 
    m = 1, \ldots, N_K.
\end{equation}
These weights satisfy $\eta_m^{(t)} > 0$ and $\sum_{m=1}^{N_K} \eta_m^{(t)} = 1$.
Jensen's inequality then yields
\begin{equation*}
    A(\boldsymbol{v}) 
    = \log \left(\sum_{m = 1}^{N_K} \eta_m^{(t)} \frac{-h_m q_m(\boldsymbol{v})}{\eta_m^{(t)}}\right)
    \geq \sum_{m = 1}^{N_K} \eta_m^{(t)} \left[\log(-h_m) + \log q_{m}(\boldsymbol{v}) - \log \eta_m^{(t)}\right].
\end{equation*}
Denote the right-hand side by $M(\boldsymbol{v}, \boldsymbol{v}^{(t)})$.
Because each $\log q_m(\boldsymbol{v})$ is concave, $M(\boldsymbol{v}, \boldsymbol{v}^{(t)})$ is concave in $\boldsymbol{v}$.
The lower bound holds for all $\boldsymbol{v} \in \mathbb{R}^{K-1}$ and is tight at the current iterate:
\begin{equation*}
    M(\boldsymbol{v}^{(t)}, \boldsymbol{v}^{(t)}) = A(\boldsymbol{v}^{(t)})
\end{equation*}
Thus, $M(\boldsymbol{v},\boldsymbol{v}^{(t)})$ is a concave minorizer of $A$.

We next construct a concave minorizer of $\log Q_{\mathcal{C}}(\boldsymbol{v})$.
Consider the nontrivial case $0 < |\mathcal{I}_{\mathcal{C}}| < N_K$.
At the current feasible iterate $\boldsymbol{v}^{(t)}$, define
\begin{equation} \label{eq:pl_mm_feasible_set_compatibility_weights_omega}
    \omega_m^{(t)} = \frac{q_m(\boldsymbol{v}^{(t)})}{Q_{\mathcal{C}}(\boldsymbol{v}^{(t)})},
    \qquad 
    m \in \mathcal{I}_{\mathcal{C}}.
\end{equation}
These weights satisfy $\omega_m^{(t)} > 0$ for every $m \in \mathcal{I}_{\mathcal{C}}$ and $\sum_{m \in \mathcal{I}_{\mathcal{C}}} \omega_m^{(t)} = 1$.
A second application of Jensen's inequality yields
\begin{equation*}
    \log Q_{\mathcal{C}}(\boldsymbol{v})
    = \log \left(\sum_{m \in \mathcal{I}_{\mathcal{C}}} \omega_m^{(t)} \frac{q_m(\boldsymbol{v})}{\omega_m^{(t)}}\right)
    \geq \sum_{m \in \mathcal{I}_{\mathcal{C}}} \omega_m^{(t)} \left[\log q_m(\boldsymbol{v}) - \log \omega_m^{(t)}\right].
\end{equation*}
Denote the right-hand side by $\psi(\boldsymbol{v}, \boldsymbol{v}^{(t)})$.
Since each $\log q_m(\boldsymbol{v})$ is concave, $\psi(\boldsymbol{v}, \boldsymbol{v}^{(t)})$ is concave in $\boldsymbol{v}$, so its superlevel sets are convex.
The lower bound holds for all $\boldsymbol{v} \in \mathbb{R}^{K-1}$ and is tight at the current iterate $\boldsymbol{v}^{(t)}$.
Consequently,
\begin{equation*}
    \psi(\boldsymbol{v}, \boldsymbol{v}^{(t)}) \geq \log(1 - \varepsilon_{\alpha, \gamma})
    \quad \Rightarrow \quad
    Q_{\mathcal{C}}(\boldsymbol{v}) \geq 1 - \varepsilon_{\alpha, \gamma}.
\end{equation*}
Moreover, feasibility of $\boldsymbol{v}^{(t)}$ and tightness of the minorizer imply
\begin{equation*}
    \psi(\boldsymbol{v}^{(t)}, \boldsymbol{v}^{(t)})
    = \log Q_{\mathcal{C}}(\boldsymbol{v}^{(t)})
    \geq \log (1 - \varepsilon_{\alpha, \gamma}).
\end{equation*}
Thus, the surrogate feasible set is convex, contains the current iterate, and is contained in the original feasible region.

If $\mathcal{I}_{\mathcal{C}} = \varnothing$, the compatibility constraint is omitted under the empty-set convention.
If $\mathcal{I}_{\mathcal{C}} = \{1, \ldots, N_K\}$, then $Q_{\mathcal{C}}(\boldsymbol{v}) = 1$ for every $\boldsymbol{v}$, so the constraint is automatically satisfied and may again be omitted.

\paragraph{Maximization Step.}
In the nontrivial constrained case, holding the objective and compatibility weights fixed at iteration $t$, we update the latent-score coordinates by maximizing the concave objective minorizer over the surrogate feasible set.
Specifically,
\begin{equation*}
    \boldsymbol{v}^{(t+1)} \in \operatorname*{argmax}_{\boldsymbol{v} \in \mathbb{R}^{K-1}}\ M(\boldsymbol{v}, \boldsymbol{v}^{(t)})
    \quad
    \text{s.t.}
    \quad
    \psi(\boldsymbol{v}, \boldsymbol{v}^{(t)}) \geq \log(1 - \varepsilon_{\alpha, \gamma}).
\end{equation*}
Dropping the terms in $M(\boldsymbol{v}, \boldsymbol{v}^{(t)})$ that are constant with respect to $\boldsymbol{v}$ and expanding the surrogate constraint gives the equivalent update
\begin{equation} \label{eq:pl_mm_iterate_v_t_plus_one}
    \begin{aligned}
        \boldsymbol{v}^{(t+1)} \in \operatorname*{argmax}_{\boldsymbol{v} \in \mathbb{R}^{K-1}}&\ \sum_{m = 1}^{N_K} \eta_m^{(t)} \log q_{m}(\boldsymbol{v})\\
        \text{s.t.}&\ \sum_{m \in \mathcal{I}_{\mathcal{C}}} \omega_m^{(t)} \log q_m(\boldsymbol{v}) \geq \log(1 - \varepsilon_{\alpha, \gamma}) + \sum_{m \in \mathcal{I}_{\mathcal{C}}} \omega_m^{(t)} \log \omega_m^{(t)}.
    \end{aligned}
\end{equation}
This is a convex optimization problem: the objective is concave in $\boldsymbol{v}$, and the feasible set is convex because it is a superlevel set of a concave function.

\paragraph{Feasible Initialization.}
The MM updates preserve feasibility but require a feasible starting point.
The following lemma provides an explicit construction, with the proof provided in Appendix~\ref{app:pl_mm_proofs}.

\begin{lemma} \label{lem:pl_mm_feasible_initialization}
    Suppose that $\mathcal{I}_{\mathcal{C}} \neq \varnothing$. 
    Choose any ranking $y^{(m_0)} \in \widehat{\mathcal{C}}_\gamma(D)$ and define $\boldsymbol{u}^{(0)} \in \mathbb{R}^K$ by setting
    \begin{equation*}
        u^{(0), y^{(m_0)}_k} = \lambda \left(\frac{K+1}{2} - k\right),
        \qquad
        k = 1, \ldots, K,
    \end{equation*}
    where
    \begin{equation*}
        \lambda \geq \left[\log \frac{K - 1}{(1 - \varepsilon_{\alpha, \gamma})^{-1/(K-1)} - 1}\right]_{+}.
    \end{equation*}
    Then $\boldsymbol{u}^{(0)} \in \boldsymbol{1}_K^\perp$ and $\boldsymbol{v}^{(0)} = B_K^\top \boldsymbol{u}^{(0)}$ satisfies $Q_{\mathcal{C}}(\boldsymbol{v}^{(0)}) \geq 1 - \varepsilon_{\alpha, \gamma}$.
    Hence, $\boldsymbol{v}^{(0)} \in \mathcal{V}$.
\end{lemma}

When $\mathcal{I}_{\mathcal{C}} = \varnothing$, the compatibility constraint is omitted under the conservative empty-set convention, and one may simply initialize $\boldsymbol{v}^{(0)} = \boldsymbol{0}$.

\begin{algorithm}[t]
    \caption{Feasibility-Preserving MM Algorithm for the Pessimistic PL Inner Problem}
    \label{alg:pl_mm_for_pessimistic_pl_inner_problem}
    \begin{algorithmic}[1]
        \REQUIRE Fixed context $D$ and policy $\pi$;
        conformal index set $\mathcal{I}_{\mathcal{C}}$;
        feasible initialization $\boldsymbol{v}^{(0)} \in \mathcal{V}$;
        maximum number of iterations $T_{\mathrm{MM}} \geq 1$;
        tolerance $\delta_{\mathrm{MM}} > 0$.

        \ENSURE A feasible approximate solution $\widetilde{\boldsymbol{v}}$ to the inner problem in (\ref{eq:pl_pessimistic_objective_computational_formulation}).

        \STATE Set $\widetilde{\boldsymbol{v}} \gets \boldsymbol{v}^{(0)}$.

        \FOR{$t = 0, \ldots , T_{\mathrm{MM}} - 1$}
        \STATE Compute the objective weights $\{\eta_m^{(t)}\}_{m=1}^{N_K}$ using (\ref{eq:pl_mm_objective_weights_eta}).
        \IF{$0 < |\mathcal{I}_{\mathcal{C}}| < N_K$}
        \STATE Compute the compatibility weights $\{\omega_m^{(t)}\}_{m \in \mathcal{I}_{\mathcal{C}}}$ using (\ref{eq:pl_mm_feasible_set_compatibility_weights_omega}).
        \STATE Obtain $\boldsymbol{v}^{(t+1)}$ by solving the convex subproblem in (\ref{eq:pl_mm_iterate_v_t_plus_one}).
        \ELSE
        \STATE Obtain $\boldsymbol{v}^{(t+1)}$ by solving the convex subproblem in (\ref{eq:pl_mm_iterate_v_t_plus_one}) with the compatibility constraint omitted.
        \ENDIF

        \STATE Set $\widetilde{\boldsymbol{v}} \gets \boldsymbol{v}^{(t+1)}$.

        \IF{$A(\boldsymbol{v}^{(t+1)}) - A(\boldsymbol{v}^{(t)}) \leq \delta_{\mathrm{MM}}$}
        \STATE \textbf{break}.
        \ENDIF
        \ENDFOR

        \RETURN $\widetilde{\boldsymbol{v}}$.
    \end{algorithmic}
\end{algorithm}

The preceding steps are summarized in Algorithm~\ref{alg:pl_mm_for_pessimistic_pl_inner_problem}.
Let $\widetilde{\boldsymbol{v}}$ denote the feasible approximate solution returned by the algorithm. 
The corresponding MM approximation satisfies $\widetilde{\mathcal{L}}_{\alpha, \gamma}^{\mathrm{pess}}(\pi; D) = \boldsymbol{q}(\widetilde{\boldsymbol{v}})^\top \boldsymbol{h}(\pi)$.

\paragraph{Theoretical Properties.}
We next establish the main properties of the proposed MM algorithm, with the proof provided in Appendix~\ref{app:pl_mm_proofs}.

\begin{theorem}[Feasibility, monotonicity, and stationary accumulation points] \label{thm:pl_mm_properties}
    Fix a context $D$ and a policy $\pi$. 
    Suppose that $\boldsymbol{v}^{(0)} \in \mathcal{V}$ and that every MM subproblem is solved exactly. 
    Then
    \begin{equation*}
        \boldsymbol{v}^{(t)} \in \mathcal{V} 
        \qquad
        \text{for every } t \geq 0,
    \end{equation*}
    and
    \begin{equation*}
        A(\boldsymbol{v}^{(t+1)}) \geq A(\boldsymbol{v}^{(t)}),
        \qquad
        F(\boldsymbol{v}^{(t+1)}) \leq F(\boldsymbol{v}^{(t)}).
    \end{equation*}
    Moreover, both $\{A(\boldsymbol{v}^{(t)})\}_{t \geq 0}$ and $\{F(\boldsymbol{v}^{(t)})\}_{t \geq 0}$ converge to finite limits.

    Suppose, in addition, that $\{\boldsymbol{v}^{(t)}\}_{t \geq 0}$ is bounded, and let $\boldsymbol{v}^{\star}$ be any of its accumulation points.
    In the nontrivial constrained case $0 < |\mathcal{I}_{\mathcal{C}}| < N_K$, assume that the Mangasarian--Fromovitz constraint qualification holds at $\boldsymbol{v}^{\star}$; equivalently,
    \begin{equation*}
        Q_{\mathcal{C}}(\boldsymbol{v}^{\star}) = 1 - \varepsilon_{\alpha, \gamma}
        \quad \Rightarrow \quad
        \nabla Q_{\mathcal{C}}(\boldsymbol{v}^{\star}) \neq \boldsymbol{0}.
    \end{equation*}
    Then $\boldsymbol{v}^{\star}$ satisfies the KKT conditions of the transformed inner maximization problem $\sup_{\boldsymbol{v} \in \mathcal{V}} A(\boldsymbol{v})$ and, equivalently, of the original inner minimization problem in (\ref{eq:pl_pessimistic_objective_computational_formulation}).
\end{theorem}

The theorem establishes feasibility and monotonic improvement of the inner objective, together with stationarity guarantees under the stated assumptions.

\paragraph{Computational Complexity.}
Consider a fixed context $D$ and policy $\pi$, excluding the cost of evaluating the policy on the $K$ candidates.
For a given latent coordinate $\boldsymbol{v}$, computing $B_K \boldsymbol{v}$ costs $O(K^2)$.
Given $B_K \boldsymbol{v}$, the PL probability of one ranking can be evaluated in $O(K)$ time using backward cumulative sums, so evaluating all $N_K = K!$ ranking probabilities costs $O(N_K K)$.
Once these probabilities are available, updating the weights $\{\eta_m^{(t)}\}_{m=1}^{N_K}$ and $\{\omega_m^{(t)}\}_{m \in \mathcal I_{\mathcal{C}}}$ costs $O(N_K)$.

Each MM iteration additionally solves a convex problem in $K - 1$ variables with at most one nontrivial scalar inequality constraint.
Let $C_{\mathrm{cvx}}(K-1, N_K)$ denote the cost of solving this subproblem to the prescribed tolerance, including evaluations of its surrogate objective and constraint.
Up to one-time preprocessing for fixed $(D, \pi)$, an MM run of at most $T_{\mathrm{MM}}$ iterations therefore has time complexity
\begin{equation*}
    O\left(T_{\mathrm{MM}}\left[N_K K + C_{\mathrm{cvx}}(K-1, N_K)\right]\right).
\end{equation*}
Since $N_K = K!$, explicit enumeration induces factorial dependence on $K$.
The proposed implementation is therefore intended for settings with a small-to-moderate number of candidates.

\paragraph{A Simplex-Relaxation Bound.}
A feasible MM output upper-bounds the exact inner infimum. 
To obtain a complementary lower bound, we relax the PL family of ranking distributions to the full probability simplex while retaining the compatibility constraint. 
This yields the following computable, generally conservative bound.

\begin{proposition}[Simplex-Relaxation Bound] \label{prop:pl_inner_value_bound}
    Fix a context $D$ and a policy $\pi$, and define
    \begin{equation*}
        L_{\mathrm{relax}}(\pi; D) = \begin{cases}
            (1 - \varepsilon_{\alpha,\gamma}) \displaystyle \min_{m \in \mathcal{I}_{\mathcal{C}}} h_m(\pi) + \varepsilon_{\alpha,\gamma} \displaystyle\min_{1 \leq m \leq N_K} h_m(\pi), & \mathcal{I}_{\mathcal{C}} \neq \varnothing, \\
            \displaystyle\min_{1 \leq m \leq N_K} h_m(\pi), & \mathcal{I}_{\mathcal{C}} = \varnothing.
        \end{cases}
    \end{equation*}
    For any feasible approximate inner solution $\widetilde{\boldsymbol{v}} \in \mathcal{V}$, let $\widetilde{\mathcal{L}}_{\alpha, \gamma}^{\mathrm{pess}}(\pi; D) = \boldsymbol{q}(\widetilde{\boldsymbol{v}})^\top\boldsymbol{h}(\pi)$.
    Then
    \begin{equation} \label{eq:pl_inner_value_error_bound}
        0 \leq
        \widetilde{\mathcal L}_{\alpha,\gamma}^{\mathrm{pess}}(\pi;D) - \mathcal L_{\alpha,\gamma}^{\mathrm{pess}}(\pi;D)
        \leq
        \widetilde{\mathcal L}_{\alpha,\gamma}^{\mathrm{pess}}(\pi;D) - L_{\mathrm{relax}}(\pi;D).
    \end{equation}
\end{proposition}

The right-hand side of (\ref{eq:pl_inner_value_error_bound}) gives a computable upper bound on the inner value error. 
This bound can be conservative because the relaxation replaces the $(K-1)$-dimensional PL family by the full $(K!-1)$-dimensional simplex of ranking distributions.
The proof is provided in Appendix~\ref{app:pl_mm_proofs}.

\subsection{Proofs for the PL MM Algorithm} \label{app:pl_mm_proofs}
We provide the proofs of Lemma~\ref{lem:pl_mm_feasible_initialization}, Theorem~\ref{thm:pl_mm_properties}, and Proposition~\ref{prop:pl_inner_value_bound} here.

\begin{proof}[\bf Proof of Lemma~\ref{lem:pl_mm_feasible_initialization}]
    By construction,
    \begin{equation*}
        \boldsymbol{1}_K^\top\boldsymbol{u}^{(0)}
        = \sum_{k=1}^{K} u^{(0), y^{(m_0)}_k} 
        = \lambda \sum_{k=1}^{K} \left(\frac{K+1}{2} - k\right) = 0.
    \end{equation*}
    Thus, $\boldsymbol{u}^{(0)} \in \boldsymbol{1}_K^\perp$.
    Since the columns of $B_K$ form an orthonormal basis of $\boldsymbol{1}_K^\perp$, setting $\boldsymbol{v}^{(0)} = B_K^\top\boldsymbol{u}^{(0)}$ gives $B_K\boldsymbol{v}^{(0)} = B_KB_K^\top\boldsymbol{u}^{(0)} = \boldsymbol{u}^{(0)}$.

    We next lower-bound the PL probability of the selected ranking $y^{(m_0)}$. 
    For every $k = 1, \ldots, K-1$, its stage-$k$ selection probability satisfies
    \begin{equation*}
        \frac{\exp\left(u^{(0), y^{(m_0)}_k}\right)}{\sum_{\ell=k}^{K}\exp\left(u^{(0), y^{(m_0)}_\ell}\right)}
        = \frac{1}{1 + \sum_{\ell=k+1}^{K}\exp\left(\lambda \left(k - \ell\right)\right)}
        \geq \frac{1}{1 + (K - 1) e^{-\lambda}},
    \end{equation*}
    because $\lambda \geq 0$ and each of the $K-k \leq K-1$ terms in the sum is at most $e^{-\lambda}$.
    Multiplying these bounds over the $K - 1$ nontrivial stages yields
    \begin{equation*}
        q_{m_0}(\boldsymbol{v}^{(0)}) 
        = \prod_{k=1}^{K-1} \frac{\exp\left(u^{(0), y^{(m_0)}_k}\right)}{\sum_{\ell=k}^{K}\exp\left(u^{(0), y^{(m_0)}_\ell}\right)}
        \geq \left(1 + (K - 1) e^{-\lambda}\right)^{-(K-1)}
        \geq 1 - \varepsilon_{\alpha,\gamma}.
    \end{equation*}
    The last inequality follows from the prescribed lower bound on $\lambda$.

    Finally, since $y^{(m_0)} \in \widehat{\mathcal{C}}_\gamma(D)$, we have $m_0 \in \mathcal{I}_{\mathcal{C}}$ and hence
    \begin{equation*}
        Q_{\mathcal{C}}(\boldsymbol{v}^{(0)})
        = \sum_{m \in \mathcal{I}_{\mathcal{C}}} q_m(\boldsymbol{v}^{(0)})
        \geq q_{m_0}(\boldsymbol{v}^{(0)})
        \geq 1 - \varepsilon_{\alpha, \gamma}.
    \end{equation*}
    Therefore, $\boldsymbol{v}^{(0)} \in \mathcal{V}$, completing the proof.
\end{proof}

\begin{proof}[\bf Proof of Theorem~\ref{thm:pl_mm_properties}]
    First consider the nontrivial constrained case $0 < |\mathcal{I}_{\mathcal{C}}|< N_K$.
    Whenever $\boldsymbol{v}^{(t)}\in\mathcal{V}$, the tightness property of $\psi$ gives
    \begin{equation*}
        \psi(\boldsymbol{v}^{(t)}, \boldsymbol{v}^{(t)}) = \log Q_{\mathcal{C}}(\boldsymbol{v}^{(t)}) \geq \log(1 - \varepsilon_{\alpha, \gamma}).
    \end{equation*}
    Thus, the current iterate is feasible for the surrogate subproblem.
    Moreover, $\psi(\boldsymbol{v},\boldsymbol{v}^{(t)}) \leq \log Q_{\mathcal{C}}(\boldsymbol{v})$, so every surrogate-feasible point belongs to $\mathcal{V}$.
    Starting from $\boldsymbol{v}^{(0)} \in \mathcal{V}$, feasibility of all iterates follows by induction.
    When the compatibility constraint is omitted, $\mathcal{V} = \mathbb{R}^{K-1}$ and feasibility is immediate.

    By exact maximization and the minorization and tightness properties of $M$,
    \begin{equation*}
        A(\boldsymbol{v}^{(t+1)})
        \geq M(\boldsymbol{v}^{(t+1)}, \boldsymbol{v}^{(t)})
        \geq M(\boldsymbol{v}^{(t)}, \boldsymbol{v}^{(t)})
        = A(\boldsymbol{v}^{(t)}).
    \end{equation*}
    Since $F(\boldsymbol{v}) = -\exp(A(\boldsymbol{v}))$, this implies $F(\boldsymbol{v}^{(t+1)}) \leq F(\boldsymbol{v}^{(t)})$.
    Furthermore, $\boldsymbol{q}(\boldsymbol{v})$ is a probability vector, so
    \begin{equation*}
        \min_m h_m 
        \leq F(\boldsymbol{v}) 
        = \boldsymbol{q}(\boldsymbol{v})^\top \boldsymbol{h}
        \leq \max_m h_m < 0.
    \end{equation*}
    Consequently, $\{F(\boldsymbol{v}^{(t)})\}_{t \geq 0}$ converges to a finite negative limit.
    By continuity of $t \mapsto \log(-t)$ on $(-\infty, 0)$, $\{A(\boldsymbol{v}^{(t)})\}_{t\geq0}$ also converges to a finite limit.

    It remains to establish stationarity under the additional assumptions of the theorem.
    Direct calculation gives
    \begin{equation*}
        \nabla_1 M(\boldsymbol{v}, \boldsymbol{v})
        = \nabla A(\boldsymbol{v}),
    \end{equation*}
    and, in the nontrivial constrained case,
    \begin{equation*}
        \nabla_1 \psi(\boldsymbol{v}, \boldsymbol{v})
        = \nabla \log Q_{\mathcal{C}}(\boldsymbol{v}),
    \end{equation*}
    where $\nabla_1$ denotes differentiation with respect to the first argument.
    By boundedness, the iterates lie in a compact convex set $\mathcal{K} \subseteq \mathbb{R}^{K-1}$.
    Let $\boldsymbol{p}^{(1)}(\boldsymbol{v}) \in \mathbb{R}^{K}$ denote the first-stage PL choice-probability vector, with components
    \begin{equation*}
        p_\ell^{(1)}(\boldsymbol{v})
        = \frac{\exp((B_K\boldsymbol{v})_\ell)}{\sum_{\ell'=1}^{K}\exp((B_K\boldsymbol{v})_{\ell'})},
        \qquad 
        \ell = 1, \ldots, K.
    \end{equation*}
    For every ranking, the first-stage contribution to $- \nabla^2 \log q_m(\boldsymbol{v})$ is
    \begin{equation*}
        B_K ^\top\left[\operatorname{Diag}\bigl(\boldsymbol{p}^{(1)}(\boldsymbol{v})\bigr)- \boldsymbol{p}^{(1)}(\boldsymbol{v}) \boldsymbol{p}^{(1)}(\boldsymbol{v})^\top\right] B_K
        \succ 0.
    \end{equation*}
    Indeed, $\boldsymbol{p}^{(1)}(\boldsymbol{v})$ has full support, and the columns of $B_K$ span $\boldsymbol{1}_K^\perp$.
    The remaining stages contribute positive semidefinite matrices.
    Continuity, compactness, and $\sum_{m=1}^{N_K} \eta_m^{(t)} = 1$ therefore imply that the functions $M(\cdot, \boldsymbol{v}^{(t)})$ are uniformly $\mu_{\mathrm{MM}}$-strongly concave on $\mathcal{K}$ for some $\mu_{\mathrm{MM}}>0$.
    Exact maximization over the convex surrogate feasible set then gives
    \begin{equation*}
        A(\boldsymbol{v}^{(t+1)}) - A(\boldsymbol{v}^{(t)})
        \geq M(\boldsymbol{v}^{(t+1)},\boldsymbol{v}^{(t)}) - M(\boldsymbol{v}^{(t)},\boldsymbol{v}^{(t)})
        \geq \frac{\mu_{\mathrm{MM}}}{2} \left\|\boldsymbol{v}^{(t+1)} - \boldsymbol{v}^{(t)}\right\|^2.
    \end{equation*}
    Convergence of the objective values thus implies $\|\boldsymbol{v}^{(t+1)} - \boldsymbol{v}^{(t)}\| \to 0$.

    Now take a subsequence such that $\boldsymbol{v}^{(t_j)} \to \boldsymbol{v}^{\star}$.
    Then $\boldsymbol{v}^{(t_j+1)} \to \boldsymbol{v}^{\star}$ as well.
    Since $\mathcal{V}$ is closed, $\boldsymbol{v}^{\star} \in \mathcal{V}$.
    In the nontrivial constrained case, the KKT conditions of the corresponding surrogate subproblems hold, for all sufficiently large $j$, with some multiplier $\lambda_j \geq 0$:
    \begin{equation*}
        \begin{aligned}
            &\nabla_1 M\bigl(\boldsymbol{v}^{(t_j+1)}, \boldsymbol{v}^{(t_j)}\bigr) + \lambda_j \nabla_1 \psi\bigl(\boldsymbol{v}^{(t_j+1)}, \boldsymbol{v}^{(t_j)}\bigr) = \boldsymbol{0},\\
            &\lambda_j \left[\psi\bigl(\boldsymbol{v}^{(t_j+1)}, \boldsymbol{v}^{(t_j)}\bigr) - \log(1 - \varepsilon_{\alpha,\gamma})\right] = 0.
        \end{aligned}
    \end{equation*} 
    If the original compatibility constraint is inactive at $\boldsymbol{v}^{\star}$, continuity implies that the surrogate constraint is also inactive for all sufficiently large $j$.
    Complementary slackness then gives $\lambda_j=0$.
    If the original constraint is active at $\boldsymbol{v}^{\star}$, continuity, first-order tightness, and the assumed constraint qualification imply
    \begin{equation*}
        \nabla_1\psi\bigl(\boldsymbol{v}^{(t_j+1)}, \boldsymbol{v}^{(t_j)}\bigr)
        \to
        \nabla \log Q_{\mathcal{C}}(\boldsymbol{v}^{\star})
        \neq \boldsymbol{0}.
    \end{equation*}
    Hence, the norms of the surrogate constraint gradients are bounded away from zero for all sufficiently large $j$.
    In this active-constraint case, since $\nabla_1 M$ is continuous on the compact set $\mathcal{K} \times \mathcal{K}$, there exists $C<\infty$ such that
    \begin{equation*}
        \left\|\nabla_1 M\bigl(\boldsymbol{v}^{(t_j+1)}, \boldsymbol{v}^{(t_j)}\bigr)\right\| \leq C.
    \end{equation*}
    Moreover, the preceding convergence implies that, for some $c>0$,
    \begin{equation*}
        \left\|\nabla_1 \psi\bigl(\boldsymbol{v}^{(t_j+1)}, \boldsymbol{v}^{(t_j)}\bigr)\right\| \geq c
    \end{equation*}
    for all sufficiently large $j$.
    Taking norms in the surrogate stationarity condition therefore gives
    \begin{equation*}
        0 \leq \lambda_j \leq \frac{C}{c}.
    \end{equation*}
    Together with $\lambda_j = 0$ in the inactive-constraint case, this shows that $\{\lambda_j\}$ is bounded in either case.

    Passing to a further subsequence if necessary, let $\lambda_j \to \lambda^\star \geq 0$.
    Continuity and first-order tightness yield
    \begin{equation*}
        \begin{aligned}
            &\nabla A(\boldsymbol{v}^{\star}) + \lambda^\star \nabla\log Q_{\mathcal{C}}(\boldsymbol{v}^{\star}) = \boldsymbol{0},\\
            \log Q_{\mathcal{C}}(\boldsymbol{v}^\star)& \geq \log(1 - \varepsilon_{\alpha, \gamma}),
            \qquad
            \lambda^\star\left[\log Q_{\mathcal{C}}(\boldsymbol{v}^{\star}) - \log(1 - \varepsilon_{\alpha, \gamma})\right] = 0.
        \end{aligned}
    \end{equation*}
    These are the KKT conditions for the logarithmic form of the transformed inner problem. 
    Since $Q_{\mathcal{C}} > 0$, replacing the logarithmic constraint by $Q_{\mathcal{C}} \geq 1 - \varepsilon_{\alpha, \gamma}$ only rescales the multiplier by a positive factor.

    Finally, $A=\log(-F)$ and $F < 0$ imply $\nabla A = \nabla F / F$.
    The limiting stationarity condition can therefore be written as
    \begin{equation*}
        \nabla F(\boldsymbol{v}^{\star}) - \frac{-F(\boldsymbol{v}^{\star})\lambda^\star}{Q_{\mathcal{C}}(\boldsymbol{v}^{\star})} \nabla Q_{\mathcal{C}}(\boldsymbol{v}^{\star}) = \boldsymbol{0}.
    \end{equation*}
    The rescaled multiplier is nonnegative, and complementary slackness is preserved.
    Thus, $\boldsymbol{v}^{\star}$ also satisfies the KKT conditions of the original inner minimization problem.

    In the unconstrained cases, exact maximization gives
    \begin{equation*}
        \nabla_1 M\bigl(\boldsymbol{v}^{(t_j+1)},\boldsymbol{v}^{(t_j)}\bigr) = \boldsymbol{0}.
    \end{equation*}
    Passing to the limit and using first-order tightness yields $\nabla A(\boldsymbol{v}^{\star}) = \boldsymbol{0}$, and hence $\nabla F(\boldsymbol{v}^{\star}) = \boldsymbol{0}$.
    This completes the proof.
\end{proof}

\begin{proof}[\bf Proof of Proposition~\ref{prop:pl_inner_value_bound}]
    Fix $D$ and $\pi$, and use the enumeration $\mathfrak{S}_K = \{y^{(1)}, \ldots, y^{(N_K)}\}$ and the notation $q_m(\boldsymbol{v})$, $h_m(\pi)$, $\mathcal{I}_{\mathcal{C}}$, $Q_{\mathcal{C}}(\boldsymbol{v})$, $\mathcal{V}$, and $F(\boldsymbol{v})$ defined above.

    If $\mathcal{I}_{\mathcal{C}} \neq \varnothing$, then every $\boldsymbol{v} \in \mathcal{V}$ satisfies
    \begin{equation*}
        \begin{aligned}
            F(\boldsymbol{v}) 
            & = \sum_{m=1}^{N_K} q_m(\boldsymbol{v}) h_m(\pi)\\
            &\geq Q_{\mathcal{C}}(\boldsymbol{v}) \min_{m \in \mathcal{I}_{\mathcal{C}}}h_m(\pi) + (1 - Q_{\mathcal{C}}(\boldsymbol{v})) \min_{1 \leq m \leq N_K} h_m(\pi)\\
            &\geq (1 - \varepsilon_{\alpha,\gamma}) \min_{m \in \mathcal{I}_{\mathcal{C}}}h_m(\pi) + \varepsilon_{\alpha,\gamma} \min_{1 \leq m \leq N_K}h_m(\pi)
            = L_{\mathrm{relax}}(\pi; D).
        \end{aligned}
    \end{equation*}
    The second inequality follows from $Q_{\mathcal{C}}(\boldsymbol{v}) \geq 1 - \varepsilon_{\alpha,\gamma}$ and $\min_{m \in \mathcal{I}_{\mathcal{C}}}h_m(\pi) \geq \min_{1 \leq m \leq N_K}h_m(\pi)$.
    If $\mathcal{I}_{\mathcal{C}} = \varnothing$, the compatibility constraint is omitted, and $F(\boldsymbol{v}) \geq \min_{1 \leq m \leq N_K}h_m(\pi) = L_{\mathrm{relax}}(\pi; D)$.
    Taking the infimum over $\boldsymbol{v} \in \mathcal{V}$ gives $L_{\mathrm{relax}}(\pi;D) \leq \mathcal{L}_{\alpha,\gamma}^{\mathrm{pess}}(\pi; D)$.
    Since $\widetilde{\boldsymbol{v}} \in \mathcal{V}$,
    \begin{equation*}
        \mathcal{L}_{\alpha,\gamma}^{\mathrm{pess}}(\pi;D)
        \leq F(\widetilde{\boldsymbol{v}})
        = \widetilde{\mathcal{L}}_{\alpha,\gamma}^{\mathrm{pess}}(\pi;D).
    \end{equation*}
    Combining these inequalities proves (\ref{eq:pl_inner_value_error_bound}).
\end{proof}

\section{Suboptimality Upper Bound} \label{app:suboptimality_upper_bound}

\subsection{Full Theorem Statement and Discussion} \label{app:suboptimality_theorem_and_discussion}
Recall the conformal uncertainty envelope
\begin{equation*}
    \Psi_{\alpha,\gamma}(\delta)
    = \min\left\{1,\,
        \varepsilon_{\alpha,\gamma}^2
        + (1-\varepsilon_{\alpha,\gamma}^2)
        \left(\widetilde{\eta}_{\alpha,\gamma}+\gamma+
        \frac{\log(2/\delta)}{n_{\mathrm{cal}}}\right)
    \right\},
\end{equation*}
and the combined error term
\begin{equation*}
    \mathfrak{E}_{\alpha,\gamma}(\delta)
    = \underbrace{\mathcal{E}_\Pi}_{\text{policy approximation error}}
    + \underbrace{\Psi_{\alpha,\gamma}(\delta)}_{\text{conformal uncertainty}}
    + \underbrace{\frac{d_\Pi + \log(2/\delta)}{n_{\mathrm{opt}}}}_{\text{statistical estimation error}}.
    \tag{\ref{eq:subopt_main_error_term}}
\end{equation*}
We now give the full statement of Theorem~\ref{thm:subopt_main_guarantees}, including explicit model-specific constants.

\medskip
\noindent
\textbf{Theorem~\ref{thm:subopt_main_guarantees}} (Finite-Sample Suboptimality Guarantee, full statement)\textbf{.}
\emph{
    Suppose that Assumptions~\ref{ass:subopt_sampling} and \ref{ass:bounded_comparison_scores} hold.
    Assume that the exact empirical LatentCP-DPO objective attains its maximum over $\Pi$, and that the localized-complexity condition
    \begin{equation} \label{eq:subopt_localized_complexity_condition}
        \mathfrak{R}_{n_{\mathrm{opt}}}(\zeta) 
        \leq C_{\mathrm{loc}}\left(\sqrt{\frac{d_\Pi\zeta}{n_{\mathrm{opt}}}} + \frac{d_\Pi}{n_{\mathrm{opt}}}\right),
        \qquad 
        \zeta > 0
    \end{equation}
    holds for some $C_{\mathrm{loc}} \geq 1$ and $d_\Pi \geq 1$ independent of $n_{\mathrm{opt}}$, where $\mathfrak{R}_{n_{\mathrm{opt}}}(\zeta)$ is defined in the next subsection.
    Assume further that $C_{\mathrm{comp}}(\pi^\star;\Pi) < \infty$, and that $\widehat{\pi}_{\mathrm{LCP}}$ shares the response support of $\pi^\star$ for $\rho$-almost every prompt.
    Fix $0< \gamma < \alpha < 1$ independently of the calibration sample, and let $\delta \in (0, 1)$.
    Conditional on the pilot sample, with probability at least $1 - \delta$ over the calibration and optimization samples, every comparator policy $\pi^\dagger$ for which $J(\pi)$ and $H(\pi)$ are finite for $\pi \in \{\pi^\star, \widehat\pi_{\mathrm{LCP}}, \pi^\dagger\}$ satisfies
    \begin{equation*}
        \mathrm{SubOpt}_\beta(\widehat{\pi}_{\mathrm{LCP}}, \pi^\dagger)
        \leq \frac{C_{\mathrm{M}} C_{\mathrm{comp}}(\pi^\star; \Pi)}{2\beta} \mathfrak{E}_{\alpha,\gamma}(\delta)
            - \beta\mathbb{E}_{s \sim \rho} \left[D_{\mathrm{KL}}\left(\pi^\dagger(\cdot \mid s) \middle\| \pi^\star(\cdot \mid s)\right)\right].
        \tag{\ref{eq:subopt_main_regularized}}
    \end{equation*}
    Here $C_{\mathrm{M}} > 0$ depends on $B$, the model $\mathrm{M}$, the localized-complexity constant $C_{\mathrm{loc}}$, and, for PL feedback, the fixed number of candidates $K$.
    For BT feedback, one may take
    \begin{equation} \label{eq:model_specific_constant_bt}
        C_{\mathrm{BT}} 
        = \max\left\{\frac{1}{\sigma(B)\sigma(-B)} + 4,\ \frac{20}{\sigma(B)^2\sigma(-B)^2},\ \frac{4096\log(1+e^B)C_{\mathrm{loc}}^2}{\sigma(B)\sigma(-B)}\right\},
    \end{equation}
    whereas for PL feedback, one may take
    \begin{equation} \label{eq:model_specific_constant_pl}
        C_{\mathrm{PL}}
        = \max\Bigl\{2K (K - 1)e^{2B} + 4,\ 40K^2(K - 1)^2e^{4B},\ 4096K e^{2B}\bigl[\log(K!) + BK\bigr] C_{\mathrm{loc}}^2\Bigr\}.
    \end{equation}
    In particular, setting $\pi^\dagger = \pi^\star$ removes the comparator-dependent KL correction.
}

The error term $\mathfrak{E}_{\alpha,\gamma}(\delta)$ separates three contributions to the upper bound.
The policy approximation error $\mathcal{E}_\Pi$ vanishes when $\pi^\star\in\Pi$, because the prompt-dependent normalizing term in (\ref{eq:kl_regularized_solution_r_star}) cancels from the comparison scores.
The conformal uncertainty term $\Psi_{\alpha,\gamma}(\delta)$ accounts for the compatibility tolerance $\varepsilon_{\alpha,\gamma}$, disagreement within the conformal prediction sets, preference-outcome miscoverage, and finite-sample calibration.
Finally, $(d_\Pi+\log(2/\delta))/n_{\mathrm{opt}}$ is the estimation contribution from the independent label-free optimization contexts.

The bound also reveals a trade-off in choosing $\gamma$.
For fixed pilot scores and calibration data, decreasing $\gamma$ reduces $\varepsilon_{\alpha, \gamma}$ but can only enlarge the conformal prediction sets $\widehat{\mathcal{C}}_\gamma(D)$, potentially leaving more preference comparisons unresolved.
Thus, smaller $\gamma$ need not yield a smaller conformal uncertainty term; for the stated guarantee, $\gamma$ must be chosen independently of the calibration data.
For fixed $\alpha$, $\gamma$, and $\delta$, the explicit calibration correction in $\Psi_{\alpha,\gamma}(\delta)$ is $O(n_{\mathrm{cal}}^{-1})$, but this is not a rate for the entire conformal term because $\widetilde{\eta}_{\alpha, \gamma}$ also depends on the calibrated sets.
Increasing $n_{\mathrm{opt}}$ reduces the optimization-sample contribution, while the approximation and conformal contributions need not vanish.

The explicit constants $C_{\mathrm{BT}}$ and $C_{\mathrm{PL}}$ show how the upper bound depends on the comparison-score range $B$ and, for PL feedback, the number of candidates $K$.
A larger $B$ allows more extreme preference probabilities and increases these constants.
For fixed $B$ and $C_{\mathrm{loc}}$, the displayed PL constant grows as $O(K^4)$, although the space of complete rankings contains $K!$ outcomes.
Polynomial dependence on $K$ also appears in related PL guarantees \citep[Theorem~4.1]{zhu2023principled}, although the estimator and performance criterion differ.

In (\ref{eq:subopt_localized_complexity_condition}), $\mathfrak{R}_{n_{\mathrm{opt}}}(\zeta)$ is the conditional expected Rademacher complexity of the localized excess pessimistic-loss class, defined in (\ref{eq:subopt_rademacher_complexity}).
Lemma~\ref{lem:subopt_common_policy_classes} gives sufficient conditions for this bound, covering finite-dimensional affine comparison-score classes (including those induced by log-linear policy classes with a shared finite-dimensional feature map \citep{zhou2025unified}) and uniformly Lipschitz comparison-score parameterizations over bounded parameter domains.
Its proof is provided in Appendix~\ref{app:subopt_proofs_upper_bounds}.

\begin{lemma}[Sufficient Conditions for Localized Complexity] \label{lem:subopt_common_policy_classes}
    Under Assumptions~\ref{ass:subopt_sampling} and~\ref{ass:bounded_comparison_scores}, condition on the pilot and calibration samples, and treat $\Pi$ and its parameterization as fixed under this conditioning.
    Let $\mathcal{D}_0$ be a set of target contexts with probability one on which the bounds in Assumption~\ref{ass:bounded_comparison_scores} hold simultaneously for all $\pi \in \Pi$.
    Suppose that, for some integer $p \geq 1$, at least one of the following conditions holds:
    \begin{enumerate}
        \item \textbf{Finite-dimensional comparison scores.}
        The functions $\{D \mapsto x_\pi(D): \pi \in \Pi\}$, restricted to $\mathcal{D}_0$, lie in an affine function space of dimension at most $p$;
        
        \item \textbf{Uniformly Lipschitz parameterization.}
        The policy class admits a representation $\Pi = \{\pi_\vartheta: \vartheta \in \mathcal{W}\}$, where $\mathcal{W} \subseteq \mathbb{R}^p$ is bounded and, for some $L_x < \infty$,
        \begin{equation*}
            \sup_{D\in\mathcal{D}_0} \left\|x_{\pi_\vartheta}(D) - x_{\pi_{\vartheta'}}(D)\right\| \leq L_x\|\vartheta-\vartheta'\|,
            \qquad
            \vartheta, \vartheta' \in \mathcal{W}.
        \end{equation*}
    \end{enumerate}
    Assume that $p$ and the model bounds, together with $\operatorname{diam}(\mathcal{W})$ and $L_x$ in the second case, do not depend on $n_{\mathrm{opt}}$.
    Then there exists $C_{\mathrm{loc}} \geq 1$, depending only on the model bounds and, in the second case, $L_x\operatorname{diam}(\mathcal{W})$, but not on $p$, $n_{\mathrm{opt}}$, or $\zeta$, such that
    \begin{equation*}
        \mathfrak{R}_{n_{\mathrm{opt}}}(\zeta)
        \leq C_{\mathrm{loc}}\left(\sqrt{\frac{p\zeta}{n_{\mathrm{opt}}}} + \frac{p}{n_{\mathrm{opt}}}\right),
        \qquad \zeta>0.
    \end{equation*}
    Consequently, (\ref{eq:subopt_localized_complexity_condition}) holds with $d_\Pi = p$.
\end{lemma}

\begin{remark}[Role of the pilot sample]
    Theorem~\ref{thm:subopt_main_guarantees} analyzes conformal calibration and pessimistic optimization conditional on the realized pilot sample. 
    Its bound has no separate error term involving $n_{\mathrm{pil}}$.
    The pilot fit still affects the bound: the nonconformity scores, together with the calibration data, determine the conformal prediction sets, whose informativeness is reflected in $\widetilde{\eta}_{\alpha, \gamma}$.
    The stated bound requires neither consistency nor a convergence rate for the pilot estimator.
\end{remark}

\begin{remark}[Comparison coefficient for log-linear policies]
    The upper bound in Theorem~\ref{thm:subopt_main_guarantees} amplifies error by $C_{\mathrm{comp}}(\pi^\star;\Pi)$.
    This is analogous to the role of relative condition numbers in log-linear offline alignment \citep{zhou2025unified}, although the coefficients differ.
    For BT feedback, suppose that $\Pi = \{\pi_\theta: \theta \in \Theta\}$ has a common log-linear parameterization with bounded feature map $\phi$, that $\pi^\star = \pi_{\theta^\star} \in \Pi$, and that the two comparison responses are drawn conditionally independently from $\pi_{\mathrm{ref}}$.
    For a fixed evaluation response distribution $\nu$, write $v_\theta = \theta - \theta^\star$.
    Then
    \begin{equation*}
        C_{\mathrm{comp}}(\nu;\Pi) 
        = \sup_{\theta: \pi_\theta \in \Pi} \frac{v_\theta^\top \Sigma_{\nu,\pi_\theta}^{\mathrm{diff}} v_\theta}{v_\theta^\top \Sigma_{\pi_{\mathrm{ref}},\pi_{\mathrm{ref}}}^{\mathrm{diff}} v_\theta},
    \end{equation*}
    where
    \begin{equation*}
        \Sigma_{\nu_1, \nu_2}^{\mathrm{diff}} = \mathbb{E}_{\substack{s \sim \rho \\ a \sim \nu_1(\cdot \mid s), a' \sim \nu_2(\cdot \mid s)}} \left[(\phi(s,a) - \phi(s,a'))(\phi(s,a) - \phi(s,a'))^\top\right],
    \end{equation*}
    with $a$ and $a'$ conditionally independent given $s$.
    The coefficient involves candidate-policy response distributions but evaluates each policy only along its own parameter difference $v_\theta = \theta - \theta^\star$, rather than over arbitrary feature-space directions.
\end{remark}

\paragraph{Additional Discussion}
Theorem~\ref{thm:subopt_main_guarantees} concerns an exact empirical optimizer.
For a possibly data-dependent output $\widetilde{\pi} \in \Pi$, define the exact empirical pessimistic loss and its optimization gap by
\begin{equation*}
    \widehat{L}(\pi) 
    = -\frac{1}{n_{\mathrm{opt}}} \sum_{i=1}^{n_{\mathrm{opt}}} \mathcal{L}_{\alpha,\gamma}^{\mathrm{pess}}(\pi;D_i^{\mathrm{opt}}),
    \qquad
    \Delta_{\mathrm{opt}} = \widehat{L}(\widetilde{\pi}) - \inf_{\pi\in\Pi}\widehat L(\pi)\geq 0.
\end{equation*}
The empirical infimum need not be attained.

\begin{corollary}[Suboptimality under Approximate Empirical Optimization] \label{cor:subopt_approximate_optimization}
    Suppose the assumptions of Theorem~\ref{thm:subopt_main_guarantees} hold.
    Conditional on the pilot sample, with probability at least $1 - \delta$ over the calibration and optimization samples, the following bound holds simultaneously for every measurable output $\widetilde{\pi} \in \Pi$ that shares the response support of $\pi^\star$ for $\rho$-almost every prompt and every comparator $\pi^\dagger$ such that $J(\pi)$ and $H(\pi)$ are finite for each $\pi \in \{\pi^\star, \widetilde{\pi}, \pi^\dagger\}$:
    \begin{equation}\label{eq:subopt_approximate_optimization_bound}
        \begin{aligned}
            \mathrm{SubOpt}_{\beta}(\widetilde{\pi},\pi^\dagger)
            \leq& \frac{C_{\mathrm{M}}C_{\mathrm{comp}}(\pi^\star;\Pi)}{2\beta} \mathfrak{E}_{\alpha,\gamma}(\delta)
            + \frac{3C_{\mathrm{comp}}(\pi^\star;\Pi)}{\beta\kappa_{\mathrm{M}}} \Delta_{\mathrm{opt}}\\
            & - \beta\mathbb{E}_{s\sim\rho}\left[D_{\mathrm{KL}}\left(\pi^\dagger(\cdot\mid s)\middle\|\pi^\star(\cdot\mid s)\right)\right],
        \end{aligned}
    \end{equation}
    where $\kappa_{\mathrm{M}}$ is the curvature constant in Lemma~\ref{lem:pointwise_benchmark_quadratic_growth}.
\end{corollary}

The additional term measures the gap in the exact empirical objective.
The proof is given in Appendix~\ref{app:subopt_proof_of_the_main_upper_bound}.

For PL feedback, the effect of the MM approximation on $\Delta_{\mathrm{opt}}$ can be made explicit.
Let $\widetilde{\boldsymbol{v}}_i$ be a feasible MM output evaluated at $\widetilde{\pi}$ on $D_i^{\mathrm{opt}}$.
For each $D_i^{\mathrm{opt}}$, let $\boldsymbol{q}_i$ and $\boldsymbol{h}_i$ denote the vectors in (\ref{eq:pl_pessimistic_objective_computational_formulation}) formed using that context's ranking enumeration.
Holding the MM outputs fixed as $\pi$ varies, define
\begin{equation*}
    \widehat{L}_{\mathrm{fr}}(\pi)
    = -\frac{1}{n_{\mathrm{opt}}}\sum_{i=1}^{n_{\mathrm{opt}}} \boldsymbol{q}_i(\widetilde{\boldsymbol{v}}_i)^\top \boldsymbol{h}_i(\pi),
\end{equation*}
and set
\begin{equation*}
    \epsilon_{\mathrm{in}}
    = \widehat{L}(\widetilde{\pi}) - \widehat{L}_{\mathrm{fr}}(\widetilde{\pi}),
    \qquad
    \epsilon_{\mathrm{out}}
    = \widehat{L}_{\mathrm{fr}}(\widetilde{\pi}) - \inf_{\pi\in\Pi}\widehat{L}_{\mathrm{fr}}(\pi).
\end{equation*}
We have $\widehat{L}_{\mathrm{fr}}(\pi) \leq \widehat{L}(\pi)$ for every $\pi\in\Pi$.
Consequently, Proposition~\ref{prop:pl_inner_value_bound} gives
\begin{equation}
    \Delta_{\mathrm{opt}}
    \leq \epsilon_{\mathrm{in}}+\epsilon_{\mathrm{out}}
    \leq \epsilon_{\mathrm{out}} + \frac{1}{n_{\mathrm{opt}}} \sum_{i=1}^{n_{\mathrm{opt}}} \left[\widetilde{\mathcal{L}}_{\alpha, \gamma}^{\mathrm{pess}}(\widetilde{\pi}; D_i^{\mathrm{opt}}) - L_{\mathrm{relax}}(\widetilde{\pi}; D_i^{\mathrm{opt}})\right].
\end{equation}
Thus, $\Delta_{\mathrm{opt}}$ is bounded by the sum of the outer optimization gap and a computable upper bound on the inner approximation error.

\paragraph{Proof Roadmap}
The proof of Theorem~\ref{thm:subopt_main_guarantees} proceeds in three ingredients.
Appendix~\ref{app:subopt_excess_preference_risk_bounds} controls the excess pessimistic preference risk of the exact empirical optimizer.
Appendix~\ref{app:preference_to_policy_transfer} then uses $C_{\mathrm{comp}}$ to obtain a KL-regularized suboptimality bound in terms of pessimistic benchmark bias, approximation error, and estimation error.
Appendix~\ref{app:effect_of_conformal_uncertainty} bounds the bias and approximation terms using conformal uncertainty and finite-sample calibration.
Appendix~\ref{app:subopt_proof_of_the_main_upper_bound} then combines these results to prove the Theorem~\ref{thm:subopt_main_guarantees} and Corollary~\ref{cor:subopt_approximate_optimization}.
Proofs of the auxiliary results are collected in Appendix~\ref{app:subopt_proofs_upper_bounds}.

\subsection{Setup and Auxiliary Quantities}
For the optimization-sample analysis, we condition on the realized pilot and calibration samples.
The sets $\widehat{\mathcal{C}}_\gamma(D)$ and $\widehat{\mathcal{U}}_\alpha(D;\gamma)$ are then fixed functions of $D$, while Assumption~\ref{ass:subopt_sampling} ensures that $D_1^{\mathrm{opt}},\ldots,D_{n_{\mathrm{opt}}}^{\mathrm{opt}}$ remain i.i.d. draws from the target context distribution.
When deriving probability bounds over the calibration sample, we instead condition only on the pilot sample.

For a BT scalar score or a centered PL score $x$, let $p_{\mathrm{M},x}(y\mid D)$ denote the corresponding preference likelihood, and define
\begin{equation*}
    \ell_x(y,D)=-\log p_{\mathrm{M},x}(y\mid D).
\end{equation*}
The unit-level pessimistic loss is
\begin{equation}\label{eq:subopt_unit_level_pessimistic_loss}
    f_D(x)
    =\sup_{\boldsymbol{u}\in\widehat{\mathcal{U}}_\alpha(D;\gamma)}
    \mathbb{E}_{y\sim P_{\mathrm{M},\boldsymbol{u}}(\cdot\mid D)}
    [\ell_x(y,D)].
\end{equation}
By the definition of $x_\pi(D)$, we have $p_{\mathrm{M}, x_\pi(D)}(y \mid D) = p_{\mathrm{M}, \pi}(y \mid D)$.
Thus, $f_D(x_\pi(D)) = -\mathcal{L}_{\alpha, \gamma}^{\mathrm{pess}}(\pi; D)$.

Define the conditional population risk and its empirical counterpart by
\begin{equation*}
    L(\pi)=\mathbb{E}_D[f_D(x_\pi(D))],
    \qquad
    \widehat L(\pi)
    =\frac{1}{n_{\mathrm{opt}}}\sum_{i=1}^{n_{\mathrm{opt}}}
    f_{D_i^{\mathrm{opt}}}(x_\pi(D_i^{\mathrm{opt}})),
\end{equation*}
where $\mathbb{E}_D$ is over a fresh target context, conditional on the pilot and calibration samples.
Under the empirical-attainment condition of Theorem~\ref{thm:subopt_main_guarantees}, the exact LatentCP-DPO estimator satisfies
\begin{equation}\label{eq:subopt_empirical_minimizer}
    \widehat{\pi}_{\mathrm{LCP}}
    \in\operatorname*{argmin}_{\pi\in\Pi}\widehat L(\pi).
\end{equation}
The guarantees below concern exact inner supremum values in (\ref{eq:subopt_unit_level_pessimistic_loss}) and a global empirical minimizer in (\ref{eq:subopt_empirical_minimizer}).

\subsection{Excess Preference-Risk Bounds} \label{app:subopt_excess_preference_risk_bounds}
Define the bounded comparison-score domain by
\begin{equation*}
    \mathcal{X}(D)
    = \begin{cases}
        [-B, B], & \mathrm{M} = \mathrm{BT},\\
        \left\{x \in \mathbb{R}^{K}: \boldsymbol{1}_{K}^\top x = 0, \|x\|_\infty \leq B\right\}, & \mathrm{M} = \mathrm{PL}.
    \end{cases}
\end{equation*}
By Assumption~\ref{ass:bounded_comparison_scores}, $\mathcal{X}(D)$ contains $x^\star(D)$ and $x_\pi(D)$ for every $\pi\in\Pi$, almost surely.
We first establish uniform bounds on the negative log-likelihood and the corresponding pessimistic loss.

\begin{lemma}[Uniform Boundedness of the Pessimistic Loss] \label{lem:uniform_boundedness_pessimistic_loss}
    Under Assumption~\ref{ass:bounded_comparison_scores}, define
    \begin{equation*}
        H_{\mathrm{M}} = \begin{cases}
            \log(1 + e^B), & \mathrm{M} = \mathrm{BT},\\
            \log(K!) + BK, & \mathrm{M} = \mathrm{PL}.
        \end{cases}
    \end{equation*}
    For almost every target context $D$, every $x \in \mathcal{X}(D)$, and every $y \in \mathcal{Y}(D)$,
    \begin{equation*}
        0 \leq \ell_x(y, D) \leq H_{\mathrm{M}},
        \qquad
        0 \leq f_D(x) \leq H_{\mathrm{M}}.
    \end{equation*}
\end{lemma}

Having established uniform boundedness, we next relate excess pessimistic loss to the squared distance from a pointwise minimizer.

\begin{lemma}[Pointwise Benchmark and Quadratic Growth] \label{lem:pointwise_benchmark_quadratic_growth}
    Under Assumption~\ref{ass:bounded_comparison_scores}, define
    \begin{equation*}
        \kappa_{\mathrm{M}}
        = \begin{cases}
            \sigma(B) \sigma(-B), & \mathrm{M} = \mathrm{BT},\\
            e^{-2B} / K, & \mathrm{M} = \mathrm{PL}.
        \end{cases}
    \end{equation*}
    For almost every target context $D$, the function $f_D$ is continuous and $\kappa_{\mathrm{M}}$-strongly convex on $\mathcal{X}(D)$.
    Consequently, it admits a unique minimizer%
    \footnote{The benchmark minimizes the pessimistic loss over the bounded score domain, not over the policy class.
    It need not equal $x^\star(D)$ or be realizable by any policy in $\Pi$.}
    \begin{equation*}
        \bar{x}(D) = \operatorname*{argmin}_{x\in\mathcal{X}(D)} f_D(x).
    \end{equation*}
    Moreover, for every $x \in \mathcal{X}(D)$,
    \begin{equation} \label{eq:subopt_pointwise_quadratic_growth}
        \frac{\kappa_{\mathrm{M}}}{2} \|x - \bar{x}(D)\|^2
        \leq f_D(x) - f_D(\bar{x}(D))
        \leq H_{\mathrm{M}}.
    \end{equation}
\end{lemma}

We next control sampling fluctuations uniformly over the policy class.
Define the pointwise excess pessimistic loss and its best-in-class population excess pessimistic risk by
\begin{equation*}
    e_\pi(D) 
    = f_D(x_\pi(D)) - f_D(\bar{x}(D)),
    \qquad
    \mathcal{E}_{\Pi}^{\mathrm{pess}} = \inf_{\pi \in \Pi} \mathbb{E}_D[e_\pi(D)].
\end{equation*}
Lemma~\ref{lem:pointwise_benchmark_quadratic_growth} implies that $0 \leq e_\pi(D) \leq H_{\mathrm{M}}$ and hence
\begin{equation*}
    \operatorname{Var}_D(e_\pi(D)) 
    \leq \mathbb{E}_D[e_\pi(D)^2]
    \leq H_{\mathrm{M}} \mathbb{E}_D[e_\pi(D)].
\end{equation*}
To exploit the mean-dependent variance bound, for $\zeta > 0$, define the localized class of rescaled excess-loss functions by
\begin{equation*}
    \mathcal{H}_\zeta
    = \left\{\frac{\lambda e_\pi}{H_{\mathrm{M}}}: \pi \in \Pi, 0 \leq \lambda \leq 1, \frac{\lambda\mathbb{E}_D[e_\pi(D)]}{H_{\mathrm{M}}} \leq \zeta\right\}.
\end{equation*}
Every $h \in \mathcal{H}_\zeta$ takes values in $[0, 1]$ and satisfies $\operatorname{Var}_D(h(D)) \leq \zeta$.
This rescaling is used only for analysis and does not alter the policy class or the training objective.

Define the conditional expected Rademacher complexity of $\mathcal{H}_\zeta$ by
\begin{equation} \label{eq:subopt_rademacher_complexity}
    \mathfrak{R}_{n_{\mathrm{opt}}}(\zeta)
    = \mathbb{E} \left[\sup_{h \in \mathcal{H}_\zeta}\left|\frac{1}{n_{\mathrm{opt}}} \sum_{i=1}^{n_{\mathrm{opt}}} \epsilon_i h(D_i^{\mathrm{opt}})\right|\right].
\end{equation}
Here, $\epsilon_1, \ldots, \epsilon_{n_{\mathrm{opt}}}$ are independent Rademacher signs, each taking the values $+1$ and $-1$ with equal probability, and are independent of the optimization contexts.
The expectation is over both the optimization contexts and these signs, conditional on the pilot and calibration samples.
We assume that the displayed suprema define random variables, so that their expectations are well defined.

\begin{lemma}[Relative Uniform-Convergence Bound] \label{lem:relative_uniform_convergence}
    Under Assumptions~\ref{ass:subopt_sampling} and~\ref{ass:bounded_comparison_scores}, fix $\delta \in (0, 1)$.
    Let $\zeta_{n_{\mathrm{opt}}}>0$ be fixed conditional on the pilot and calibration samples and satisfy
    \begin{equation} \label{eq:subopt_localized_radius}
        16\mathfrak{R}_{n_{\mathrm{opt}}}(\zeta_{n_{\mathrm{opt}}}) \leq \zeta_{n_{\mathrm{opt}}},
        \qquad
        \frac{64\log(1/\delta)}{n_{\mathrm{opt}}} \leq \zeta_{n_{\mathrm{opt}}}.
    \end{equation}
    Set $\xi_{n_{\mathrm{opt}}} = H_{\mathrm{M}} \zeta_{n_{\mathrm{opt}}}$.
    Then, with conditional probability at least $1 - \delta$, simultaneously for all $\pi \in \Pi$,
    \begin{equation} \label{eq:subopt_relative_uniform_deviation}
        \left|\frac{1}{n_{\mathrm{opt}}} \sum_{i=1}^{n_{\mathrm{opt}}} e_\pi(D_i^{\mathrm{opt}}) - \mathbb{E}_D[e_\pi(D)]\right|
        \leq \frac{1}{3} \max\left\{\mathbb{E}_D[e_\pi(D)], \xi_{n_{\mathrm{opt}}}\right\}.
    \end{equation}
    
    On the same event, the exact estimator satisfies
    \begin{equation} \label{eq:subopt_localized_oracle_bound}
        \mathbb{E}_D\left[e_{\widehat{\pi}_{\mathrm{LCP}}}(D)\right]
        \leq \min\left\{H_{\mathrm{M}}, \mathcal{E}_{\Pi}^{\mathrm{pess}} + \max\{\mathcal{E}_{\Pi}^{\mathrm{pess}}, \xi_{n_{\mathrm{opt}}}\}\right\}.
    \end{equation}
    Consequently,
    \begin{equation} \label{eq:subopt_estimation_score_bound}
        \mathbb{E}_D\left[\left\|x_{\widehat{\pi}_{\mathrm{LCP}}}(D) -\bar{x}(D)\right\|^2\right]
        \leq \frac{2}{\kappa_{\mathrm{M}}} \min\left\{H_{\mathrm{M}}, \mathcal{E}_{\Pi}^{\mathrm{pess}} + \max\{\mathcal{E}_{\Pi}^{\mathrm{pess}}, \xi_{n_{\mathrm{opt}}}\}\right\}.
    \end{equation}
\end{lemma}

\begin{remark}
    Since every $h \in \mathcal{H}_\zeta$ takes values in $[0,1]$, we have $\mathfrak{R}_{n_{\mathrm{opt}}}(\zeta) \leq 1$.
    Thus, the conditions in (\ref{eq:subopt_localized_radius}) always admit a solution; for example, one may take
    \begin{equation*}
        \zeta_{n_{\mathrm{opt}}} = \max\left\{16, \frac{64\log(1/\delta)}{n_{\mathrm{opt}}}\right\}.
    \end{equation*}
    This choice establishes existence but yields only a trivial bound.
    Under the localized-complexity condition, Corollary~\ref{cor:subopt_explicit_preference_rate} gives a radius that tends to zero as $n_{\mathrm{opt}} \to \infty$ for fixed $C_{\mathrm{loc}}$, $d_\Pi$, and $\delta$; Lemma~\ref{lem:subopt_common_policy_classes} gives sufficient conditions for that complexity bound.
\end{remark}

We now relate the preceding estimation bound to the true preference distribution. 
Let $P_{\mathrm{M}, x}(\cdot\mid D)$ denote the distribution with probability mass function $p_{\mathrm{M}, x}(\cdot \mid D)$.
For any $\pi \in \Pi$, define the excess preference risk by
\begin{equation*}
    \mathcal{R}(\pi) 
    = \mathbb{E}_D \left[D_{\mathrm{KL}}\left(P_{\mathrm{M},x^\star(D)}(\cdot\mid D) \middle\| P_{\mathrm{M},x_\pi(D)}(\cdot\mid D)\right)\right].
\end{equation*}
Under model correctness, this is the excess population negative log-likelihood relative to the true comparison score $x^\star(D)$, with both losses evaluated under the true preference distribution.
Since the pessimistic benchmark $\bar{x}(D)$ need not coincide with $x^\star(D)$, define its preference-risk bias by
\begin{equation*}
    b_{\alpha, \gamma}
    = \mathbb{E}_D\left[D_{\mathrm{KL}}\left(P_{\mathrm{M}, x^\star(D)}(\cdot \mid D) \middle\| P_{\mathrm{M},\bar{x}(D)}(\cdot\mid D)\right)\right].
\end{equation*}
This bias is fixed conditional on the realized pilot and calibration samples and is not assumed to vanish.

\begin{theorem}[Excess Preference-Risk Bound] \label{thm:subopt_excess_preference_risk}
    Under the conditions of Lemma~\ref{lem:relative_uniform_convergence}, define
    \begin{equation*}
        \Lambda_{\mathrm{M}}
        = \begin{cases}
            \dfrac{1}{4}, & \mathrm{M} = \mathrm{BT},\\
            \dfrac{K-1}{2}, & \mathrm{M}=\mathrm{PL}.
        \end{cases}
    \end{equation*}
    Then, with conditional probability at least $1 - \delta$,
    \begin{equation} \label{eq:subopt_true_preference_risk_bound}
        \sqrt{\mathcal{R}(\widehat{\pi}_{\mathrm{LCP}})}
        \leq \sqrt{b_{\alpha, \gamma}}
         + \sqrt{\frac{\Lambda_{\mathrm{M}}}{\kappa_{\mathrm{M}}} \min\left\{H_{\mathrm{M}}, \mathcal{E}_{\Pi}^{\mathrm{pess}} + \max\left\{\mathcal{E}_{\Pi}^{\mathrm{pess}}, \xi_{n_{\mathrm{opt}}}\right\}\right\}}.
    \end{equation}
    Moreover, deterministically, $\mathcal{R}(\pi) \leq H_{\mathrm{M}}$ for every $\pi \in \Pi$.
\end{theorem}

Theorem~\ref{thm:subopt_excess_preference_risk} bounds the learned policy's excess preference risk in terms of the pessimistic benchmark bias, approximation to the pointwise pessimistic benchmark, and optimization-sample error.
The following corollary makes the last term explicit under a localized-complexity condition.

\begin{corollary}[Explicit Statistical Rate under Localized Complexity] \label{cor:subopt_explicit_preference_rate}
    Under the conditions of Theorem~\ref{thm:subopt_excess_preference_risk}, suppose that, conditional on the pilot and calibration samples, there exist constants $C_{\mathrm{loc}} \geq 1$ and $d_\Pi \geq 1$, independent of $n_{\mathrm{opt}}$, such that
    \begin{equation*}
        \mathfrak{R}_{n_{\mathrm{opt}}}(\zeta) 
        \leq C_{\mathrm{loc}}\left(\sqrt{\frac{d_\Pi\zeta}{n_{\mathrm{opt}}}} + \frac{d_\Pi}{n_{\mathrm{opt}}}\right),
        \qquad 
        \zeta > 0
        \tag{\ref{eq:subopt_localized_complexity_condition}}
    \end{equation*}
    For any $\delta \in (0,1)$, the choice
    \begin{equation} \label{eq:subopt_explicit_localized_radius}
        \zeta_{n_{\mathrm{opt}}}
        = \frac{\max\left\{1024C_{\mathrm{loc}}^2d_\Pi, 64\log(1/\delta)\right\}}{n_{\mathrm{opt}}}
    \end{equation}
    satisfies (\ref{eq:subopt_localized_radius}).
    Consequently, with conditional probability at least $1 - \delta$,
    \begin{equation} \label{eq:subopt_explicit_preference_rate}
        \mathcal{R}(\widehat{\pi}_{\mathrm{LCP}})
        \leq \underbrace{2b_{\alpha,\gamma} + \frac{4\Lambda_{\mathrm{M}}}{\kappa_{\mathrm{M}}} \mathcal{E}_{\Pi}^{\mathrm{pess}}}_{\text{benchmark bias and approximation}}
        + \underbrace{\frac{2\Lambda_{\mathrm{M}}H_{\mathrm{M}}}{\kappa_{\mathrm{M}}n_{\mathrm{opt}}} \max\left\{1024C_{\mathrm{loc}}^2d_\Pi, 64\log(1/\delta)\right\}}_{\text{statistical error}}.
    \end{equation}
    The right-hand side may additionally be capped at $H_{\mathrm{M}}$.
\end{corollary}

Conditional on the pilot and calibration samples, condition (\ref{eq:subopt_localized_complexity_condition}) constrains $\Pi$ through the localized excess pessimistic-loss class it induces.
Lemma~\ref{lem:subopt_common_policy_classes} gives sufficient conditions for this bound.

\subsection{Preference-to-Policy Transfer} \label{app:preference_to_policy_transfer}
We next use $C_{\mathrm{comp}}(\pi^\star;\Pi)$ to transfer comparison-score error to policy KL divergence.

\begin{lemma}[Comparison-Score Error to Policy KL Divergence] \label{lem:subopt_score_error_to_policy_kl}
    Suppose that Assumption~\ref{ass:bounded_comparison_scores} holds and that $C_{\mathrm{comp}}(\pi^\star; \Pi) < \infty$.
    Then, for every $\pi \in \Pi$ sharing the response support of $\pi^\star$ for $\rho$-almost every prompt,
    \begin{equation} \label{eq:subopt_score_error_to_policy_kl}
        \mathbb{E}_{s \sim \rho} \left[D_{\mathrm{KL}}\left(\pi(\cdot \mid s) \middle\| \pi^\star(\cdot \mid s)\right)\right]
        \leq \frac{C_{\mathrm{comp}}(\pi^\star; \Pi)}{2\beta^2} \mathbb{E}_D\left[\|x_\pi(D) - x^\star(D)\|^2\right].
    \end{equation}
\end{lemma}

We next combine the comparison-score estimation bound with Lemma~\ref{lem:subopt_score_error_to_policy_kl} to control KL-regularized policy suboptimality.

\begin{theorem}[Regularized Suboptimality Bound] \label{thm:subopt_regularized}
    Under the conditions of Lemma~\ref{lem:relative_uniform_convergence}, let $\xi_{n_{\mathrm{opt}}}$ be as defined there.
    Suppose that the exact empirical LatentCP-DPO objective attains its maximum over $\Pi$, and let $\widehat{\pi}_{\mathrm{LCP}}$ be any maximizer.
    Assume that $C_{\mathrm{comp}}(\pi^\star;\Pi) < \infty$.
    Let $\pi^\dagger$ be a comparator such that $J(\pi)$ and $H(\pi)$ are finite for $\pi \in \{\pi^\star, \widehat{\pi}_{\mathrm{LCP}}, \pi^\dagger\}$.
    Conditional on the pilot and calibration samples, there is an event of probability at least $1-\delta$ over the optimization sample on which, provided that $\widehat{\pi}_{\mathrm{LCP}}$ shares the response support of $\pi^\star$ for $\rho$-almost every prompt, the following bound holds:
    \begin{equation} \label{eq:subopt_regularized_bound}
        \begin{aligned}
            \mathrm{SubOpt}_{\beta} (\widehat{\pi}_{\mathrm{LCP}},\pi^\dagger)
            \leq& \frac{C_{\mathrm{comp}}(\pi^\star; \Pi)}{\beta\kappa_{\mathrm{M}}} \min\left\{H_{\mathrm{M}}, \left(\sqrt{b_{\alpha, \gamma}} + \sqrt{\mathcal{E}_{\Pi}^{\mathrm{pess}} + \max\left\{\mathcal{E}_{\Pi}^{\mathrm{pess}}, \xi_{n_{\mathrm{opt}}}\right\}}\right)^2\right\}\\
            &- \beta \mathbb{E}_{s \sim \rho}\left[D_{\mathrm{KL}}\left(\pi^\dagger(\cdot \mid s) \middle\| \pi^\star(\cdot \mid s)\right)\right].
        \end{aligned}
    \end{equation}
\end{theorem}

To quantify the effect of conformal uncertainty on this bound, we next control $b_{\alpha, \gamma}$ and $\mathcal{E}_{\Pi}^{\mathrm{pess}}$ in terms of the calibrated prediction sets and the policy-class approximation error $\mathcal{E}_{\Pi}$.

\subsection{Effect of Conformal Uncertainty} \label{app:effect_of_conformal_uncertainty}
To quantify the effect of conformal prediction, we first analyze calibration-sample variability conditional only on the pilot sample.
For each realized calibration sample, define the conditional preference-outcome miscoverage by
\begin{equation*}
    \bar{\gamma}_{\mathrm{cal}} = \mathbb{E}_D\left[1 - p_\gamma\bigl(\boldsymbol{u}^\star(D), D\bigr)\right],
\end{equation*}
where the expectation is over the target context distribution, with the pilot and calibration samples held fixed.
Under model correctness, this quantity equals the probability that a target preference outcome falls outside $\widehat{\mathcal{C}}_\gamma(D)$, conditional on these samples.
The following lemma controls this miscoverage in terms of $\gamma$ and the calibration sample size.

\begin{lemma}[Conditional Calibration Miscoverage] \label{lem:subopt_conditional_calibration_miscoverage}
    Under Assumption~\ref{ass:subopt_sampling}, fix $\gamma \in (0, \alpha)$ independently of the calibration sample, and let $n_{\mathrm{cal}} \geq 1$.
    Set $j_\gamma = \lfloor(n_{\mathrm{cal}}+1) \gamma\rfloor$.
    For $\delta_{\mathrm{cal}} \in (0, 1)$, let $u_{n_{\mathrm{cal}},\gamma}(\delta_{\mathrm{cal}})=0$ if $j_\gamma=0$; otherwise, let it be the $(1-\delta_{\mathrm{cal}})$-quantile of $\operatorname{Beta}(j_\gamma, n_{\mathrm{cal}}+1-j_\gamma)$.
    Then, conditional on the pilot sample, with probability at least $1 - \delta_{\mathrm{cal}}$ over the calibration sample,
    \begin{equation} \label{eq:subopt_exact_calibration_envelope}
        \bar\gamma_{\mathrm{cal}}
        \leq u_{n_{\mathrm{cal}},\gamma}(\delta_{\mathrm{cal}})
        \leq \min\left\{1,2\gamma+
        \frac{4\log(1/\delta_{\mathrm{cal}})}{3n_{\mathrm{cal}}}\right\}.
    \end{equation}
    These bounds allow ties and infinite scores.
    If the score distribution conditional on the pilot sample is continuous and $j_\gamma \geq 1$, then $\bar{\gamma}_{\mathrm{cal}}$ has the $\operatorname{Beta}(j_\gamma, n_{\mathrm{cal}}+1-j_\gamma)$ distribution conditional on the pilot sample.
\end{lemma}

We next relate the miscoverage and informativeness of the conformal prediction sets to the benchmark bias and the policy-class approximation error.
Throughout the following analysis, we condition on the realized pilot and calibration samples.
Recall
\begin{equation*}
    \eta_\gamma
    = \mathbb{P}_D\left\{|\widehat{\mathcal{C}}_\gamma(D)| \neq 1\right\},
    \qquad
    \mathcal{E}_\Pi = \inf_{\pi\in\Pi} \mathbb{E}_D\left[\|x_\pi(D) - x^\star(D)\|^2\right].
\end{equation*}
Here, $\eta_\gamma$ is the probability of a non-singleton or empty conformal prediction set.
The quantity $\mathcal{E}_\Pi$ measures the policy class's mean squared approximation error relative to the true comparison score, whereas $\mathcal{E}_\Pi^{\mathrm{pess}}$ measures excess pessimistic risk relative to the benchmark $\bar{x}(D)$.

To retain partial ranking information, for PL feedback, write $R_k(y) = \{y_k, \ldots, y_K\}$.
Define the normalized set disagreement by
\begin{equation*}
    \omega_D=
    \begin{cases}
        \mathbb I\{|\widehat{\mathcal{C}}_\gamma(D)| \neq 1\}, & \mathrm{M} = \mathrm{BT},\\
        1,&\mathrm{M}=\mathrm{PL},\ \widehat{\mathcal{C}}_\gamma(D) = \varnothing,\\
        \frac{1}{K-1}\max_{y,z \in \widehat{\mathcal{C}}_\gamma(D)} \sum_{k=2}^K \mathbb{I}\{R_k(y) \neq R_k(z)\}, &\mathrm{M} = \mathrm{PL},\ \widehat{\mathcal{C}}_\gamma(D) \neq \varnothing.
    \end{cases}
\end{equation*}
Set
\begin{equation} \label{eq:subopt_structural_set_disagreement}
    r_\varepsilon(w) = \frac{2 \varepsilon_{\alpha,\gamma} w + (1 - \varepsilon_{\alpha,\gamma}) w^2}{1 + \varepsilon_{\alpha,\gamma}},
    \qquad
    \widetilde{\eta}_{\alpha,\gamma}=\mathbb E_D[r_\varepsilon(\omega_D)].
\end{equation}
Since $w^2 \leq r_\varepsilon(w) \leq w$ for $w \in [0,1]$ and $\omega_D \leq \mathbb{I}\{|\widehat{\mathcal{C}}_\gamma(D)| \neq 1\}$, we have $0 \leq \widetilde\eta_{\alpha,\gamma} \leq \eta_\gamma$, and $\widetilde\eta_{\alpha,\gamma} = \eta_\gamma$ for BT feedback.
For example, the PL set $\widehat{\mathcal{C}}_\gamma(D) = \{(1,2,3),(2,1,3)\}$ has $\omega_D=1/2$ and contributes $(1+3\varepsilon_{\alpha,\gamma})/(4(1+\varepsilon_{\alpha,\gamma}))$ rather than one.
This measure quantifies unresolved ranking structure; a common pairwise order alone does not identify the full preference distribution.

Let $t_D = 1-p_\gamma(\boldsymbol u^\star(D),D)$ and define
\begin{equation*}
    \Xi_{\alpha, \gamma}
    = \mathbb E_D\left[
        \left(\omega_D+(1-\omega_D)\max\{\varepsilon_{\alpha,\gamma},t_D\}\right)^2
    \right].
\end{equation*}
For singleton sets, the integrand is $\max\{\varepsilon_{\alpha,\gamma},t_D\}^2$, a squared total-variation envelope; for empty sets, it equals one.
For non-singleton PL sets, it instead uses a bound on the difference between expected loss gradients, retaining information shared by the rankings.
The following lemma connects this discrepancy envelope to the benchmark bias and pessimistic approximation error.

\begin{lemma}[Conformal Control of Benchmark and Approximation Bias] \label{lem:subopt_conformal_benchmark_bias}
    Under Assumption~\ref{ass:bounded_comparison_scores}, let $H_{\mathrm{M}}$, $\kappa_{\mathrm{M}}$, and $\Lambda_{\mathrm{M}}$ be defined as above, and set
    \begin{equation*}
        G_{\mathrm{M}} = \begin{cases}
            1, & \mathrm{M} = \mathrm{BT},\\
            \sqrt{2}(K-1), & \mathrm{M} = \mathrm{PL}.
        \end{cases}
    \end{equation*}
    Then
    \begin{equation}
        \label{eq:subopt_conformal_discrepancy_bound}
        \Xi_{\alpha, \gamma} \leq \min\left\{1, \varepsilon_{\alpha, \gamma}^{2} + (1 - \varepsilon_{\alpha,\gamma}^{2})(\widetilde\eta_{\alpha,\gamma}+\bar{\gamma}_{\mathrm{cal}})\right\}.
    \end{equation}
    Moreover, the benchmark bias satisfies
    \begin{equation}
        \label{eq:subopt_conformal_benchmark_kl_bias}
        b_{\alpha, \gamma} 
        \leq \min\left\{H_{\mathrm{M}}, \frac{G_{\mathrm{M}}^2}{2\kappa_{\mathrm{M}}}\Xi_{\alpha, \gamma}\right\},
    \end{equation}
    and the pessimistic approximation error satisfies
    \begin{equation} \label{eq:subopt_conformal_policy_approximation_bias}
        \mathcal{E}_\Pi^{\mathrm{pess}}
        \leq \frac{\Lambda_{\mathrm{M}}}{2}\mathcal{E}_\Pi + G_{\mathrm{M}} \sqrt{\Xi_{\alpha, \gamma}\mathcal{E}_\Pi} + \frac{G_{\mathrm{M}}^2}{2\kappa_{\mathrm{M}}} \Xi_{\alpha, \gamma}.
    \end{equation}
\end{lemma}

\subsection{Proof of the Main Upper Bounds} \label{app:subopt_proof_of_the_main_upper_bound}
Having established the auxiliary results, we now prove Theorem~\ref{thm:subopt_main_guarantees} and Corollary~\ref{cor:subopt_approximate_optimization}.

\begin{proof}[\bf Proof of Theorem~\ref{thm:subopt_main_guarantees}]
    Throughout the proof, we condition on the pilot sample.
    Applying Lemma~\ref{lem:subopt_conditional_calibration_miscoverage} with failure probability $\delta/2$, followed by (\ref{eq:subopt_conformal_discrepancy_bound}), gives
    \begin{equation} \label{eq:subopt_main_proof_conformal_bound}
        \begin{aligned}
            \Xi_{\alpha,\gamma}
            &\leq\min\left\{1,\,
            \varepsilon_{\alpha, \gamma}^2 + (1 - \varepsilon_{\alpha,\gamma}^{2})
            \left(\widetilde\eta_{\alpha,\gamma} + 2\gamma + \frac{4\log(2/\delta)}{3n_{\mathrm{cal}}}\right)\right\}\\
            &\leq 2\Psi_{\alpha,\gamma}(\delta),
        \end{aligned}
    \end{equation}
    with probability at least $1-\delta/2$ over the calibration sample.
    
    Conditional on the realized calibration sample, apply Lemma~\ref{lem:relative_uniform_convergence} with failure probability $\delta/2$ and the explicit localized radius in (\ref{eq:subopt_explicit_localized_radius}).
    The corresponding statistical term is
    \begin{equation*}
        \xi_{n_{\mathrm{opt}}}
        = \frac{H_{\mathrm{M}}}{n_{\mathrm{opt}}} \max\left\{1024C_{\mathrm{loc}}^2d_\Pi, 64\log(2/\delta)\right\}.
    \end{equation*}
    On this optimization-sample event, (\ref{eq:subopt_true_comparison_score_bound}) and $\mathcal{E}_{\Pi}^{\mathrm{pess}} + \max\{\mathcal{E}_{\Pi}^{\mathrm{pess}},\xi_{n_{\mathrm{opt}}}\} \leq 2\mathcal{E}_{\Pi}^{\mathrm{pess}} + \xi_{n_{\mathrm{opt}}}$ imply
    \begin{equation} \label{eq:subopt_main_intrinsic_score_bound}
        \mathbb{E}_D\left[\|x_{\widehat{\pi}_{\mathrm{LCP}}}(D) - x^\star(D)\|^2\right]
        \leq \frac{4}{\kappa_{\mathrm{M}}}b_{\alpha, \gamma} + \frac{8}{\kappa_{\mathrm{M}}}\mathcal{E}_\Pi^{\mathrm{pess}} + \frac{4}{\kappa_{\mathrm{M}}} \xi_{n_{\mathrm{opt}}}.
    \end{equation}
    Substituting the bias bounds from Lemma~\ref{lem:subopt_conformal_benchmark_bias} and applying $2uv \leq u^2 + v^2$ yields
    \begin{equation*}
        \begin{aligned}
            &\mathbb{E}_D\left[\|x_{\widehat{\pi}_{\mathrm{LCP}}}(D) - x^\star(D)\|^2\right]\\
            \leq& \frac{4\Lambda_{\mathrm{M}}}{\kappa_{\mathrm{M}}}\mathcal{E}_\Pi + \frac{8G_{\mathrm{M}}}{\kappa_{\mathrm{M}}}\sqrt{\Xi_{\alpha,\gamma}\mathcal{E}_\Pi} + \frac{6G_{\mathrm{M}}^2}{\kappa_{\mathrm{M}}^2}\Xi_{\alpha,\gamma} + \frac{4}{\kappa_{\mathrm{M}}}\xi_{n_{\mathrm{opt}}}\\
            \leq& \left(\frac{4\Lambda_{\mathrm{M}}}{\kappa_{\mathrm{M}}} + 4\right)\mathcal{E}_\Pi + \frac{10G_{\mathrm{M}}^2}{\kappa_{\mathrm{M}}^2}\Xi_{\alpha,\gamma} + \frac{4}{\kappa_{\mathrm{M}}}\xi_{n_{\mathrm{opt}}}.
        \end{aligned}
    \end{equation*}
    
    By the tower property and a union bound, the calibration and optimization-sample events hold jointly with probability at least $1 - \delta$, conditional on the pilot sample.
    On this joint event, (\ref{eq:subopt_main_proof_conformal_bound}) bounds the second term above by $20G_{\mathrm{M}}^2\Psi_{\alpha, \gamma}(\delta) / \kappa_{\mathrm{M}}^2$.
    Moreover, since $C_{\mathrm{loc}} \geq 1$,
    \begin{equation*}
        \frac{4}{\kappa_{\mathrm{M}}}\xi_{n_{\mathrm{opt}}}
        \leq \frac{4096H_{\mathrm{M}}C_{\mathrm{loc}}^2}{\kappa_{\mathrm{M}}} \frac{d_\Pi+\log(2/\delta)}{n_{\mathrm{opt}}}.
    \end{equation*}
    Thus, with
    \begin{equation*}
        C_{\mathrm{M}}
        = \max\left\{\frac{4\Lambda_{\mathrm{M}}}{\kappa_{\mathrm{M}}}+4, \frac{20G_{\mathrm{M}}^2}{\kappa_{\mathrm{M}}^2}, \frac{4096H_{\mathrm{M}}C_{\mathrm{loc}}^2}{\kappa_{\mathrm{M}}}\right\}
    \end{equation*}
    we obtain
    \begin{equation} \label{eq:subopt_main_proof_score_bound}
        \mathbb{E}_D\left[\|x_{\widehat{\pi}_{\mathrm{LCP}}}(D) - x^\star(D)\|^2\right]
        \leq C_{\mathrm{M}} \mathfrak{E}_{\alpha, \gamma}(\delta).
    \end{equation}
    Substituting the BT and PL values of $\Lambda_{\mathrm{M}}$, $\kappa_{\mathrm{M}}$, $G_{\mathrm{M}}$, and $H_{\mathrm{M}}$ gives the respective constants in (\ref{eq:model_specific_constant_bt}) and (\ref{eq:model_specific_constant_pl}).

    Finally, under the support and finiteness conditions of the theorem, (\ref{eq:subopt_beta_bound_proof_in_theorem}) and (\ref{eq:subopt_main_proof_score_bound}) give
    \begin{equation*}
        \begin{aligned}
            \mathrm{SubOpt}_{\beta}(\widehat{\pi}_{\mathrm{LCP}},\pi^\dagger)
            \leq& \frac{C_{\mathrm{M}}C_{\mathrm{comp}}(\pi^\star;\Pi)}{2\beta}\mathfrak{E}_{\alpha,\gamma}(\delta)\\
            & - \beta\mathbb{E}_{s\sim\rho}\left[D_{\mathrm{KL}}\left(\pi^\dagger(\cdot\mid s)\middle\|\pi^\star(\cdot\mid s)\right)\right],
        \end{aligned}
    \end{equation*}
    which is (\ref{eq:subopt_main_regularized}).
\end{proof}

\begin{proof}[\bf Proof of Corollary~\ref{cor:subopt_approximate_optimization}]
    Throughout, we condition on the pilot sample.
    Work on the joint calibration and optimization-sample event used in the proof of Theorem~\ref{thm:subopt_main_guarantees}, which has conditional probability at least $1 - \delta$.
    On this event, (\ref{eq:subopt_main_proof_conformal_bound}) holds, and (\ref{eq:subopt_relative_uniform_deviation}) holds simultaneously for all $\pi\in\Pi$ with
    \begin{equation*}
        \xi_{n_{\mathrm{opt}}}
        = \frac{H_{\mathrm{M}}}{n_{\mathrm{opt}}} \max\left\{1024C_{\mathrm{loc}}^2d_\Pi, 64\log(2/\delta)\right\}.
    \end{equation*}
    
    For $\pi \in \Pi$, write
    \begin{equation*}
        m_\pi = \mathbb{E}_D[e_\pi(D)],
        \qquad
        \widehat{m}_\pi = \frac{1}{n_{\mathrm{opt}}}\sum_{i=1}^{n_{\mathrm{opt}}}e_\pi(D_i^{\mathrm{opt}}).
    \end{equation*}
    Since
    \begin{equation*}
        \widehat{L}(\pi) - \widehat{m}_\pi
        = \frac{1}{n_{\mathrm{opt}}}\sum_{i=1}^{n_{\mathrm{opt}}}f_{D_i^{\mathrm{opt}}}(\bar{x}(D_i^{\mathrm{opt}}))
    \end{equation*}
    is independent of $\pi$, we have $\widehat{m}_{\widetilde\pi} = \inf_{\pi \in \Pi}\widehat{m}_\pi + \Delta_{\mathrm{opt}}$.
    Moreover, (\ref{eq:subopt_relative_uniform_deviation}) and $\max\{m_\pi,\xi_{n_{\mathrm{opt}}}\} \leq m_\pi + \xi_{n_{\mathrm{opt}}}$ imply, simultaneously for all $\pi\in\Pi$,
    \begin{equation*}
        m_\pi \leq \frac{3}{2} \widehat{m}_\pi + \frac{1}{2}\xi_{n_{\mathrm{opt}}},
        \qquad
        \widehat{m}_\pi \leq \frac{4}{3}m_\pi + \frac{1}{3}\xi_{n_{\mathrm{opt}}}.
    \end{equation*}
    Therefore,
    \begin{equation*}
        m_{\widetilde{\pi}}
        \leq \frac{3}{2}\widehat{m}_{\widetilde{\pi}} + \frac{1}{2}\xi_{n_{\mathrm{opt}}}
        \leq \frac{3}{2}\inf_{\pi\in\Pi}\widehat{m}_\pi + \frac{3}{2}\Delta_{\mathrm{opt}} + \frac{1}{2}\xi_{n_{\mathrm{opt}}}
        \leq 2\mathcal{E}_\Pi^{\mathrm{pess}} + \xi_{n_{\mathrm{opt}}} + \frac{3}{2}\Delta_{\mathrm{opt}}.
    \end{equation*}
    The passage to the infimum does not require it to be attained.
    
    Lemma~\ref{lem:pointwise_benchmark_quadratic_growth} gives $\mathbb{E}_D[\|x_{\widetilde\pi}(D) - \bar{x}(D)\|^2] \leq 2m_{\widetilde\pi}/\kappa_{\mathrm M}$.
    The benchmark-bias argument (\ref{eq:subopt_upper_bound_of_x_bar_and_x_star}) in the proof of Theorem~\ref{thm:subopt_regularized} gives $\mathbb{E}_D[\|\bar{x}(D)-x^\star(D)\|^2] \leq 2b_{\alpha,\gamma}/\kappa_{\mathrm M}$.
    Applying the squared triangle inequality yields
    \begin{equation*}
        \mathbb{E}_D \left[\|x_{\widetilde\pi}(D)-x^\star(D)\|^2\right]
        \leq \frac{4}{\kappa_{\mathrm{M}}}\left(b_{\alpha,\gamma} + m_{\widetilde\pi}\right)
        \leq \frac{4}{\kappa_{\mathrm{M}}}b_{\alpha,\gamma} + \frac{8}{\kappa_{\mathrm{M}}}\mathcal{E}_\Pi^{\mathrm{pess}} + \frac{4}{\kappa_{\mathrm{M}}}\xi_{n_{\mathrm{opt}}} + \frac{6}{\kappa_{\mathrm{M}}}\Delta_{\mathrm{opt}}.
    \end{equation*}
    By Lemma~\ref{lem:subopt_conformal_benchmark_bias}, (\ref{eq:subopt_main_proof_conformal_bound}), and the same constant calculation as in the proof of Theorem~\ref{thm:subopt_main_guarantees}, the first three terms are bounded by $C_{\mathrm{M}}\mathfrak{E}_{\alpha,\gamma}(\delta)$.
    Thus,
    \begin{equation*}
        \mathbb{E}_D \left[\|x_{\widetilde\pi}(D)-x^\star(D)\|^2\right]
        \leq C_{\mathrm{M}}\mathfrak{E}_{\alpha,\gamma}(\delta) + \frac{6}{\kappa_{\mathrm{M}}}\Delta_{\mathrm{opt}}.
    \end{equation*}
    Finally, under the stated support and finiteness conditions, the KL representation of regularized suboptimality derived in the proof of Theorem~\ref{thm:subopt_regularized} gives
    \begin{equation*}
        \mathrm{SubOpt}_{\beta}(\widetilde{\pi},\pi^\dagger) 
        = \beta\mathbb{E}_{s\sim\rho}\left[D_{\mathrm{KL}}\left(\widetilde{\pi}(\cdot\mid s)\middle\|\pi^\star(\cdot\mid s)\right)\right] - \beta\mathbb{E}_{s\sim\rho}\!\left[D_{\mathrm{KL}}\left(\pi^\dagger(\cdot\mid s)\middle\|\pi^\star(\cdot\mid s)\right)\right].
    \end{equation*}
    Applying Lemma~\ref{lem:subopt_score_error_to_policy_kl} to the first term and then using the preceding score-error bound proves (\ref{eq:subopt_approximate_optimization_bound}).
    The event is uniform over $\Pi$, so the conclusion holds simultaneously for every measurable, possibly data-dependent output and every comparator satisfying the stated conditions.
\end{proof}

\subsection{Proofs of Auxiliary Results} \label{app:subopt_proofs_upper_bounds}
We provide the proofs of the auxiliary results below.

\begin{proof}[\bf Proof of Lemma~\ref{lem:subopt_common_policy_classes}]
    Throughout the proof, we condition on the pilot and calibration samples and restrict all functions to $\mathcal{D}_0$.
    For a comparison-score function $v$, define
    \begin{equation*}
        \|v\|_{\infty, 2} = \sup_{D \in \mathcal{D}_0} \|v(D)\|,
    \end{equation*}
    and use $\|\cdot\|_\infty$ for the uniform norm of scalar-valued functions on $\mathcal{D}_0$.
    For a function class $\mathcal{F}$, let $\mathcal{N}(\eta, \mathcal{F}, \|\cdot\|)$ denote its $\eta$-covering number: the smallest number of balls of radius $\eta$, with centers in $\mathcal{F}$, needed to cover $\mathcal{F}$.
    
    We first establish a uniform covering bound for the comparison-score class.
    Under the first condition, fix $\pi_0 \in \Pi$ and define
    \begin{equation*}
        V = \operatorname{span}\{x_\pi - x_{\pi_0}: \pi \in \Pi\}.
    \end{equation*}
    Equipped with $\|\cdot\|_{\infty, 2}$, this is a normed space of bounded functions with dimension at most $p$.
    The translated class $\{x_\pi - x_{\pi_0}: \pi \in \Pi\}$ is contained in the ball centered at zero with radius
    \begin{equation*}
        R_{\mathrm{M}} = \begin{cases}
            2B, & \mathrm{M} = \mathrm{BT},\\
            2B\sqrt{K}, & \mathrm{M} = \mathrm{PL}.
        \end{cases}
    \end{equation*}
    If $\dim V = 0$, the score class is a singleton and its covering number is one.
    Otherwise, consider any finite collection of points in the translated class with pairwise distances greater than $\eta$. 
    The balls of radius $\eta/2$ centered at these points are disjoint and contained in the ball of radius $R_{\mathrm M} + \eta/2$.
    Comparing volumes in $V$ bounds the cardinality of this collection by $(1 + 2R_{\mathrm{M}}/\eta)^{\dim V}$.
    A maximal such collection therefore exists and forms an $\eta$-cover.
    Since translation preserves distances and $\dim V \leq p$, we obtain
    \begin{equation}\label{eq:subopt_score_covering_bound}
        \mathcal{N}\left(\eta, \{x_\pi: \pi \in \Pi\}, \|\cdot\|_{\infty,2}\right)
        \leq \left(1 + \frac{A_x}{\eta}\right)^p,
        \qquad \eta > 0,
    \end{equation}
    with $A_x = 2R_{\mathrm{M}}$.

    Under the second condition, suppose first that $L_x > 0$.
    An internal $\eta/L_x$-net of $\mathcal{W}$ induces an $\eta$-net of the score class.
    Since $\mathcal{W}$ is contained in a Euclidean ball of radius $\operatorname{diam}(\mathcal{W})$, the same volumetric argument in $\mathbb{R}^p$ yields (\ref{eq:subopt_score_covering_bound}) with $A_x = 2L_x \operatorname{diam}(\mathcal{W})$.
    If $L_x = 0$, all comparison-score functions coincide, and the bound holds with $A_x = 0$.

    We next transfer this covering bound to the excess-loss class.
    The outcome-wise negative log-likelihood is uniformly Lipschitz in the comparison score. 
    Specifically,
    \begin{equation*}
        |\ell_x(y,D)-\ell_z(y,D)| \leq G_{\mathrm M}\|x-z\|,
    \end{equation*}
    where $G_{\mathrm{BT}} = 1$ and $G_{\mathrm{PL}} = \sqrt{2}(K-1)$.
    For BT, the derivative of the negative log-likelihood has absolute value at most one.
    For PL, each stagewise gradient is the difference between a probability vector over the remaining candidates and a standard basis vector, so its Euclidean norm is at most $\sqrt{2}$.
    Summing over the $K-1$ nontrivial stages gives the stated bound.
    Hence, for each fixed feasible $\boldsymbol{u}$, taking expectation under $P_{\mathrm{M}, \boldsymbol{u}}(\cdot \mid D)$ preserves this Lipschitz bound.
    The uncertainty set does not depend on the candidate score.
    Taking suprema over the same set for $x$ and $z$, and applying the resulting inequality in both directions, therefore yields $|f_D(x) - f_D(z)| \leq G_{\mathrm M} \|x - z\|$.
    Since the common benchmark term cancels in $e_\pi(D) - e_{\pi'}(D)$, it follows that
    \begin{equation*}
        |e_\pi(D) - e_{\pi'}(D)| \leq G_{\mathrm M} \|x_\pi(D) - x_{\pi'}(D)\|.
    \end{equation*}
    Consequently,
    \begin{equation*}
        \|e_\pi - e_{\pi'}\|_\infty
        \leq G_{\mathrm{M}} \|x_\pi - x_{\pi'}\|_{\infty,2}.
    \end{equation*}

    The bound $0 \leq e_\pi \leq H_{\mathrm{M}}$ implies that the mean constraint defining $\mathcal{H}_1$ is automatically satisfied. 
    Thus,
    \begin{equation*}
        \mathcal{H}_1
        = \left\{\frac{\lambda e_\pi}{H_{\mathrm M}}: \pi \in \Pi, \lambda \in [0,1]\right\}.
    \end{equation*}
    Moreover, for $\lambda, \lambda' \in [0,1]$,
    \begin{equation} \label{eq:subopt_rescaled_loss_lipschitz}
        \left\|\frac{\lambda e_\pi}{H_{\mathrm{M}}} - \frac{\lambda'e_{\pi'}}{H_{\mathrm{M}}}\right\|_\infty
        \leq \frac{\lambda}{H_{\mathrm M}} \|e_\pi-e_{\pi'}\|_\infty + \frac{|\lambda-\lambda'|}{H_{\mathrm M}} \|e_{\pi'}\|_\infty
        \leq \frac{G_{\mathrm{M}}}{H_{\mathrm{M}}} \|x_\pi-x_{\pi'}\|_{\infty, 2} + |\lambda - \lambda'|.
    \end{equation}

    Fix $0 < \eta \leq 1$.
    By (\ref{eq:subopt_score_covering_bound}), the score class admits a cover of radius $H_{\mathrm{M}} \eta / (2G_{\mathrm{M}})$, with centers $x_{\pi_1}, \ldots, x_{\pi_{M_x}}$ and cardinality
    \begin{equation*}
        M_x \leq \left(1 + \frac{2G_{\mathrm M}A_x}{H_{\mathrm M}\eta}\right)^p.
    \end{equation*}
    For the scalar parameter, consider the grid
    \begin{equation*}
        \lambda_k = \frac{k\eta}{2},
        \qquad
        k = 0, \ldots, \left\lfloor\frac{2}{\eta}\right\rfloor.
    \end{equation*}
    All grid points belong to $[0, 1]$, and every $\lambda \in [0, 1]$ lies within $\eta/2$ of a grid point.
    The grid contains at most $1 + 2/\eta$ points.
    
    Pairing the score representatives with these grid points gives the functions $\lambda_k e_{\pi_j} / H_{\mathrm{M}} \in \mathcal{H}_1$.
    For every $\lambda e_\pi / H_{\mathrm M} \in \mathcal{H}_1$, choose $j$ and $k$ from the respective covers. 
    By (\ref{eq:subopt_rescaled_loss_lipschitz}),
    \begin{equation*}
        \left\|\frac{\lambda e_\pi}{H_{\mathrm M}} - \frac{\lambda_k e_{\pi_j}}{H_{\mathrm M}}\right\|_\infty
        \leq \frac{G_{\mathrm M}}{H_{\mathrm M}} \frac{H_{\mathrm M}\eta}{2G_{\mathrm M}} + \frac{\eta}{2}
        = \eta.
    \end{equation*}
    Consequently,
    \begin{equation*}
        \mathcal{N}(\eta, \mathcal{H}_1, \|\cdot\|_\infty)
        \leq \left(1 + \frac{2G_{\mathrm{M}}A_x}{H_{\mathrm{M}}\eta}\right)^p \left(1 + \frac{2}{\eta}\right).
    \end{equation*}
    Set
    \begin{equation*}
        A = \max\left\{3, 1+\frac{2G_{\mathrm{M}}A_x}{H_{\mathrm{M}}}\right\}.
    \end{equation*}
    Since $0 < \eta \leq 1$ and $p \geq 1$, we obtain
    \begin{equation*}
        \log\mathcal{N}(\eta, \mathcal{H}_1, \|\cdot\|_\infty)
        \leq (p + 1) \log(A/\eta)
        \leq 2p \log(A/\eta).
    \end{equation*}
    The constant $A$ depends only on the fixed model and class bounds.
    
    Fix $\zeta > 0$.
    Since $\mathcal{H}_\zeta \subseteq \mathcal{H}_1$, an $\eta/2$-cover of $\mathcal{H}_1$ induces an internal $\eta$-cover of $\mathcal{H}_\zeta$: discard balls that do not intersect $\mathcal{H}_\zeta$, and choose one member of $\mathcal{H}_\zeta$ from each remaining ball as a new center.
    The triangle inequality shows that doubling the radius preserves coverage.
    For $\eta_j = 2^{-j}$, $j \geq 1$, there exist finite $\eta_j$-nets $\mathcal{T}_j \subseteq \mathcal{H}_\zeta$ satisfying
    \begin{equation*}
        \log|\mathcal{T}_j| 
        \leq 2p\log\left(\frac{2A}{\eta_j}\right)
        = 2p \bigl[\log(2A) + j\log2\bigr]
        \leq C_1 p(j+1),
    \end{equation*}
    where $C_1 = 2 \max\{\log(2A), \log 2\}$.
    Set $\eta_0 = 1$ and $\mathcal{T}_0 = \{0\}$.
    Because $0 \in \mathcal{H}_\zeta$ and every $h \in \mathcal{H}_\zeta$ takes values in $[0,1]$, $\mathcal{T}_0$ is an internal $\eta_0$-net.

    For any $h, g \in \mathcal{H}_\zeta$, nonnegativity and localization give
    \begin{equation*}
        \mathbb{E}_D[(h(D) - g(D))^2]
        \leq \|h - g\|_\infty \mathbb{E}_D[h(D) + g(D)]
        \leq 2 \zeta \|h - g\|_\infty.
    \end{equation*}
    Define the sets of nearby consecutive-net pairs by
    \begin{equation*}
        \mathcal{A}_j 
        = \left\{(h, g) \in \mathcal{T}_j \times \mathcal{T}_{j-1}: \|h-g\|_\infty \leq 3\eta_j\right\}.
    \end{equation*}
    These sets are nonempty, as can be seen by choosing consecutive-net approximants to the zero function.
    Moreover,
    \begin{equation*}
        \log 2|\mathcal{A}_j|
        \leq \log 2 + \log|\mathcal{T}_j| + \log|\mathcal{T}_{j - 1}| 
        \leq \log 2 + C_1 p(j+1) + C_1 pj
        \leq C_2 p(j+1),
    \end{equation*}
    where $C_2 = 2 C_1 + \log 2$.
    For $j = 1$, the same bound holds because $|\mathcal{T}_0|=1$.

    For every $(h, g) \in \mathcal{A}_j$, the random variables $\epsilon_i(h - g)(D_i^{\mathrm{opt}})$ are independent and centered, have absolute value at most $3\eta_j$, and satisfy
    \begin{equation*}
        \mathbb{E}\left[\epsilon_i^2 (h - g)^2(D_i^{\mathrm{opt}})\right]
        \leq 6 \zeta \eta_j.
    \end{equation*}
    Bernstein's maximal inequality therefore gives
    \begin{equation*}
        \begin{aligned}
            \mathbb{E}\max_{(h, g) \in \mathcal{A}_j}\left|\frac{1}{n_{\mathrm{opt}}} \sum_{i=1}^{n_{\mathrm{opt}}} \epsilon_i(h-g)(D_i^{\mathrm{opt}})\right|
            \leq& \sqrt{\frac{12 \zeta \eta_j \log(2|\mathcal{A}_j|)}{n_{\mathrm{opt}}}} + \frac{\eta_j\log(2|\mathcal{A}_j|)}{n_{\mathrm{opt}}}\\
            &\leq C \left(\sqrt{\frac{p \zeta \eta_j(j+1)}{n_{\mathrm{opt}}}} + \frac{p\eta_j(j+1)}{n_{\mathrm{opt}}}\right).
        \end{aligned}
    \end{equation*}
    Here, $C$ depends only on $C_2$, and the expectation is over both the optimization sample and the Rademacher signs.

    Finally, for each $h \in \mathcal{H}_\zeta$, choose $h^j \in \mathcal{T}_j$ with $\|h - h^j\|_\infty \leq \eta_j$, and set $h^0 = 0$.
    Then $(h^j, h^{j-1}) \in \mathcal{A}_j$.
    For every $J \geq 1$, the telescoping identity
    \begin{equation*}
        h = (h - h^J) + \sum_{j=1}^{J} (h^j - h^{j-1})
    \end{equation*}
    and $|\epsilon_i|=1$ imply that the remainder contributes at most $\eta_J$ to the absolute Rademacher average.
    Taking the supremum over $h \in \mathcal{H}_\zeta$ and then expectation yields
    \begin{equation*}
        \mathfrak{R}_{n_{\mathrm{opt}}}(\zeta)
        \leq \eta_J + \sum_{j=1}^{J}\mathbb{E} \max_{(h, g) \in \mathcal{A}_j}\left|\frac{1}{n_{\mathrm{opt}}} \sum_{i=1}^{n_{\mathrm{opt}}} \epsilon_i(h - g)(D_i^{\mathrm{opt}})\right|.
    \end{equation*}
    Substituting the preceding maximal bound and letting $J \to \infty$, we obtain
    \begin{equation*}
        \begin{aligned}
            \mathfrak{R}_{n_{\mathrm{opt}}}(\zeta)
            &\leq C \sqrt{\frac{p\zeta}{n_{\mathrm{opt}}}}\sum_{j=1}^{\infty}\sqrt{2^{-j}(j+1)} + \frac{Cp}{n_{\mathrm{opt}}} \sum_{j=1}^{\infty}2^{-j}(j+1)\\
            &\leq C_{\mathrm{loc}}\left(\sqrt{\frac{p\zeta}{n_{\mathrm{opt}}}} + \frac{p}{n_{\mathrm{opt}}}\right).
        \end{aligned}
    \end{equation*}
    Both series converge.
    Since $C_1$ and $C_2$ depend only on the fixed model bounds and, in the second case, $L_x\operatorname{diam}(\mathcal{W})$, the resulting $C_{\mathrm{loc}}$ is independent of $p$, $n_{\mathrm{opt}}$, and $\zeta$.
    Increasing it if necessary ensures $C_{\mathrm{loc}}\geq 1$ and completes the proof.
\end{proof}

\begin{proof}[\bf Proof of Lemma~\ref{lem:uniform_boundedness_pessimistic_loss}]
    Fix a target context $D$ for which the stated bounds hold, and let $x \in \mathcal{X}(D)$ and $y \in \mathcal{Y}(D)$.
    For BT feedback, $y \in \{0, 1\}$ and $|x| \leq B$ imply
    \begin{equation*}
        \ell_x(y, D)
        = \log\left(1 + \exp(-(2y-1) x)\right)
        \leq \log(1 + e^B).
    \end{equation*}
    For PL feedback, let $y = (y_1, \ldots, y_K)$ be a complete ranking.
    Since the comparison scores are centered, $\sum_{k=1}^{K-1} x^{y_k} = -x^{y_K}$.
    Consequently,
    \begin{equation*}
        \begin{aligned}
            \ell_x(y,D)
            = \sum_{k=1}^{K-1} \left[\log\left(\sum_{\ell=k}^{K} e^{x^{y_\ell}}\right) - x^{y_k}\right]
            = \sum_{k=1}^{K-1} \log\left(\sum_{\ell=k}^{K} e^{x^{y_\ell}}\right) + x^{y_K}\\
            \leq \sum_{k=1}^{K-1} \left[\log(K - k + 1) + B\right] + B
            = \log(K!) + KB
            \leq H_{\mathrm{PL}}.
        \end{aligned}
    \end{equation*}
    In both cases, $\ell_x(y,D) \geq 0$ because $p_{\mathrm{M}, x}(y \mid D) \leq 1$.
    Taking expectations under any $P_{\mathrm{M}, \boldsymbol{u}}(\cdot \mid D)$ preserves these bounds.
    The empty-set convention ensures that $\widehat{\mathcal{U}}_\alpha(D; \gamma)$ is nonempty; hence taking the supremum over this set gives $0 \leq f_D(x) \leq H_{\mathrm{M}}$.
\end{proof}

\begin{proof}[\bf Proof of Lemma~\ref{lem:pointwise_benchmark_quadratic_growth}]
    Fix a target context $D$ satisfying the stated assumptions.
    The argument in Lemma~\ref{lem:uniform_boundedness_pessimistic_loss} applies to every $x \in \mathcal{X}(D)$ and yields $0 \leq f_D(x)\leq H_{\mathrm{M}}$ throughout this domain.
    
    We first establish strong convexity.
    For BT feedback,
    \begin{equation*}
        \ell_x(y,D) 
        = -y \log\sigma(x) - (1-y)\log\sigma(-x) 
        = \log(1 + e^x) - yx.
    \end{equation*}
    Consequently,
    \begin{equation*}
        \frac{\partial^2}{\partial x^2} \ell_x(y, D)
        = \sigma(x) \sigma(-x)
        \geq \sigma(B)\sigma(-B)
        = \kappa_{\mathrm{BT}},
        \qquad 
        x \in [-B, B].
    \end{equation*}
    For PL feedback, let $\boldsymbol{\omega} = \operatorname{softmax}(x)$.
    The first-stage contribution to the Hessian is $\operatorname{diag}(\boldsymbol{\omega}) - \boldsymbol{\omega} \boldsymbol{\omega}^\top$.
    Each remaining stage is a log-sum-exp function minus a linear term and therefore has a positive semidefinite Hessian.
    Hence,
    \begin{equation*}
        \nabla_x^2 \ell_x(y,D) \succeq \operatorname{diag}(\boldsymbol{\omega}) -\boldsymbol{\omega} \boldsymbol{\omega}^\top.
    \end{equation*}
    Moreover, since $\|x\|_\infty \leq B$,
    \begin{equation*}
        \omega^j 
        = \frac{e^{x^j}}{\sum_{i=1}^{K} e^{x^i}}
        \geq \frac{e^{-B}}{\sum_{i=1}^{K} e^{B}}
        \geq \frac{e^{-2B}}{K},
        \qquad 
        j = 1, \ldots, K.
    \end{equation*}
    Fix $\boldsymbol{h} \in \boldsymbol{1}_K^\perp$.
    Using $\sum_{j=1}^{K} \omega^j = 1$, we obtain, for any $c \in \mathbb{R}$,
    \begin{equation*}
        \sum_{j=1}^{K} \omega^j  (h^j - c)^2
        = \sum_{j=1}^{K}\omega^j  (h^j)^2 - \left(\sum_{j=1}^{K} \omega^j  h^j\right)^2 + \left(c - \sum_{j=1}^{K} \omega^j  h^j\right)^2.
    \end{equation*}
    Thus, the minimum over $c$ is attained at $c = \sum_{j=1}^{K} \omega^j  h^j$, giving
    \begin{equation*}
        \min_{c \in \mathbb{R}} \sum_{j=1}^{K} \omega^j  (h^j - c)^2
        = \sum_{j=1}^{K} \omega^j  (h^j)^2 - \left(\sum_{j=1}^{K} \omega^j  h^j\right)^2
        = \boldsymbol{h}^\top \left[\operatorname{diag}(\boldsymbol{\omega}) - \boldsymbol{\omega} \boldsymbol{\omega}^\top\right] \boldsymbol{h}.
    \end{equation*}
    Similarly, since $\sum_{j=1}^{K} h^j = 0$,
    \begin{equation*}
        \sum_{j=1}^{K} (h^j - c)^2 
        = \|\boldsymbol{h}\|^2 + Kc^2,
        \quad \Rightarrow \quad
        \min_{c \in \mathbb{R}} \sum_{j=1}^{K} (h^j - c)^2
        = \|\boldsymbol{h}\|^2.
    \end{equation*}
    Combining these identities with the lower bound on $\omega^j $ yields
    \begin{equation*}
        \begin{aligned}
            \boldsymbol{h}^\top \nabla_x^2 \ell_x(y,D) \boldsymbol{h}
            \geq \boldsymbol{h}^\top \left[\operatorname{diag}(\boldsymbol{\omega}) - \boldsymbol{\omega} \boldsymbol{\omega}^\top\right] \boldsymbol{h}
            = \min_{c \in \mathbb{R}} \sum_{j=1}^{K} \omega^j (h^j - c)^2\\
            \geq \frac{e^{-2B}}{K} \min_{c \in \mathbb{R}} \sum_{j=1}^{K} (h^j - c)^2
            = \frac{e^{-2B}}{K} \|\boldsymbol{h}\|^2 
            = \kappa_{\mathrm{PL}} \|\boldsymbol{h}\|^2.
        \end{aligned}
    \end{equation*}
    The domain $\mathcal{X}(D)$ is convex, and in the PL case the difference between any two points in this domain belongs to $\boldsymbol{1}_K^\perp$.
    The preceding curvature bounds therefore establish that $\ell_x(y,D)$ is $\kappa_{\mathrm{M}}$-strongly convex on $\mathcal{X}(D)$ for any $\mathrm{M} \in \{\mathrm{BT}, \mathrm{PL}\}$ and every outcome $y$.

    For fixed $D$, the latent uncertainty set and its indexed preference distributions do not depend on $x$.
    For each such distribution, taking the expectation preserves convexity of $x \mapsto \ell_x(y,D) - \kappa_{\mathrm{M}} \|x\|^2/2$.
    Taking the supremum over the same uncertainty set also preserves convexity.
    Consequently, $x \mapsto f_D(x) - \kappa_{\mathrm{M}} \|x\|^2/2$ is convex, so $f_D$ is $\kappa_{\mathrm{M}}$-strongly convex on $\mathcal{X}(D)$.

    We next establish continuity.
    Because the suprema defining $f_D(x)$ and $f_D(z)$ range over the same set, for any $x, z \in \mathcal{X}(D)$,
    \begin{equation*}
        \begin{aligned}
            |f_D(x) - f_D(z)|
            &\leq \sup_{\boldsymbol{u} \in \widehat{\mathcal{U}}_\alpha(D; \gamma)} \mathbb{E}_{y \sim P_{\mathrm{M}, \boldsymbol{u}}(\cdot \mid D)} \left[|\ell_x(y,D) - \ell_z(y,D)|\right]\\
            &\leq \max_{y \in \mathcal{Y}(D)} |\ell_x(y,D)-\ell_z(y,D)|.
        \end{aligned}
    \end{equation*}
    Since $\mathcal{Y}(D)$ is finite and each outcome-wise loss is continuous, the right-hand side tends to zero as $z \to x$.
    Thus, $f_D$ is continuous.
    Continuity and compactness of the nonempty domain $\mathcal{X}(D)$ guarantee that a minimizer exists, and strong convexity guarantees its uniqueness.

    Finally, fix $x\in\mathcal{X}(D)$ and $t\in(0,1)$.
    Convexity of $\mathcal{X}(D)$ ensures that $(1 - t) \bar{x}(D) + t x$ remains in the same domain.
    By optimality of $\bar{x}(D)$ and strong convexity,
    \begin{equation*}
        \begin{aligned}
            f_D(\bar{x}(D))
            &\leq f_D\bigl((1-t)\bar{x}(D) + tx\bigr)\\
            &\leq (1-t)f_D(\bar{x}(D)) + tf_D(x) - \frac{\kappa_{\mathrm{M}}}{2}t(1-t)\|x-\bar{x}(D)\|^2.
        \end{aligned}
    \end{equation*}
    Rearranging and dividing by $t>0$ gives
    \begin{equation*}
        f_D(x) - f_D(\bar{x}(D))
        \geq \frac{\kappa_{\mathrm{M}}}{2} (1-t)\|x-\bar{x}(D)\|^2.
    \end{equation*}
    Letting $t \downarrow 0$ proves the quadratic lower bound in (\ref{eq:subopt_pointwise_quadratic_growth}).
    The upper bound follows from $f_D(x) \leq H_{\mathrm{M}}$ and $f_D(\bar{x}(D)) \geq 0$.
\end{proof}

\begin{proof}[\bf Proof of Lemma~\ref{lem:relative_uniform_convergence}]
    Throughout the proof, all probabilities and expectations are conditional on the pilot and calibration samples.
    Define
    \begin{equation*}
        Z_{\zeta_{n_{\mathrm{opt}}}} = \sup_{h \in \mathcal{H}_{\zeta_{n_{\mathrm{opt}}}}}\left|\frac{1}{n_{\mathrm{opt}}} \sum_{i=1}^{n_{\mathrm{opt}}} h(D_i^{\mathrm{opt}}) - \mathbb{E}_D[h(D)]\right|.
    \end{equation*}
    
    We first bound its expectation by symmetrization.
    Let $D_1', \ldots, D_{n_{\mathrm{opt}}}'$ be an independent copy of the optimization sample.
    Write $\mathbb{E}_{\mathrm{opt}}$, $\mathbb{E}'$, and $\mathbb{E}_\epsilon$ for expectations over the original sample, its independent copy, and the Rademacher signs, respectively, with the signs independent of both samples.
    For every $h \in \mathcal{H}_{\zeta_{n_{\mathrm{opt}}}}$,
    \begin{equation*}
        \mathbb{E}_D[h(D)]
        = \mathbb{E}'\left[\frac{1}{n_{\mathrm{opt}}} \sum_{i=1}^{n_{\mathrm{opt}}} h(D_i')\right].
    \end{equation*}
    Thus, with all suprema in the following display taken over $h \in \mathcal{H}_{\zeta_{n_{\mathrm{opt}}}}$,
    \begin{equation*}
        \begin{aligned}
            \mathbb{E}[Z_{\zeta_{n_{\mathrm{opt}}}}] 
            &= \mathbb{E}_{\mathrm{opt}} \sup_{h \in \mathcal{H}_{\zeta_{n_{\mathrm{opt}}}}} \left|\mathbb{E}'\left[\frac{1}{n_{\mathrm{opt}}} \sum_i \bigl(h(D_i^{\mathrm{opt}})-h(D_i')\bigr)\right]\right|\\
            &\leq \mathbb{E}_{\mathrm{opt}} \mathbb{E}' \sup_{h \in \mathcal{H}_{\zeta_{n_{\mathrm{opt}}}}}\left|\frac{1}{n_{\mathrm{opt}}}\sum_i \bigl(h(D_i^{\mathrm{opt}})-h(D_i')\bigr)\right|\\
            &= \mathbb{E}_{\mathrm{opt}} \mathbb{E}' \mathbb{E}_\epsilon \sup_{h \in \mathcal{H}_{\zeta_{n_{\mathrm{opt}}}}} \left|\frac{1}{n_{\mathrm{opt}}}\sum_i \epsilon_i \bigl(h(D_i^{\mathrm{opt}})-h(D_i')\bigr)\right|\\
            &\leq \mathbb{E}_{\mathrm{opt}}\mathbb{E}_\epsilon \sup_{h \in \mathcal{H}_{\zeta_{n_{\mathrm{opt}}}}}\left|\frac{1}{n_{\mathrm{opt}}}\sum_i \epsilon_i h(D_i^{\mathrm{opt}})\right|
            + \mathbb{E}'\mathbb{E}_\epsilon \sup_{h \in \mathcal{H}_{\zeta_{n_{\mathrm{opt}}}}}\left|\frac{1}{n_{\mathrm{opt}}}\sum_i \epsilon_i h(D_i')\right|\\
            &= 2 \mathfrak{R}_{n_{\mathrm{opt}}}(\zeta_{n_{\mathrm{opt}}}).
        \end{aligned}
    \end{equation*}
    The first inequality follows from Jensen's inequality.
    The equality introducing the signs follows from invariance of the joint sample distribution under independent exchanges of the pairs $(D_i^{\mathrm{opt}}, D_i')$.
    The next inequality is the triangle inequality.
    The final line uses the identical distributions of the two samples and (\ref{eq:subopt_rademacher_complexity}).

    We next apply Bousquet's inequality \citep[Theorem A.1]{bartlett2005local} to the centered signed class
    \begin{equation*}
        \left\{\pm\left(h(D) - \mathbb{E}_D[h(D)]\right): h \in \mathcal{H}_{\zeta_{n_{\mathrm{opt}}}}\right\}.
    \end{equation*}
    Every member of this class has mean zero, is bounded in absolute value by one, and has variance at most $\zeta_{n_{\mathrm{opt}}}$.
    Conditional on the pilot and calibration samples, this class is fixed and the optimization contexts remain i.i.d.
    Since its supremum of empirical sums is $n_{\mathrm{opt}} Z_{\zeta_{n_{\mathrm{opt}}}}$,
    Bousquet's inequality, divided by $n_{\mathrm{opt}}$, gives, with conditional probability at least $1-\delta$,
    \begin{equation*}
        \begin{aligned}
            Z_{\zeta_{n_{\mathrm{opt}}}} 
            \leq \mathbb{E}[Z_{\zeta_{n_{\mathrm{opt}}}}] + \sqrt{\frac{2 \bigl(\zeta_{n_{\mathrm{opt}}} + 2\mathbb{E}[Z_{\zeta_{n_{\mathrm{opt}}}}]\bigr) \log(1/\delta)}{n_{\mathrm{opt}}}} + \frac{\log(1/\delta)}{3n_{\mathrm{opt}}}\\
            \leq \left(\frac{1}{8} + \sqrt{\frac{5}{128}} + \frac{1}{192}\right)\zeta_{n_{\mathrm{opt}}}
            < \frac{\zeta_{n_{\mathrm{opt}}}}{3},
        \end{aligned}
    \end{equation*}
    where the second inequality uses the preceding expectation bound and (\ref{eq:subopt_localized_radius}).

    Work on this event.
    For each $\pi \in \Pi$, set
    \begin{equation*}
        \lambda_\pi = \frac{\xi_{n_{\mathrm{opt}}}}{\max\left\{\mathbb{E}_D[e_\pi(D)], \xi_{n_{\mathrm{opt}}}\right\}}.
    \end{equation*}
    Then $0 < \lambda_\pi \leq 1$, and $\lambda_\pi\mathbb{E}_D[e_\pi(D)]/H_{\mathrm{M}} \leq \zeta_{n_{\mathrm{opt}}}$.
    Hence, $\lambda_\pi e_\pi / H_{\mathrm{M}} \in \mathcal{H}_{\zeta_{n_{\mathrm{opt}}}}$.
    By the definition of $Z_{\zeta_{n_{\mathrm{opt}}}}$, simultaneously for all $\pi\in\Pi$,
    \begin{equation*}
        \left|\frac{1}{n_{\mathrm{opt}}} \sum_{i=1}^{n_{\mathrm{opt}}} e_\pi(D_i^{\mathrm{opt}}) - \mathbb{E}_D[e_\pi(D)]\right|
        \leq \frac{H_{\mathrm{M}} Z_{\zeta_{n_{\mathrm{opt}}}}}{\lambda_\pi}
        \leq \frac{1}{3} \max\left\{\mathbb{E}_D[e_\pi(D)], \xi_{n_{\mathrm{opt}}}\right\}.
    \end{equation*}
    This proves (\ref{eq:subopt_relative_uniform_deviation}).

    Subtracting the policy-independent term $f_D(\bar{x}(D))$ does not affect empirical minimization.
    Thus, $\widehat{\pi}_{\mathrm{LCP}}$ also minimizes the empirical mean of $e_\pi$.
    For every $\pi \in \Pi$, the uniform deviation bound and empirical optimality therefore imply
    \begin{equation*}
        \begin{aligned}
            \mathbb{E}_D[e_{\widehat{\pi}_{\mathrm{LCP}}}(D)]
            &\leq \frac{1}{n_{\mathrm{opt}}} \sum_{i=1}^{n_{\mathrm{opt}}} e_{\widehat{\pi}_{\mathrm{LCP}}}(D_i^{\mathrm{opt}}) + \frac{1}{3}\max\left\{\mathbb{E}_D[e_{\widehat{\pi}_{\mathrm{LCP}}}(D)], \xi_{n_{\mathrm{opt}}}\right\}\\
            &\leq \frac{1}{n_{\mathrm{opt}}} \sum_{i=1}^{n_{\mathrm{opt}}} e_\pi(D_i^{\mathrm{opt}}) + \frac{1}{3}\max\left\{\mathbb{E}_D[e_{\widehat{\pi}_{\mathrm{LCP}}}(D)], \xi_{n_{\mathrm{opt}}}\right\}\\
            &\leq \mathbb{E}_D[e_\pi(D)] + \frac{1}{3}\max\left\{\mathbb{E}_D[e_\pi(D)], \xi_{n_{\mathrm{opt}}}\right\} + \frac{1}{3}\max\left\{\mathbb{E}_D[e_{\widehat{\pi}_{\mathrm{LCP}}}(D)], \xi_{n_{\mathrm{opt}}}\right\}.
        \end{aligned}
    \end{equation*}
    Taking the infimum over $\pi \in \Pi$ gives
    \begin{equation*}
        \mathbb{E}_D[e_{\widehat{\pi}_{\mathrm{LCP}}}(D)]
        \leq \mathcal{E}_{\Pi}^{\mathrm{pess}} + \frac{1}{3} \max\{\mathcal{E}_{\Pi}^{\mathrm{pess}}, \xi_{n_{\mathrm{opt}}}\} + \frac{1}{3} \max\{\mathbb{E}_D[e_{\widehat{\pi}_{\mathrm{LCP}}}(D)], \xi_{n_{\mathrm{opt}}}\}.
    \end{equation*}
    This step does not require the infimum to be attained, because $t \mapsto t + \max\{t, \xi_{n_{\mathrm{opt}}}\}/3$ is continuous and nondecreasing.
    Moreover, Lemma~\ref{lem:pointwise_benchmark_quadratic_growth} gives $\mathbb{E}_D[e_{\widehat{\pi}_{\mathrm{LCP}}}(D)] \leq H_{\mathrm{M}}$.
    If $\mathbb{E}_D[e_{\widehat{\pi}_{\mathrm{LCP}}}(D)] \leq \max\{\mathcal{E}_{\Pi}^{\mathrm{pess}}, \xi_{n_{\mathrm{opt}}}\}$, then (\ref{eq:subopt_localized_oracle_bound}) follows immediately from $\mathcal{E}_{\Pi}^{\mathrm{pess}} \geq 0$.
    Otherwise, rearranging the preceding inequality yields
    \begin{equation*}
        \mathbb{E}_D[e_{\widehat{\pi}_{\mathrm{LCP}}}(D)]
        \leq \frac{3}{2} \mathcal{E}_{\Pi}^{\mathrm{pess}} + \frac{1}{2} \max\left\{\mathcal{E}_{\Pi}^{\mathrm{pess}}, \xi_{n_{\mathrm{opt}}}\right\}
        \leq \mathcal{E}_{\Pi}^{\mathrm{pess}} + \max\left\{\mathcal{E}_{\Pi}^{\mathrm{pess}}, \xi_{n_{\mathrm{opt}}}\right\}.
    \end{equation*}
    Combining this with the bound by $H_{\mathrm{M}}$ proves (\ref{eq:subopt_localized_oracle_bound}).
    
    Finally, Lemma~\ref{lem:pointwise_benchmark_quadratic_growth} implies
    \begin{equation*}
        \mathbb{E}_D\left[\left\|x_{\widehat{\pi}_{\mathrm{LCP}}}(D)-\bar{x}(D)\right\|^2\right] \leq \frac{2 \mathbb{E}_D[e_{\widehat{\pi}_{\mathrm{LCP}}}(D)]}{\kappa_{\mathrm{M}}}.
    \end{equation*}
    Applying (\ref{eq:subopt_localized_oracle_bound}) establishes (\ref{eq:subopt_estimation_score_bound}).
\end{proof}

\begin{proof}[\bf Proof of Theorem~\ref{thm:subopt_excess_preference_risk}]
    Fix a target context $D$ for which the score bounds hold, and define
    \begin{equation*}
        \begin{aligned}
            g_D(x) 
            &= D_{\mathrm{KL}}\left(P_{\mathrm{M}, x^\star(D)}(\cdot\mid D) \middle\| P_{\mathrm{M},x}(\cdot\mid D)\right)\\
            &= \mathbb{E}_{y \sim P_{\mathrm{M}, x^\star(D)}(\cdot \mid D)}\left[\ell_x(y,D) - \ell_{x^\star(D)}(y, D)\right].
        \end{aligned}
    \end{equation*}
    We first establish global smoothness.
    Since the true preference distribution does not depend on $x$ and $\mathcal{Y}(D)$ is finite, differentiation commutes with the expectation defining $g_D$.
    For BT feedback,
    \begin{equation*}
        0 \leq g_D''(x) = \sigma(x) \sigma(-x) \leq \frac{1}{4}.
    \end{equation*}
    For PL feedback, each stage contributes a positive semidefinite categorical covariance matrix to the Hessian of $\ell_x(y,D)$.
    For any probability vector $\boldsymbol{\omega}$, symmetry and the maximum absolute row-sum bound give
    \begin{equation*}
        \left\|\operatorname{diag}(\boldsymbol{\omega}) - \boldsymbol{\omega} \boldsymbol{\omega}^{\top}\right\|
        \leq \max_j 2\omega_j(1-\omega_j)
        \leq \frac{1}{2}.
    \end{equation*}
    Embedding the stagewise matrices in the full score space, summing the $K - 1$ contributions, and taking expectation under the true preference distribution therefore yields
    \begin{equation*}
        0 \preceq \nabla^2 g_D(x)
        \preceq \frac{K - 1}{2} I
        = \Lambda_{\mathrm{PL}} I.
    \end{equation*}
    Thus, in either model, $g_D$ is globally $\Lambda_{\mathrm{M}}$-smooth.
    
    Because $g_D$ is nonnegative, applying the smoothness inequality at $z-\nabla g_D(z)/\Lambda_{\mathrm{M}}$ gives
    \begin{equation*}
        0 \leq g_D\left(z - \frac{\nabla g_D(z)}{\Lambda_{\mathrm{M}}}\right) \leq g_D(z) - \frac{\|\nabla g_D(z)\|^2}{2\Lambda_{\mathrm{M}}}.
    \end{equation*}
    The auxiliary point $z-\nabla g_D(z)/\Lambda_{\mathrm{M}}$ need not belong to $\mathcal{X}(D)$: $g_D$ is nonnegative and globally $\Lambda_{\mathrm{M}}$-smooth on the unrestricted score space, with the PL likelihood defined by the same formula for all $x \in \mathbb{R}^K$.
    Consequently,
    \begin{equation*} 
        \|\nabla g_D(z)\| \leq \sqrt{2\Lambda_{\mathrm{M}}g_D(z)}.
    \end{equation*}
    Applying smoothness once more and using Cauchy--Schwarz inequality, we obtain
    \begin{equation*}
        g_D(x) \leq g_D(z) + \langle\nabla g_D(z), x - z\rangle + \frac{\Lambda_{\mathrm{M}}}{2} \|x - z\|^2
        \leq \left(\sqrt{g_D(z)} + \sqrt{\frac{\Lambda_{\mathrm{M}}}{2}} \|x - z\|\right)^2.
    \end{equation*}

    Set $z = \bar{x}(D)$ and $x = x_{\widehat{\pi}_{\mathrm{LCP}}}(D)$.
    Applying Minkowski's inequality in $L^2$ over the target context distribution gives
    \begin{equation*}
        \begin{aligned}
            \sqrt{\mathcal{R}(\widehat{\pi}_{\mathrm{LCP}})}
            &= \sqrt{\mathbb{E}_D[g_D(x_{\widehat{\pi}_{\mathrm{LCP}}}(D))]}\\
            &\leq \sqrt{\mathbb{E}_D\left[\left(\sqrt{g_D(\bar{x}(D))} + \sqrt{\frac{\Lambda_{\mathrm{M}}}{2}} \|x_{\widehat{\pi}_{\mathrm{LCP}}}(D) - \bar{x}(D)\|\right)^2\right]}\\
            &\leq \sqrt{\mathbb{E}_D[g_D(\bar{x}(D))]} + \sqrt{\frac{\Lambda_{\mathrm{M}}}{2} \mathbb{E}_D\left[\left\|x_{\widehat{\pi}_{\mathrm{LCP}}}(D) - \bar{x}(D)\right\|^2\right]}.
        \end{aligned}
    \end{equation*}
    The first term is $\sqrt{b_{\alpha,\gamma}}$.
    Substituting (\ref{eq:subopt_estimation_score_bound}) into the second term proves (\ref{eq:subopt_true_preference_risk_bound}) on the event of Lemma~\ref{lem:relative_uniform_convergence}.

    Finally, for every $\pi \in \Pi$, nonnegativity of $\ell_{x^\star(D)}(y, D)$ and Lemma~\ref{lem:uniform_boundedness_pessimistic_loss} imply
    \begin{equation*}
        g_D(x_\pi(D)) \leq \mathbb{E}_{y \sim P_{\mathrm{M}, x^\star(D)}(\cdot \mid D)}[\ell_{x_\pi(D)}(y, D)] \leq H_{\mathrm{M}}.
    \end{equation*}
    Taking expectation over $D$ proves $\mathcal{R}(\pi) \leq H_{\mathrm{M}}$.
\end{proof}

\begin{proof} [\bf Proof of Corollary~\ref{cor:subopt_explicit_preference_rate}]
    The choice in (\ref{eq:subopt_explicit_localized_radius}) gives
    \begin{equation*}
        \frac{d_\Pi}{n_{\mathrm{opt}}} \leq \frac{\zeta_{n_{\mathrm{opt}}}}{1024C_{\mathrm{loc}}^2}.
    \end{equation*}
    Hence, using $C_{\mathrm{loc}} \geq 1$,
    \begin{equation*}
        16C_{\mathrm{loc}}\sqrt{\frac{d_\Pi\zeta_{n_{\mathrm{opt}}}}{n_{\mathrm{opt}}}} \leq \frac{\zeta_{n_{\mathrm{opt}}}}{2},
        \qquad
        \frac{16C_{\mathrm{loc}}d_\Pi}{n_{\mathrm{opt}}} \leq \frac{\zeta_{n_{\mathrm{opt}}}}{64C_{\mathrm{loc}}} \leq \frac{\zeta_{n_{\mathrm{opt}}}}{2}.
    \end{equation*}
    Combining these inequalities with (\ref{eq:subopt_localized_complexity_condition}) gives $16 \mathfrak{R}_{n_{\mathrm{opt}}}(\zeta_{n_{\mathrm{opt}}}) \leq \zeta_{n_{\mathrm{opt}}}$.
    The remaining condition in (\ref{eq:subopt_localized_radius}) holds by construction.

    We may therefore apply Theorem~\ref{thm:subopt_excess_preference_risk} with $\xi_{n_{\mathrm{opt}}} = H_{\mathrm{M}} \zeta_{n_{\mathrm{opt}}}$.
    Using the inequalities $\min\{H_{\mathrm{M}}, A\} \leq A$, $(\sqrt{u} + \sqrt{v})^2 \leq 2u + 2v$ and
    \begin{equation*}
        \mathcal{E}_{\Pi}^{\mathrm{pess}} + \max\left\{\mathcal{E}_{\Pi}^{\mathrm{pess}}, \xi_{n_{\mathrm{opt}}}\right\}
        \leq 2 \mathcal{E}_{\Pi}^{\mathrm{pess}} + \xi_{n_{\mathrm{opt}}},
    \end{equation*}
    we obtain (\ref{eq:subopt_explicit_preference_rate}).
    The additional cap follows from $\mathcal{R}(\widehat{\pi}_{\mathrm{LCP}}) \leq H_{\mathrm{M}}$.
\end{proof}

\begin{proof}[\bf Proof of Lemma~\ref{lem:subopt_score_error_to_policy_kl}]
    Fix a policy $\pi$ satisfying the stated conditions.
    Let $\mu$ denote the joint distribution of $(s, a, a')$ obtained by drawing $s \sim \rho$ and, conditionally on $s$, drawing $a \sim \pi^\star(\cdot \mid s)$ and $a' \sim \pi(\cdot \mid s)$ independently.
    Let $\widetilde{\mu}$ be the distribution obtained by swapping the two responses.
    The common-support condition implies that these measures are mutually absolutely continuous.

    Define
    \begin{equation*}
        T(s, a, a')
        = \log \frac{\mathrm{d}\mu}{\mathrm{d}\widetilde{\mu}} (s, a ,a')
        = \log \frac{\pi^\star(a \mid s) \pi(a' \mid s)}{\pi(a \mid s) \pi^\star(a' \mid s)}.
    \end{equation*}
    By (\ref{eq:kl_regularized_solution_r_star}) and the definition of $r_\pi$, the prompt-dependent normalizing term cancels, giving
    \begin{equation*}
        \beta T(s,a,a') = [r_\pi(s, a') - r^\star(s, a')] - [r_\pi(s,a) - r^\star(s,a)].
    \end{equation*}
    Consequently, the definition of $C_{\mathrm{comp}}(\pi^\star;\Pi)$ yields
    \begin{equation} \label{eq:subopt_cross_log_ratio_second_moment}
        \begin{aligned}
            \mathbb{E}_{\mu}[T^2] 
            &= \frac{1}{\beta^2} \mathbb{E}_{\substack{s \sim \rho\\ a \sim \pi^\star(\cdot \mid s), a' \sim \pi(\cdot \mid s)}}\left[\left([r_\pi(s,a) - r^\star(s,a)] - [r_\pi(s,a') - r^\star(s,a')]\right)^2\right]\\
            &\leq \frac{C_{\mathrm{comp}}(\pi^\star; \Pi)}{\beta^2} \mathbb{E}_D\left[\|x_\pi(D) - x^\star(D)\|^2\right]
            < \infty,
        \end{aligned}
    \end{equation}
    where finiteness follows from the bounded comparison scores and the assumed finite comparison coefficient.

    Swapping $a$ and $a'$ exchanges $\mu$ and $\widetilde{\mu}$ and changes $T$ to $-T$.
    Therefore,
    \begin{equation*}
        \mathbb{E}_{\widetilde{\mu}}[T^2]
        = \mathbb{E}_{\mu}[T^2],
        \qquad
        \mathbb{E}_{\widetilde{\mu}}[T]
        = -\mathbb{E}_{\mu}[T].
    \end{equation*}
    The first moments are finite by Cauchy--Schwarz.
    Using $\mathrm{d}\mu = e^T \mathrm{d}\widetilde{\mu}$ and $t \tanh(t/2) \leq t^2/2$ for every $t \in \mathbb{R}$, we obtain
    \begin{equation*}
        \begin{aligned}
            D_{\mathrm{KL}}(\mu\|\widetilde{\mu})
            = \mathbb{E}_{\mu}[T]
            = \frac{1}{2} \int T(\mathrm{d}\mu - \mathrm{d}\widetilde{\mu})
            \overset{\text{(i)}}{=} \frac{1}{2} \int T\tanh(T/2) (\mathrm{d}\mu + \mathrm{d}\widetilde{\mu}) \\
            \leq \frac{1}{4} \int T^2(\mathrm{d}\mu + \mathrm{d}\widetilde{\mu})
            = \frac{1}{2}\mathbb{E}_{\mu}[T^2],
        \end{aligned}
    \end{equation*}
    where equality (i) follows from
    \begin{equation*}
        \mathrm{d}\mu - \mathrm{d}\widetilde{\mu}
        = (e^T-1) \mathrm{d}\widetilde{\mu}
        = \frac{e^T-1}{e^T+1} (e^T+1) \mathrm{d}\widetilde{\mu}
        =\frac{e^T-1}{e^T+1} (\mathrm{d}\mu + \mathrm{d}\widetilde{\mu}).
    \end{equation*}
    Both $\mu$ and $\widetilde{\mu}$ have prompt marginal $\rho$.
    Hence, the prompt-marginal term in the KL chain rule vanishes, since $D_{\mathrm{KL}}(\rho \| \rho) = 0$.
    Conditional on $s$, their joint response distributions are $\pi^\star(\cdot \mid s) \otimes \pi(\cdot \mid s)$ and $\pi(\cdot \mid s) \otimes \pi^\star(\cdot \mid s)$, respectively.
    Applying the chain rule and expanding the conditional log-likelihood ratio therefore gives
    \begin{equation*}
        \begin{aligned}
            D_{\mathrm{KL}}(\mu\|\widetilde{\mu})
            &= \mathbb{E}_{s \sim \rho} \mathbb{E}_{\substack{a \sim \pi^\star(\cdot \mid s)\\ a' \sim \pi(\cdot \mid s)}} \left[\log \frac{\pi^\star(a \mid s) \pi(a' \mid s)}{\pi(a \mid s) \pi^\star(a' \mid s)}\right]\\
            &= \mathbb{E}_{s \sim \rho} \left[D_{\mathrm{KL}}\left(\pi^\star(\cdot \mid s) \middle\| \pi(\cdot \mid s)\right) + D_{\mathrm{KL}}\left(\pi(\cdot \mid s) \middle\| \pi^\star(\cdot \mid s)\right)\right].
        \end{aligned}
    \end{equation*}
    Dropping the nonnegative $D_{\mathrm{KL}}(\pi^\star(\cdot\mid s)\|\pi(\cdot\mid s))$ term and applying (\ref{eq:subopt_cross_log_ratio_second_moment}) proves (\ref{eq:subopt_score_error_to_policy_kl}).
\end{proof}

\begin{proof}[\bf Proof of Theorem~\ref{thm:subopt_regularized}]
    Work on the event of Lemma~\ref{lem:relative_uniform_convergence}, which has conditional probability at least $1-\delta$.
    Consider a realization for which $\widehat{\pi}_{\mathrm{LCP}}$ shares the response support of $\pi^\star$ for $\rho$-almost every prompt $s$.

    We first bound the comparison-score error relative to the true score.
    For almost every target context $D$, recall that
    \begin{equation*}
        \begin{aligned}
            g_D(x) 
            &= D_{\mathrm{KL}}\left(P_{\mathrm{M}, x^\star(D)}(\cdot\mid D) \middle\| P_{\mathrm{M},x}(\cdot\mid D)\right)\\
            &= \mathbb{E}_{y \sim P_{\mathrm{M}, x^\star(D)}(\cdot \mid D)}\left[\ell_x(y,D) - \ell_{x^\star(D)}(y, D)\right].
        \end{aligned}
    \end{equation*}
    By the proof of Lemma~\ref{lem:pointwise_benchmark_quadratic_growth}, each outcome-wise loss $\ell_x(y,D)$ is $\kappa_{\mathrm M}$-strongly convex on $\mathcal{X}(D)$.
    Taking expectation under a distribution independent of $x$ and subtracting an $x$-independent term preserve the same strong-convexity constant.
    Hence, $g_D$ is also $\kappa_{\mathrm{M}}$-strongly convex on $\mathcal{X}(D)$.
    Since $g_D(x) \geq 0$ and $g_D(x^\star(D)) = 0$, the true comparison score $x^\star(D)$ minimizes $g_D$ over $\mathcal{X}(D)$.
    Both $x^\star(D)$ and $x_\pi(D)$ belong to this domain, so quadratic growth gives
    \begin{equation*}
        g_D(x_\pi(D)) \geq \frac{\kappa_{\mathrm{M}}}{2} \|x_\pi(D) - x^\star(D)\|^2.
    \end{equation*}
    Since $\bar{x}(D) \in \mathcal{X}(D)$, applying this bound at $x = \bar{x}(D)$ and taking expectation over $D$ gives
    \begin{equation} \label{eq:subopt_upper_bound_of_x_bar_and_x_star}
        \mathbb{E}_D \left[\|\bar{x}(D) - x^\star(D)\|^2\right]
        \leq \frac{2}{\kappa_{\mathrm{M}}} \mathbb{E}_D[g_D(\bar{x}(D))]
        = \frac{2b_{\alpha, \gamma}}{\kappa_{\mathrm{M}}}.
    \end{equation}
    Combining this inequality with (\ref{eq:subopt_estimation_score_bound}) and applying the triangle inequality in $L^2$ yields
    \begin{equation*}
        \begin{aligned}
            &\sqrt{\mathbb{E}_D \left[\|x_{\widehat{\pi}_{\mathrm{LCP}}}(D) - x^\star(D)\|^2\right]}\\
            \leq& \sqrt{\mathbb{E}_D \left[\|x_{\widehat{\pi}_{\mathrm{LCP}}}(D) - \bar{x}(D)\|^2\right]} + \sqrt{\mathbb{E}_D\left[\|\bar{x}(D) - x^\star(D)\|^2\right]}\\
            \leq& \sqrt{\frac{2}{\kappa_{\mathrm{M}}}} \left(\sqrt{b_{\alpha, \gamma}} + \sqrt{\mathcal{E}_{\Pi}^{\mathrm{pess}} + \max \left\{\mathcal{E}_{\Pi}^{\mathrm{pess}}, \xi_{n_{\mathrm{opt}}}\right\}}\right).
        \end{aligned}
    \end{equation*}
    Independently, the same curvature argument and the deterministic risk bound in Theorem~\ref{thm:subopt_excess_preference_risk} give
    \begin{equation*}
        \mathbb{E}_D \left[\|x_{\widehat{\pi}_{\mathrm{LCP}}}(D) - x^\star(D)\|^2\right]
        \leq \frac{2}{\kappa_{\mathrm{M}}} \mathcal{R}(\widehat{\pi}_{\mathrm{LCP}})
        \leq \frac{2H_{\mathrm{M}}}{\kappa_{\mathrm{M}}}.
    \end{equation*}
    Consequently,
    \begin{equation} \label{eq:subopt_true_comparison_score_bound}
        \mathbb{E}_D \left[\|x_{\widehat{\pi}_{\mathrm{LCP}}}(D) - x^\star(D)\|^2\right]
        \leq \frac{2}{\kappa_{\mathrm{M}}} \min\left\{H_{\mathrm{M}}, \left(\sqrt{b_{\alpha, \gamma}} + \sqrt{\mathcal{E}_{\Pi}^{\mathrm{pess}} + \max\left\{\mathcal{E}_{\Pi}^{\mathrm{pess}}, \xi_{n_{\mathrm{opt}}}\right\}}\right)^2\right\}.
    \end{equation}

    We now establish the Gibbs identity for the regularized objective.
    By (\ref{eq:kl_regularized_solution_r_star}), for each policy $\pi$ under consideration, the following identity holds for $\rho$-almost every $s$ and $\pi(\cdot\mid s)$-almost every $a$:
    \begin{equation*}
        r^\star(s, a) - \beta\log\frac{\pi(a \mid s)}{\pi_{\mathrm{ref}}(a \mid s)} 
        = \beta \log Z_\beta^\star(s) - \beta\log\frac{\pi(a \mid s)}{\pi^\star(a \mid s)}.
    \end{equation*}
    The assumed finiteness of $J$ and $H$ justifies the following expectations.
    In particular, finite $H(\pi)$ implies integrability of the policy log-ratio, since its negative part has expectation at most $1/e$.
    Applying the preceding identity first to $\pi^\star$ therefore shows that $\log Z_\beta^\star(s)$ is integrable.
    Taking expectation under $s \sim \rho$ and $a \sim \pi(\cdot \mid s)$ yields
    \begin{equation*}
        J_\beta(\pi) = \beta\mathbb{E}_{s \sim \rho} \log Z_\beta^\star(s) - \beta\mathbb{E}_{s \sim \rho}\left[D_{\mathrm{KL}}\left(\pi(\cdot \mid s) \middle\| \pi^\star(\cdot \mid s)\right)\right].
    \end{equation*}
    Setting $\pi = \pi^\star$ gives $J_\beta(\pi^\star) = \beta\mathbb{E}_{s \sim \rho}\log Z_\beta^\star(s)$.
    Hence,
    \begin{equation*}
        J_\beta(\pi^\star) - J_\beta(\pi)
        = \beta\mathbb{E}_{s \sim \rho}\left[D_{\mathrm{KL}}\left(\pi(\cdot \mid s) \middle\| \pi^\star(\cdot \mid s)\right)\right].
    \end{equation*}
    Applying this identity to $\pi = \widehat{\pi}_{\mathrm{LCP}}$ and $\pi = \pi^\dagger$ and subtracting gives
    \begin{equation*}
        \begin{aligned}
            &\mathrm{SubOpt}_\beta(\widehat{\pi}_{\mathrm{LCP}}, \pi^\dagger)
            = J_\beta(\pi^\dagger) - J_\beta(\widehat{\pi}_{\mathrm{LCP}})
            = J_\beta(\pi^\star) - J_\beta(\widehat{\pi}_{\mathrm{LCP}}) - \left(J_\beta(\pi^\star) - J_\beta(\pi^\dagger)\right)\\
            &\quad = \beta\mathbb{E}_{s \sim \rho}\left[D_{\mathrm{KL}}\left(\widehat{\pi}_{\mathrm{LCP}}(\cdot \mid s) \middle\| \pi^\star(\cdot \mid s)\right)\right] - \beta\mathbb{E}_{s \sim \rho}\left[D_{\mathrm{KL}}\left(\pi^\dagger(\cdot \mid s) \middle\| \pi^\star(\cdot \mid s)\right)\right]
        \end{aligned}
    \end{equation*}
    Applying Lemma~\ref{lem:subopt_score_error_to_policy_kl} to the learned policy yields
    \begin{equation} \label{eq:subopt_beta_bound_proof_in_theorem}
        \begin{aligned}
            \mathrm{SubOpt}_{\beta}(\widehat{\pi}_{\mathrm{LCP}}, \pi^\dagger)
            \leq& \frac{C_{\mathrm{comp}}(\pi^\star; \Pi)}{2\beta} \mathbb{E}_D\left[\|x_{\widehat{\pi}_{\mathrm{LCP}}}(D) - x^\star(D)\|^2\right]\\
            &- \beta\mathbb{E}_{s \sim \rho} \left[D_{\mathrm{KL}}\left(\pi^\dagger(\cdot \mid s) \middle\| \pi^\star(\cdot \mid s)\right)\right].
        \end{aligned}
    \end{equation}
    Substituting (\ref{eq:subopt_true_comparison_score_bound}) proves (\ref{eq:subopt_regularized_bound}).
\end{proof}

\begin{proof}[\bf Proof of Lemma~\ref{lem:subopt_conditional_calibration_miscoverage}]
    Throughout the proof, we condition on the pilot sample.
    Let $F_S$ denote the cumulative distribution function of the target nonconformity score, with $F_S(+\infty) = 1$.
    For each realized calibration sample, model correctness and the definition of the conformal prediction set imply
    \begin{equation*}
        \bar{\gamma}_{\mathrm{cal}}
        = 1 - \mathbb{P}_{(D,y)}\left\{\widehat{s}(D,y) \leq \widehat{q}_{1-\gamma}^{\mathrm{cal}}\right\}
        = 1 - F_S\bigl(\widehat{q}_{1-\gamma}^{\mathrm{cal}}\bigr).
    \end{equation*}
    Recall $k_\gamma = \lceil (n_{\mathrm{cal}} + 1)(1 - \gamma)\rceil$.
    If $k_\gamma = n_{\mathrm{cal}} + 1$, then $\widehat{q}_{1-\gamma}^{\mathrm{cal}} = +\infty$ and $\bar{\gamma}_{\mathrm{cal}} = 0$.
    It therefore suffices to consider $k_\gamma \leq n_{\mathrm{cal}}$.
    Set
    \begin{equation*}
        \tau = 2 \gamma + \frac{4\log(1/\delta_{\mathrm{cal}})}{3n_{\mathrm{cal}}}.
    \end{equation*}
    The claim is immediate if $\tau \geq 1$.
    Otherwise, define the population quantile
    \begin{equation*}
        q_\tau
        = \inf\left\{t \in [0, +\infty]: F_S(t) \geq 1 - \tau\right\},
    \end{equation*}
    allowing $q_\tau = +\infty$.
    Its defining properties give $F_S(q_\tau-) \leq 1 - \tau \leq F_S(q_\tau)$, and hence
    \begin{equation*}
        p_\tau = \mathbb{P}\{S_i^{\mathrm{cal}} \geq q_\tau\} = 1 - F_S(q_\tau-) \geq \tau.
    \end{equation*}
    Here, $F_S(q_\tau-)$ denotes the left limit.

    On the event $\{\bar{\gamma}_{\mathrm{cal}} > \tau\}$, we have $F_S(\widehat{q}_{1-\gamma}^{\mathrm{cal}}) < 1 - \tau$, and therefore $\widehat{q}_{1-\gamma}^{\mathrm{cal}} < q_\tau$.
    Since the calibrated threshold is the $k_\gamma$-th order statistic, at least $k_\gamma$ calibration scores are strictly below $q_\tau$.
    Consequently, on this event,
    \begin{equation*}
        N_\tau
        = \sum_{i \in \mathcal{I}_{\mathrm{cal}}} \mathbb{I}\{S_i^{\mathrm{cal}} \geq q_\tau\}
        \leq n_{\mathrm{cal}} - k_\gamma
        < n_{\mathrm{cal}}\gamma.
    \end{equation*}

    By the conditional i.i.d. assumption, $N_\tau \sim \operatorname{Binomial}(n_{\mathrm{cal}}, p_\tau)$.
    Applying Markov's inequality to $4^{-N_\tau}$ and using $1 - u \leq e^{-u}$ yield
    \begin{equation*}
        \begin{aligned}
            \mathbb{P}\{\bar{\gamma}_{\mathrm{cal}} > \tau\}
            \leq \mathbb{P}\left\{N_\tau \leq n_{\mathrm{cal}}\gamma\right\}
            \leq 4^{n_{\mathrm{cal}}\gamma}\mathbb{E}[4^{-N_\tau}]
            = 4^{n_{\mathrm{cal}}\gamma} \left(1 - \frac{3p_\tau}{4}\right)^{n_{\mathrm{cal}}}\\
            \leq \exp\left(n_{\mathrm{cal}}\gamma\log4 - \frac{3n_{\mathrm{cal}}p_\tau}{4}\right)
            \leq \exp\left(n_{\mathrm{cal}}\gamma\log4 - \frac{3n_{\mathrm{cal}}\tau}{4}\right)
            \leq \delta_{\mathrm{cal}}.
        \end{aligned}
    \end{equation*}
    Thus, conditional on the pilot sample, with probability at least $1 - \delta_{\mathrm{cal}}$,
    \begin{equation*}
        \bar{\gamma}_{\mathrm{cal}}
        \leq \min\left\{1, 2\gamma + \frac{4\log(1/\delta_{\mathrm{cal}})}{3n_{\mathrm{cal}}}\right\}.
    \end{equation*}

    To prove the sharper statement, note that $k_\gamma = n_{\mathrm{cal}}+1-j_\gamma$.
    The case $j_\gamma=0$ was already treated. For $j_\gamma \geq 1$, introduce i.i.d. $V_i \sim \operatorname{Unif}(0,1)$, independent of the calibration scores, solely for the proof, and set
    \begin{equation*}
        W_i=F_S(S_i^{\mathrm{cal}}-)
        +V_i\bigl[F_S(S_i^{\mathrm{cal}})-F_S(S_i^{\mathrm{cal}}-)\bigr].
    \end{equation*}
    The randomized probability integral transform gives i.i.d. uniform $W_i$, including when the scores have atoms at $+\infty$.
    Since $W_i \leq F_S(S_i^{\mathrm{cal}})$, their order statistics satisfy $W_{(k_\gamma)}\leq F_S(S_{(k_\gamma)}^{\mathrm{cal}})$.
    Consequently,
    \begin{equation*}
        \bar\gamma_{\mathrm{cal}}
        \leq 1 - W_{(k_\gamma)},
        \qquad
        1 - W_{(k_\gamma)}\sim\operatorname{Beta}(j_\gamma, n_{\mathrm{cal}}+1-j_\gamma).
    \end{equation*}
    By the definition of $u_{n_{\mathrm{cal}}, \gamma}(\delta_{\mathrm{cal}})$,
    \begin{equation*}
        \mathbb{P}\left\{\bar\gamma_{\mathrm{cal}} \leq u_{n_{\mathrm{cal}},\gamma}(\delta_{\mathrm{cal}})\right\}
        \geq \mathbb{P}\left\{1-W_{(k_\gamma)} \leq u_{n_{\mathrm{cal}},\gamma}(\delta_{\mathrm{cal}})\right\}
        = 1 - \delta_{\mathrm{cal}}.
    \end{equation*}
    This proves the first inequality in (\ref{eq:subopt_exact_calibration_envelope}).
    If the conditional score distribution is continuous, then $\bar\gamma_{\mathrm{cal}} = 1-W_{(k_\gamma)}$ almost surely, so the beta distribution is exact.
    
    If $\tau < 1$, define
    \begin{equation*}
        \widetilde{N}_\tau
        = \sum_{i\in\mathcal{I}_{\mathrm{cal}}} \mathbb{I}\{W_i\geq1-\tau\}
        \sim \operatorname{Binomial}(n_{\mathrm{cal}}, \tau).
    \end{equation*}
    On the event $\{1 - W_{(k_\gamma)} > \tau\}$, $\widetilde{N}_\tau \leq n_{\mathrm{cal}}-k_\gamma < n_{\mathrm{cal}}\gamma$.
    The same exponential-moment bound as above therefore gives
    \begin{equation*}
        \mathbb{P}\{1-W_{(k_\gamma)}>\tau\}
        \leq 4^{n_{\mathrm{cal}}\gamma}(1-3\tau/4)^{n_{\mathrm{cal}}}
        \leq \delta_{\mathrm{cal}}.
    \end{equation*}
    Hence $u_{n_{\mathrm{cal}},\gamma}(\delta_{\mathrm{cal}}) \leq \tau$.
    If $\tau \geq 1$, the bound $u_{n_{\mathrm{cal}},\gamma}(\delta_{\mathrm{cal}}) \leq 1$ is immediate.
    Together, these cases prove the second inequality in (\ref{eq:subopt_exact_calibration_envelope}).
\end{proof}

\begin{proof}[\bf Proof of Lemma~\ref{lem:subopt_conformal_benchmark_bias}]
    Define
    \begin{equation*}
        v_D = \omega_D+(1-\omega_D)\max\{\varepsilon_{\alpha,\gamma} ,t_D\}.
    \end{equation*}
    Then $0 \leq v_D \leq1$ and $\Xi_{\alpha,\gamma} = \mathbb E_D[v_D^2]$.
    For $w,t \in [0,1]$, convexity of $t \mapsto [w+(1-w)t]^2$ on $[\varepsilon_{\alpha,\gamma}, 1]$ gives
    \begin{equation*}
        [w+(1-w)\max\{\varepsilon_{\alpha,\gamma},t\}]^2
        \leq a_\varepsilon(w)+(1-a_\varepsilon(w))t,
    \end{equation*}
    where $a_\varepsilon(w) = [w+(1-w)\varepsilon_{\alpha,\gamma}]^2 = \varepsilon_{\alpha,\gamma}^2 + (1-\varepsilon_{\alpha,\gamma}^2) r_\varepsilon(w)$.
    Indeed, for $t > \varepsilon_{\alpha,\gamma}$, the chord coefficient $(t - \varepsilon_{\alpha,\gamma}) / (1 -\varepsilon_{\alpha,\gamma})$ is at most $t$; for $t \leq \varepsilon_{\alpha,\gamma}$, the left-hand side equals $a_\varepsilon(w)$.
    Taking expectations therefore yields
    \begin{equation*}
        \begin{aligned}
            \Xi_{\alpha,\gamma}
            &\leq \varepsilon_{\alpha,\gamma}^2 + (1-\varepsilon_{\alpha,\gamma}^2) \mathbb{E}_D\left[r_\varepsilon(\omega_D) + (1-r_\varepsilon(\omega_D)) t_D\right]\\
            &\leq \varepsilon_{\alpha,\gamma}^2 + (1-\varepsilon_{\alpha,\gamma}^2)(\widetilde{\eta}_{\alpha,\gamma} + \bar{\gamma}_{\mathrm{cal}}).
        \end{aligned}
    \end{equation*}
    Together with $v_D\leq1$, this proves (\ref{eq:subopt_conformal_discrepancy_bound}).

    Fix a context $D$ for which the score bounds hold.
    We first show that $G_{\mathrm{M}}$ bounds the diameter of the outcome-wise gradients, i.e.,
    \begin{equation*}
        \sup_{y, y' \in \mathcal{Y}(D)} \left\|\nabla_x \ell_x(y,D) - \nabla_x \ell_x(y',D)\right\| \leq G_{\mathrm{M}}.
    \end{equation*}
    For BT feedback, this follows directly from $\partial_x \ell_x(y,D) = \sigma(x) - y$.
    For PL feedback, let $\boldsymbol{\omega}_k(x,y)$ denote the stage-$k$ choice-probability vector, with zero coordinates for previously selected candidates.
    Including the deterministic final stage gives
    \begin{equation*}
        \nabla_x \ell_x(y,D)
        = \sum_{k=1}^{K} \boldsymbol{\omega}_k(x,y) - \boldsymbol{1}_K.
    \end{equation*}
    Since $\boldsymbol{\omega}_1(x,y) = \operatorname{softmax}(x)$ is independent of the ranking, the first-stage vectors cancel when subtracting the gradients.
    Moreover, any two probability vectors have Euclidean distance at most $\sqrt{2}$. 
    The triangle inequality therefore yields
    \begin{equation*}
        \begin{aligned}
            \left\|\nabla_x \ell_x(y,D) - \nabla_x \ell_x(y',D)\right\|
            = \left\|\sum_{k=2}^{K} \left(\boldsymbol{\omega}_k(x,y) - \boldsymbol{\omega}_k(x,y')\right)\right\|\\ 
            \leq \sum_{k=2}^{K} \left\|\boldsymbol{\omega}_k(x,y) - \boldsymbol{\omega}_k(x,y')\right\|
            \leq \sqrt{2}(K-1)
            \leq G_{\mathrm{PL}}.
        \end{aligned}
    \end{equation*}

    If $R_k(y) = R_k(y')$, the corresponding PL choice-probability vectors also coincide.
    Applying the same argument only at stages where these sets differ, and using the BT gradient formula for binary feedback, gives, whenever $\widehat{\mathcal C}_\gamma(D) \neq \varnothing$,
    \begin{equation} \label{eq:subopt_restricted_gradient_diameter}
        \sup_{x \in \mathcal{X}(D)}\max_{y,y' \in \widehat{\mathcal C}_\gamma(D)}
        \|\nabla_x \ell_x(y,D) - \nabla_x \ell_x(y',D)\|
        \leq G_{\mathrm M} \omega_D.
    \end{equation}
    Fix any $\boldsymbol{u} \in \widehat{\mathcal{U}}_\alpha(D;\gamma)$ and write $Q = P_{\mathrm{M}, \boldsymbol{u}}(\cdot\mid D)$ and $P_D^\star = P_{\mathrm{M}, x^\star(D)}(\cdot \mid D)$.
    Recall that $\nabla g_D(x)=\mathbb{E}_{P_D^\star}[\nabla_x \ell_x(y,D)]$.
    In both models, any single outcome can receive probability arbitrarily close to one under latent scores in $\Theta(D)$.
    Thus, for nonempty $\widehat{\mathcal{C}}_\gamma(D)$, the compatibility constraint is feasible and gives $Q(\widehat{\mathcal{C}}_\gamma(D)) \geq 1 -\varepsilon_{\alpha,\gamma}$; also $P_D^\star(\widehat{\mathcal{C}}_\gamma(D)) = 1 - t_D$.
    Couple $Y \sim Q$ and $Y' \sim P_D^\star$ so that
    \begin{equation*}
        \mathbb{P}\{Y \in \widehat{\mathcal{C}}_\gamma(D),\ Y'\in \widehat{\mathcal{C}}_\gamma(D)\}
        = \min\{Q(\widehat{\mathcal{C}}_\gamma(D)), P_D^\star(\widehat{\mathcal{C}}_\gamma(D))\}
        \geq 1 - \max\{\varepsilon_{\alpha, \gamma}, t_D\}.
    \end{equation*}
    Such a coupling is obtained by matching the two membership indicators with maximal overlap and then sampling from the corresponding conditional distributions.
    On this event use (\ref{eq:subopt_restricted_gradient_diameter}), and on its complement use the global gradient-diameter bound. Since $0\leq\omega_D\leq1$, the triangle inequality gives
    \begin{equation*}
        \begin{aligned}
            \|\mathbb{E}_Q[\nabla_x\ell_x(y,D)]-\nabla g_D(x)\|
            &\leq\mathbb{E}\|\nabla_x\ell_x(Y,D)-\nabla_x\ell_x(Y',D)\|\\
            &\leq G_{\mathrm{M}}\left[\omega_D+(1-\omega_D)\max\{\varepsilon_{\alpha,\gamma},t_D\}\right]
            = G_{\mathrm{M}}v_D.
        \end{aligned}
    \end{equation*}
    For $\widehat{\mathcal{C}}_\gamma(D) = \varnothing$, the same conclusion follows from the global bound and $v_D = 1$.
    Thus, $x \mapsto \mathbb{E}_{y \sim P_{\mathrm{M},\boldsymbol{u}}(\cdot \mid D)}[\ell_x(y, D)] - g_D(x)$ is $G_{\mathrm{M}}v_D$-Lipschitz on $\mathcal{X}(D)$.
    Taking the supremum over $\boldsymbol{u} \in \widehat{\mathcal{U}}_\alpha(D; \gamma)$ preserves this bound, giving
    \begin{equation} \label{eq:subopt_conformal_loss_perturbation}
        \left|(f_D - g_D)(x) - (f_D-g_D)(z)\right| \leq G_{\mathrm{M}}v_D \|x - z\|,
        \qquad 
        x, z \in \mathcal{X}(D).
    \end{equation}
    
    By Lemma~\ref{lem:pointwise_benchmark_quadratic_growth} and the curvature argument in the proof of Theorem~\ref{thm:subopt_regularized}, both $f_D$ and $g_D$ are $\kappa_{\mathrm{M}}$-strongly convex on $\mathcal{X}(D)$.
    Their respective minimizers are $\bar{x}(D)$ and $x^\star(D)$, with $g_D(x^\star(D))=0$.
    Writing $d_D = \|\bar{x}(D) - x^\star(D)\|$, quadratic growth gives
    \begin{equation*}
        g_D(\bar{x}(D)) \geq \frac{\kappa_{\mathrm{M}}}{2}d_D^2,
        \qquad
        f_D(x^\star(D)) - f_D(\bar{x}(D)) \geq \frac{\kappa_{\mathrm{M}}}{2}d_D^2.
    \end{equation*}
    On the other hand, (\ref{eq:subopt_conformal_loss_perturbation}) implies
    \begin{equation*}
        g_D(\bar{x}(D)) + f_D(x^\star(D)) - f_D(\bar{x}(D)) \leq G_{\mathrm{M}} v_D d_D.
    \end{equation*}
    Subtracting either quadratic lower bound from this inequality and completing the square yields
    \begin{equation} \label{eq:subopt_conformal_pointwise_bias_bound}
        \max\left\{g_D(\bar{x}(D)), f_D(x^\star(D)) - f_D(\bar{x}(D))\right\}
        \leq G_{\mathrm{M}} v_D d_D - \frac{\kappa_{\mathrm{M}}}{2} d_D^2
        \leq \frac{G_{\mathrm{M}}^2}{2\kappa_{\mathrm{M}}} v_D^2.
    \end{equation}
    Taking expectations and using the definitions of $b_{\alpha, \gamma}$ and $\Xi_{\alpha, \gamma}$ yields
    \begin{equation*}
        b_{\alpha, \gamma} 
        = \mathbb{E}_D[g_D(\bar{x}(D))]
        \leq \frac{G_{\mathrm{M}}^2}{2\kappa_{\mathrm{M}}} \mathbb{E}_D[v_D^2]
        = \frac{G_{\mathrm{M}}^2}{2\kappa_{\mathrm{M}}} \Xi_{\alpha, \gamma}.
    \end{equation*}
    This proves the second bound in (\ref{eq:subopt_conformal_benchmark_kl_bias}).
    Moreover, the outcome-wise loss bound on $\mathcal{X}(D)$ established in the proof of Lemma~\ref{lem:uniform_boundedness_pessimistic_loss} implies $g_D(\bar{x}(D))\leq H_{\mathrm{M}}$. 
    Combining these two bounds proves (\ref{eq:subopt_conformal_benchmark_kl_bias}).

    Finally, the proof of Theorem~\ref{thm:subopt_excess_preference_risk} establishes global $\Lambda_{\mathrm{M}}$-smoothness of $g_D$.
    Since $g_D$ is nonnegative and vanishes at $x^\star(D)$, $\nabla g_D(x^\star(D)) = 0$, and hence
    \begin{equation*}
        g_D(x) 
        \leq \frac{\Lambda_{\mathrm{M}}}{2} \|x-x^\star(D)\|^2.
    \end{equation*}
    For every $\pi \in \Pi$, decompose the excess pessimistic loss as
    \begin{equation*}
        e_\pi(D)
        = g_D(x_\pi(D)) + (f_D - g_D)(x_\pi(D)) - (f_D - g_D)(x^\star(D)) + f_D(x^\star(D))-f_D(\bar{x}(D)).
    \end{equation*}
    Combining the preceding smoothness bound with (\ref{eq:subopt_conformal_loss_perturbation}) and (\ref{eq:subopt_conformal_pointwise_bias_bound}) yields
    \begin{equation*}
        e_\pi(D)
        \leq \frac{\Lambda_{\mathrm{M}}}{2} \|x_\pi(D)-x^\star(D)\|^2 + G_{\mathrm{M}} v_D \|x_\pi(D)-x^\star(D)\| + \frac{G_{\mathrm{M}}^2}{2\kappa_{\mathrm{M}}}v_D^2.
    \end{equation*}
    Taking expectations and applying Cauchy--Schwarz gives
    \begin{equation*}
        \begin{aligned}
            \mathbb{E}_D[e_\pi(D)]
            &\leq \frac{\Lambda_{\mathrm{M}}}{2} \mathbb{E}_D\left[\|x_\pi(D)-x^\star(D)\|^2\right]\\
            &\quad + G_{\mathrm{M}}\sqrt{\Xi_{\alpha,\gamma} \mathbb{E}_D\left[\|x_\pi(D)-x^\star(D)\|^2\right]}
            +\frac{G_{\mathrm{M}}^2}{2\kappa_{\mathrm{M}}} \Xi_{\alpha,\gamma}.
        \end{aligned}
    \end{equation*}
    Since $\mathcal{E}_\Pi^{\mathrm{pess}} \leq \mathbb{E}_D[e_\pi(D)]$ for every $\pi\in\Pi$, applying this bound along a sequence of policies whose mean squared score errors converge to $\mathcal{E}_\Pi$ and passing to the limit proves (\ref{eq:subopt_conformal_policy_approximation_bias}).
\end{proof}

\section{Worst-Case Lower Bounds} \label{app:suboptimality_lower_bound}

\subsection{Joint Lower Bound for Estimation Error and Conformal Bias} \label{app:suboptimality_lower_bound_joint}
We give the full statement and proof of Theorem~\ref{thm:subopt_joint_worst_case_lower}.

\medskip
\noindent
\textbf{Theorem~\ref{thm:subopt_joint_worst_case_lower}} (Worst-case Lower Bound for LatentCP-DPO, full statement)\textbf{.}
\emph{
    Fix $\alpha \in (0, 1)$, $B, \beta > 0$, and $\gamma \in (0, \alpha/2)$ independently of the calibration sample.
    For every integer $d_\Pi \geq 1$, and all sample sizes satisfying
    \begin{equation*}
        n_{\mathrm{pil}} \geq 2d_\Pi + 1,
        \qquad
        n_{\mathrm{cal}} \geq \frac{1}{\gamma} - 1,
        \qquad
        n_{\mathrm{opt}} \geq 8 \max\left\{d_\Pi, \frac{1}{\gamma}\right\},
    \end{equation*}
    one can construct a BT instance and a log-linear policy class $\Pi$ with a fixed offset and $d_\Pi+1$ free parameters such that Assumptions~\ref{ass:subopt_sampling} and~\ref{ass:bounded_comparison_scores} hold, and
    \begin{equation*}
        \pi^\star \in \Pi,
        \qquad
        \mathcal{E}_\Pi = 0,
        \qquad
        C_{\mathrm{comp}}(\pi^\star; \Pi) \leq 1.
    \end{equation*}
    Moreover, the localized-complexity condition (\ref{eq:subopt_localized_complexity_condition}) holds with complexity parameter $d_\Pi$ and $C_{\mathrm{loc}} = \sqrt{3}$.
    For this instance, there exists a pilot-sample event $\mathcal{H}_{\mathrm{pil}}$ of positive probability such that, conditional on $\mathcal{H}_{\mathrm{pil}}$, with probability at least $1/16$ over the calibration and optimization samples, every exact LatentCP-DPO empirical optimizer satisfies
    \begin{equation*}
        \mathrm{SubOpt}_{\beta}(\widehat{\pi}_{\mathrm{LCP}}, \pi^\star)
        \geq c_{B,\beta}\left[\frac{d_\Pi}{n_{\mathrm{opt}}} + \left(1 - \frac{2\gamma}{\alpha}\right)^2 \left(\gamma + \frac{1}{n_{\mathrm{cal}}}\right)\right],
        \tag{\ref{eq:subopt_joint_worst_case_lower}}
    \end{equation*}
    where $c_{B, \beta} > 0$ depends only on $B$ and $\beta$.
    One valid choice is
    \begin{equation} \label{eq:subopt_joint_lower_constants}
        c_{B,\beta}
        = \min\left\{\frac{v_{B,\beta}a_B^2}{256\beta}, \frac{v_{B,\beta}\min\{B^2,4\}}{16\beta}\right\},
    \end{equation}
    with
    \begin{equation*}
        v_{B,\beta}
        = \sigma(B/\beta)\sigma(-B/\beta),
        \qquad
        a_B
        =\frac{\sigma(-B/4)-\sigma(-B/2)}{2}.
    \end{equation*}
}
\begin{proof}[\bf Proof of Theorem~\ref{thm:subopt_joint_worst_case_lower}]
    We organize the proof into six steps, constructing a single instance in which calibration-induced bias and optimization-sample fluctuations contribute additively to policy suboptimality.

    \noindent
    \textbf{Step 1: Construction of the instance.}
    Set
    \begin{equation*}
        j = \lfloor(n_{\mathrm{cal}} + 1) \gamma\rfloor,
        \qquad
        w = \min\left\{\frac{1}{4}, \frac{j}{2n_{\mathrm{cal}}}\right\},
        \qquad
        h = \frac{B}{16}.
    \end{equation*}
    The sample-size assumption gives $j \geq 1$.
    Choose $\ell \in \{B/4, B/2\}$ such that
    \begin{equation*}
        |\varepsilon_{\alpha,\gamma} - \sigma(-\ell)| \geq a_B.
    \end{equation*}
    Such a choice exists because
    \begin{equation*}
        2a_B = \sigma(-B/4) - \sigma(-B/2) \leq |\varepsilon_{\alpha,\gamma} - \sigma(-B/4)| + |\varepsilon_{\alpha,\gamma} - \sigma(-B/2)|.
    \end{equation*}
    Since $\sup_{z \geq 0} \sigma'(z) \leq 1/4$, we have 
    \begin{equation*}
        a_B \leq \frac{1}{2} \sup_{z \geq 0} \sigma'(z) \cdot \left(-B/4 + B/2\right) = \frac{B}{32} = \frac{h}{2}.
    \end{equation*}

    Let the prompt space be
    \begin{equation*}
        \mathcal{S} = \{s_c, s_0\} \cup \{s_{i,+}, s_{i,-}: 1 \leq i \leq d_\Pi\}.
    \end{equation*}
    Every prompt has two responses, $0$ and $1$, and the comparison context is the fixed pair $D=(s, 0, 1)$.
    Take $\pi_{\mathrm{ref}}(0 \mid s) = \pi_{\mathrm{ref}}(1 \mid s) = 1/2$.
    For
    \begin{equation*}
        \vartheta = (t, \theta_1, \ldots, \theta_{d_\Pi})
        \in[0,B] \times[-h, h]^{d_\Pi},
    \end{equation*}
    define the prompt probabilities and comparison scores by
    \begin{equation*}
        \begin{array}{l|c|c|c}
            \hline
            s & \rho(s) & x_\vartheta(s) & x^\star(s)\\ 
            \hline
            s_c & w & t & 0\\
            s_{i,+} & 1/(8d_\Pi) & \ell+\theta_i & \ell\\
            s_{i,-} & 1/(8d_\Pi) & -\ell+\theta_i & -\ell\\
            s_0 & 3/4-w & 0 & 0\\
            \hline
        \end{array}
    \end{equation*}
    where $w = \rho(s_c)$ is defined above and $i = 1, \ldots, d_\Pi$.
    Under the uniform reference policy, define
    \begin{equation*}
        \pi_\vartheta(1 \mid s)
        = \sigma\left(\frac{x_\vartheta(s)}{\beta}\right),
        \qquad
        r^\star(s,a) = \frac{2a-1}{2}x^\star(s).
    \end{equation*}
    Then the policy-induced comparison score satisfies
    \begin{equation*}
        \beta \log\frac{\pi_\vartheta(1 \mid s) / \pi_{\mathrm{ref}}(1 \mid s)}{\pi_\vartheta(0 \mid s)/\pi_{\mathrm{ref}}(0 \mid s)}
        = x_\vartheta(s).
    \end{equation*}
    The comparison score is affine in $\vartheta = (t, \theta_1, \ldots, \theta_{d_\Pi})$, with fixed offsets $\pm \ell$ on the paired prompts.
    Representing these offsets by an additional feature whose coefficient is fixed at one yields a log-linear policy class with $d_\Pi + 1$ free parameters.

    The true BT labels satisfy
    \begin{equation*}
        \mathbb{P} \{y=1 \mid s\} 
        = \frac{e^{r^\star(s,1)}}{e^{r^\star(s,0)} + e^{r^\star(s,1)}}
        = \sigma\left(r^\star(s,1) - r^\star(s,0)\right)
        = \sigma(x^\star(s)).
    \end{equation*}
    Setting all free parameters in $\vartheta$ to zero gives $x_{\boldsymbol{0}}(s) = x^\star(s)$ for every prompt.
    Under the uniform reference policy, (\ref{eq:kl_regularized_solution_r_star}) therefore yields $\pi^\star = \pi_{\boldsymbol{0}}$.
    Draw all pilot and calibration preference units independently from this distribution, and draw the optimization contexts independently from its context marginal.

    Enumerate the pilot observations as $(s_i^{\mathrm{pil}}, y_i^{\mathrm{pil}})$, $i = 1, \ldots, n_{\mathrm{pil}}$.
    Define the event
    \begin{equation*}
        \mathcal{H}_{\mathrm{pil}}
        = \bigcap_{i=1}^{d_\Pi} \left\{(s_{2i-1}^{\mathrm{pil}},y_{2i-1}^{\mathrm{pil}}) = (s_{i,+},1), (s_{2i}^{\mathrm{pil}},y_{2i}^{\mathrm{pil}}) = (s_{i,-},0)\right\}
        \cap
        \bigcap_{r=2d_\Pi+1}^{n_{\mathrm{pil}}}\left\{(s_r^{\mathrm{pil}},y_r^{\mathrm{pil}}) = (s_c,1)\right\}.
    \end{equation*}
    By independence and the specified BT label probabilities,
    \begin{equation*}
        \mathbb{P}(\mathcal H_{\mathrm{pil}})
        = \left(\frac{\sigma(\ell)}{8d_\Pi}\right)^{2d_\Pi} \left(\frac{w}{2}\right)^{n_{\mathrm{pil}}-2d_\Pi} > 0.
    \end{equation*}
    On this event, the empirical pilot loss is
    \begin{equation*}
        \widehat{L}_{\mathrm{pil}}(\vartheta)
        = \frac{n_{\mathrm{pil}}-2d_\Pi}{n_{\mathrm{pil}}}\log(1+e^{-t})
        + \frac{1}{n_{\mathrm{pil}}} \sum_{i=1}^{d_\Pi}\left[\log(1+e^{-\ell-\theta_i}) + \log(1+e^{-\ell+\theta_i})\right].
    \end{equation*}
    Since $n_{\mathrm{pil}} \geq 2d_\Pi + 1$, the coefficient of $\log(1+e^{-t})$ is positive, and this term is strictly decreasing in $t\in[0,B]$.
    Its unique minimizer is therefore $t = B$.
    For each $i$, the corresponding two terms form a strictly convex even function of $\theta_i$, whose unique minimizer on $[-h, h]$ is zero.
    Consequently, $\{(B, 0, \ldots, 0)\} = \operatorname*{argmin}_{\vartheta} \widehat{L}_{\mathrm{pil}}(\vartheta)$.

    \noindent
    \textbf{Step 2: Verification of the assumptions and complexity bound.}
    The sampling assumption holds by construction.
    Moreover, $|x_\vartheta(s)| \leq B$ and $|x^\star(s)| \leq B$ for every prompt and parameter, because $\ell + h \leq 9B/16$.
    The parameter domain is compact, and the empirical pilot and pessimistic losses are continuous, so their minima are attained.
    All policies have full response support, and $\pi^\star \in \Pi$ gives $\mathcal{E}_\Pi = 0$.

    For any two responses $a, a' \in \{0, 1\}$, the difference of implicit-reward errors is either zero or $\pm (x_\vartheta(s) - x^\star(s))$.
    Taking expectations over $s \sim \rho$ and conditionally independent responses $a \sim \pi^\star(\cdot \mid s)$ and $a' \sim \pi_\vartheta(\cdot \mid s)$ gives
    \begin{equation*}
        \begin{aligned}
            &\mathbb{E}_{\substack{s \sim \rho\\ a \sim \pi^\star(\cdot \mid s), a' \sim \pi_\vartheta(\cdot \mid s)}}\left[\left([r_{\pi_\vartheta}(s,a)-r^\star(s,a)] - [r_{\pi_\vartheta}(s,a')-r^\star(s,a')]\right)^2\right]\\
            &\qquad \leq \mathbb{E}_{s\sim\rho}\left[|x_\vartheta(s) - x^\star(s)|^2\right] 
            = \mathbb{E}_D\left[|x_{\pi_\vartheta}(D) - x^\star(D)|^2\right],
        \end{aligned}
    \end{equation*}
    where the equality uses the deterministic comparison pair $D = (s, 0, 1)$.
    Thus, every ratio in the supremum defining $C_{\mathrm{comp}}(\pi^\star; \Pi)$ is at most one.
    Taking the supremum over $\pi_\vartheta \in \Pi$ therefore yields $C_{\mathrm{comp}}(\pi^\star; \Pi) \leq 1$.

    To verify localized complexity, condition on any pilot and calibration realization.
    At $s_0$, every policy has score zero, so the prediction set is either empty or full, its pointwise pessimistic benchmark is zero, and 
    \begin{equation*}
        e_\pi(s_0) 
        = f_{s_0}(x_\pi(s_0)) - f_{s_0}(\bar x(s_0))
        = f_{s_0}(0) - f_{s_0}(0)
        = 0.
    \end{equation*}
    Consequently, every $g \in \mathcal{H}_\zeta$ vanishes at $s_0$ and can be nonzero only at the other $2d_\Pi + 1$ prompts.
    By the definition of $\mathcal{H}_\zeta$, it also satisfies $0 \leq g \leq 1$ and $\mathbb{E}_\rho[g] \leq \zeta$.

    Let
    \begin{equation*}
        Z_s = \frac{1}{n_{\mathrm{opt}}}\sum_{r=1}^{n_{\mathrm{opt}}} \epsilon_r \mathbb{I}\{s_r^{\mathrm{opt}}=s\},
        \qquad
        \mathcal{S}_+=\mathcal{S} \setminus \{s_0\},
    \end{equation*}
    where the $\epsilon_r$ are independent Rademacher signs.
    Weighted Cauchy--Schwarz gives
    \begin{equation*}
        \sup_{g \in \mathcal{H}_\zeta}\left|\sum_{s \in \mathcal{S}_+} g(s)Z_s\right|
        \leq \sup_{g \in \mathcal{H}_\zeta}\left|\sum_{s \in \mathcal{S}_+} \bigl[\sqrt{\rho(s)} g(s)\bigr]\left[\frac{Z_s}{\sqrt{\rho(s)}}\right]\right|
        \leq \sqrt{\zeta}\left(\sum_{s \in \mathcal{S}_+}\frac{Z_s^2}{\rho(s)}\right)^{1/2},
    \end{equation*}
    using $\sum_s\rho(s)g(s)^2 \leq \mathbb{E}_\rho[g] \leq \zeta$.
    Since $\mathbb{E}[Z_s^2] = \rho(s)/n_{\mathrm{opt}}$, Jensen's inequality yields
    \begin{equation*}
        \mathfrak{R}_{n_{\mathrm{opt}}}(\zeta)
        \leq \sqrt{\frac{(2d_\Pi+1)\zeta}{n_{\mathrm{opt}}}}
        \leq \sqrt{3} \sqrt{\frac{d_\Pi\zeta}{n_{\mathrm{opt}}}}.
    \end{equation*}
    This proves (\ref{eq:subopt_localized_complexity_condition}) with $C_{\mathrm{loc}} = \sqrt{3}$ and complexity parameter $d_\Pi$.

    \noindent
    \textbf{Step 3: Policy-loss decomposition.}
    Define
    \begin{equation*}
        L_\beta(x,z) = \beta D_{\mathrm{KL}}\left(\operatorname{Bern}(\sigma(x/\beta)) \middle\| \operatorname{Bern}(\sigma(z/\beta))\right).
    \end{equation*}
    For $F(u) = \log(1 + e^u)$,
    \begin{equation*}
        \frac{L_\beta(x,z)}{\beta} = F(z/\beta) - F(x/\beta) - F'(x/\beta)\frac{z-x}{\beta}.
    \end{equation*}
    Because $F''(u) \geq v_{B, \beta}$ on $[-B/\beta, B/\beta]$, it follows that
    \begin{equation*}
        L_\beta(x,z) \geq \frac{v_{B,\beta}}{2\beta}(x-z)^2,
        \qquad x, z \in [-B, B].
    \end{equation*}
    The Gibbs-policy identity used in the proof of Theorem~\ref{thm:subopt_regularized} gives
    \begin{equation} \label{eq:joint_lower_policy_decomposition}
        \begin{aligned}
            \mathrm{SubOpt}_\beta(\pi_\vartheta, \pi^\star)
            =& \beta \mathbb{E}_{s \sim \rho}\left[D_{\mathrm{KL}}\bigl(\pi_\vartheta(\cdot \mid s) \| \pi^\star(\cdot \mid s)\bigr)\right]\\
            =& \mathbb{E}_{s \sim \rho}\left[L_\beta(x_{\vartheta}(s), x^\star(s))\right]\\
            =& \sum_{s \in \mathcal{S}} \rho(s) L_\beta(x_\vartheta(s), x^\star(s))\\
            =& w L_\beta(t, 0) + \frac{1}{8d_\Pi} \sum_{i=1}^{d_\Pi}\left[L_\beta(\ell+\theta_i,\ell) + L_\beta(-\ell+\theta_i,-\ell)\right]\\
            \geq& w L_\beta(t, 0) + \frac{v_{B,\beta}}{8\beta d_\Pi} \sum_{i=1}^{d_\Pi}\theta_i^2
        \end{aligned}
    \end{equation}
    For later use, write
    \begin{equation*}
        c_\gamma = \min\left\{B, \log\frac{1-\varepsilon_{\alpha,\gamma}}{\varepsilon_{\alpha,\gamma}}\right\},
        \qquad 
        A_\gamma = L_\beta(c_\gamma, 0).
    \end{equation*}
    In particular,
    \begin{equation} \label{eq:joint_lower_amplitude_bound}
        A_\gamma
        = \frac{c_\gamma}{2} \tanh\left(\frac{c_\gamma}{2\beta}\right) - \beta\log\cosh\left(\frac{c_\gamma}{2\beta}\right)
        \geq \frac{v_{B,\beta}}{2\beta}c_\gamma^2
        \geq \frac{v_{B,\beta}\min\{B^2,4\}}{2\beta} (1 - 2\varepsilon_{\alpha,\gamma})^2.
    \end{equation}
    Here the last inequality uses $\log((1-\varepsilon_{\alpha,\gamma})/\varepsilon_{\alpha,\gamma}) \geq 2(1-2\varepsilon_{\alpha,\gamma})$ and hence $c_\gamma \geq \min\{B,2\}(1-2\varepsilon_{\alpha,\gamma})$.

    \noindent
    \textbf{Step 4: LatentCP-induced contribution.}
    For the remainder of the proof, condition on $\mathcal{H}_{\mathrm{pil}}$.
    The pilot comparison scores are
    \begin{equation*}
        x_{\widehat\pi_{\mathrm{pil}}}(s_c) = B,
        \qquad
        x_{\widehat\pi_{\mathrm{pil}}}(s_{i,\pm}) = \pm\ell,
        \qquad
        x_{\widehat\pi_{\mathrm{pil}}}(s_0) = 0.
    \end{equation*}
    By the BT nonconformity-score definition,
    \begin{equation*}
        \widehat{s}(s,y)
        = \log\left(1 + \exp\left\{-(2y-1)x_{\widehat\pi_{\mathrm{pil}}}(s)\right\}\right).
    \end{equation*}
    Since $0 < \ell < B$, the largest possible score is $H_B = \log(1+e^B)$, attained precisely at $(s_c, 0)$, whereas the smallest is $L_B = \log(1+e^{-B})$, attained precisely at $(s_c, 1)$.
    All scores at the other prompts lie strictly between $L_B$ and $H_B$.

    Let $N_H$ count calibration observations with score $H_B$.
    Then
    \begin{equation*}
        N_H \sim \operatorname{Binomial}(n_{\mathrm{cal}}, w/2),
        \qquad
        k_\gamma = \lceil(n_{\mathrm{cal}}+1)(1-\gamma)\rceil = n_{\mathrm{cal}}+1-j \leq n_{\mathrm{cal}}.
    \end{equation*}
    On $\mathcal{G}_{\mathrm{cal}} = \{N_H \leq j-1\}$, the calibrated threshold satisfies $L_B \leq \widehat{q}_{1-\gamma}^{\mathrm{cal}} < H_B$, so $\widehat{\mathcal C}_\gamma(s_c)=\{1\}$.
    Markov's inequality gives
    \begin{equation*}
        \mathbb{P}(\mathcal{G}_{\mathrm{cal}} \mid \mathcal{H}_{\mathrm{pil}})
        = 1 - \mathbb{P}(N_H \geq j \mid \mathcal{H}_{\mathrm{pil}})
        \geq 1 - \frac{\mathbb E[N_H \mid \mathcal{H}_{\mathrm{pil}}]}{j}
        \geq 1 - \frac{n_{\mathrm{cal}}w}{2j}\geq\frac{3}{4}.
    \end{equation*}
    If the optimization sample contains $s_c$, the empirical objective is uniquely minimized with respect to $t = x_\vartheta(s_c)$ at $t = c_\gamma$, by (\ref{eq:bt_unit_level_pessimistic_objective}).
    Its contribution in (\ref{eq:joint_lower_policy_decomposition}) is therefore $wA_\gamma$.

    We next lower-bound $w$ to quantify the contribution of $s_c$ and control its occurrence in the optimization sample.
    Since $(n_{\mathrm{cal}} + 1) \gamma \geq 1$, the inequality $\lfloor r \rfloor \geq r/2$ for $r \geq 1$ gives
    \begin{equation*}
        \frac{j}{2n_{\mathrm{cal}}}
        \geq \frac{(n_{\mathrm{cal}}+1)\gamma}{4n_{\mathrm{cal}}}
        \geq \frac{\gamma}{4}.
    \end{equation*}
    Together with $1/4 \geq \gamma/4$, this implies $w \geq \gamma/4$.
    Similarly, $j \geq 1$ and $n_{\mathrm{cal}} \geq 1$ imply
    \begin{equation*}
        \frac{j}{2n_{\mathrm{cal}}} \geq \frac{1}{4n_{\mathrm{cal}}},
        \qquad
        \frac{1}{4} \geq \frac{1}{4n_{\mathrm{cal}}},
    \end{equation*}
    so $w \geq 1/(4n_{\mathrm{cal}})$.
    Combining these bounds yields
    \begin{equation} \label{eq:joint_lower_mass_bounds}
        w \geq \frac{\gamma}{4},
        \quad
        w \geq \frac{1}{4n_{\mathrm{cal}}}
        \qquad \Rightarrow \quad
        w \geq \frac{1}{8}\left(\gamma + \frac1{n_{\mathrm{cal}}}\right).
    \end{equation}
    Consequently, the contribution $wA_\gamma$ is at least $A_\gamma(\gamma + 1/n_{\mathrm{cal}})/8$.
    Moreover, the sample-size condition $n_{\mathrm{opt}} \geq 8/\gamma$ gives $n_{\mathrm{opt}}w \geq 2$.
    Since $\gamma < 1/2$, it also gives $n_{\mathrm{opt}} > 16$.

    \noindent
    \textbf{Step 5: Optimization-sample contribution.}
    Let $N_c$ count optimization observations at $s_c$, and let $N_{i,+}, N_{i,-}$ count those at $s_{i,+}, s_{i,-}$.
    Write
    \begin{equation*}
        M_i = N_{i,+} + N_{i,-},
        \qquad 
        \Delta_i = N_{i,+} - N_{i,-}.
    \end{equation*}
    First fix any calibration realization.
    For each paired group, the pilot scores are $\ell$ and $-\ell$, so the two labels' nonconformity scores are exchanged.
    Applying the same calibration threshold therefore yields either $\{1\}$ at $s_{i,+}$ and $\{0\}$ at $s_{i,-}$, or prediction sets that are both empty or both full.
    Since $\ell - h > 0$, the empirical objective for coordinate $\theta_i$ can be written as
    \begin{equation*}
        F_i(\theta)
        = \frac{1}{n_{\mathrm{opt}}}\left[N_{i,+}\psi(\ell+\theta) + N_{i,-}\psi(\ell-\theta)\right],
    \end{equation*}
    where, on the positive score range under consideration,
    \begin{equation*}
        \psi(z) = \begin{cases}
            \log(1+e^{-z}) + \varepsilon_{\alpha,\gamma} z, &\text{for singleton sets},\\
            \log(1+e^z),&\text{for empty or full sets}.
        \end{cases}
    \end{equation*}
    Thus, if $M_i > 0$,
    \begin{equation*}
        F_i'(0) = \frac{\Delta_i \psi'(\ell)}{n_{\mathrm{opt}}},
        \qquad
        0 < F_i''(\theta) \leq \frac{M_i}{4n_{\mathrm{opt}}},
    \end{equation*}
    where $\psi'(\ell) = \varepsilon_{\alpha, \gamma} - \sigma(-\ell)$ in the singleton case and $\psi'(\ell) = \sigma(\ell)$ otherwise.
    By the choice of $\ell$ in the former case and $\sigma(\ell) > 1/2 > a_B$ in the latter, we have $|\psi'(\ell)| \geq a_B$ in both cases.
    
    Fix $i$ with $M_i > 0$.
    If an empirical minimizer satisfies $|\widehat{\theta}_i| < h$, first-order optimality gives $F_i'(\widehat{\theta}_i) = 0$.
    The curvature upper bound therefore implies
    \begin{equation*}
        \frac{a_B|\Delta_i|}{n_{\mathrm{opt}}}
        \leq |F_i'(0)|
        = |F_i'(0)-F_i'(\widehat\theta_i)|
        \leq \frac{M_i}{4n_{\mathrm{opt}}} |\widehat\theta_i|,
    \end{equation*}
    and hence $|\widehat{\theta}_i| \geq 4 a_B |\Delta_i| / M_i$.
    If the minimizer lies on the boundary of $[-h, h]$, then $|\widehat{\theta}_i| = h$.
    Combining the two cases yields, for every empirical minimizer,
    \begin{equation*}
        |\widehat\theta_i|
        \geq \min\left\{h, \frac{4a_B|\Delta_i|}{M_i}\right\}.
    \end{equation*}
    Since $|\Delta_i| \leq M_i$ and $h \geq 2 a_B$, this yields
    \begin{equation} \label{eq:joint_lower_coordinate_displacement}
        \widehat{\theta}_i^2
        \geq 4 a_B^2 \frac{\Delta_i^2}{M_i^2},
        \qquad M_i > 0.
    \end{equation}

    We next construct an optimization event on which these errors accumulate.
    Let $\mathcal{B} = \{1 \leq N_c \leq n_{\mathrm{opt}}/2\}$.
    Because $N_c \sim \operatorname{Binomial}(n_{\mathrm{opt}}, w)$, $n_{\mathrm{opt}}w \geq 2$, and $w \leq 1/4$,
    \begin{equation*}
        \mathbb{P}(N_c = 0) 
        = (1-w)^{n_{\mathrm{opt}}} \leq e^{-2},
        \quad
        \mathbb{P}(N_c > n_{\mathrm{opt}}/2)
        \leq \frac{\mathbb{E}[3^{N_c}]}{3^{n_{\mathrm{opt}}/2}}
        \leq \left(\frac{\sqrt3}{2}\right)^{n_{\mathrm{opt}}}
        \leq \left(\frac{3}{4}\right)^8.
    \end{equation*}
    Hence $\mathbb{P}(\mathcal{B}) > 3/4$.
    Fix an integer $k$ with $1 \leq k \leq n_{\mathrm{opt}}/2$ and condition on $N_c = k$. 
    Then each $M_i$ is binomial with mean
    \begin{equation*}
        \mu_k = \frac{n_{\mathrm{opt}}-k}{4d_\Pi(1-w)}
        \geq \frac{n_{\mathrm{opt}}}{8d_\Pi}
        \geq 1.
    \end{equation*}
    Let $\mathcal{I}_M = \{i: M_i > 0\}$ and $g = |\mathcal{I}_M|$.
    For each $i$, the conditional binomial distribution gives
    \begin{equation*}
        \mathbb{P}(M_i=0 \mid N_c=k)
        = \left(1-\frac{1}{4d_\Pi(1-w)}\right)^{n_{\mathrm{opt}}-k}
        \leq e^{-\mu_k}
        \leq e^{-1}.
    \end{equation*}
    Since $g = \sum_{i=1}^{d_\Pi} \mathbb{I}\{M_i>0\}$, linearity of expectation yields
    \begin{equation*}
        \mathbb{E}[g \mid N_c=k]
        = \sum_{i=1}^{d_\Pi} \mathbb{P}(M_i>0 \mid N_c=k)
        \geq d_\Pi(1-e^{-1})
        \geq\frac{5d_\Pi}{8}.
    \end{equation*}
    On the other hand, using $0 \leq g \leq d_\Pi$ and splitting according to whether $g \geq d_\Pi/4$, we obtain
    \begin{equation*}
        \begin{aligned}
            \mathbb{E}[g \mid N_c=k]
            &\leq \frac{d_\Pi}{4} \mathbb{P}(g<d_\Pi/4 \mid N_c=k) + d_\Pi\mathbb{P}(g\geq d_\Pi/4 \mid N_c=k)\\
            &= \frac{d_\Pi}{4} + \frac{3d_\Pi}{4} \mathbb{P}(g\geq d_\Pi/4 \mid N_c=k).
        \end{aligned}
    \end{equation*}
    Combining the preceding two bounds and rearranging gives
    \begin{equation*}
        \mathbb{P}(g\geq d_\Pi/4 \mid N_c=k) \geq \frac{1}{2}.
    \end{equation*}

    Conditional on $N_c$ and $(M_i)_{i=1}^{d_\Pi}$, the number of observations in each paired group is fixed.
    Since $\rho(s_{i,+}) = \rho(s_{i,-})$, each observation within group $i$ is independently drawn from $s_{i,+}$ or $s_{i,-}$ with equal probability.
    These within-group allocations are also independent across groups. 
    Consequently,
    \begin{equation} \label{eq:joint_lower_counts_n_m}
        N_{i,+} \mid N_c, (M_i)_{i=1}^{d_\Pi} \sim \operatorname{Binomial}(M_i, 1/2),
        \qquad
        \Delta_i = 2N_{i,+} - M_i.
    \end{equation}
    Thus, the $\Delta_i$ are conditionally independent, and each is the sum of $M_i$ independent signs, with $+1$ representing an occurrence of $s_{i,+}$ and $-1$ an occurrence of $s_{i,-}$.

    Fix a realization of $N_c$ and $(M_i)_{i=1}^{d_\Pi}$ for which $g \geq d_\Pi/4$, and define
    \begin{equation*}
        Z = \sum_{i\in\mathcal{I}_M}\frac{\Delta_i^2}{M_i^2}.
    \end{equation*}
    Conditional on these counts, by (\ref{eq:joint_lower_counts_n_m}), we have,
    \begin{equation*}
        \mathbb{E}[\Delta_i \mid N_c, (M_i)] = 0,
        \qquad
        \operatorname{Var}(\Delta_i \mid N_c,(M_i)) = M_i,
    \end{equation*}
    and hence $\mathbb{E}[\Delta_i^2 \mid N_c,(M_i)] = M_i$.
    Expanding the fourth power of the sum of the independent $\pm 1$ contributions gives
    \begin{equation*}
        \mathbb{E}[\Delta_i^4 \mid N_c,(M_i)]
        = M_i + 6\binom{M_i}{2}
        = 3M_i^2 - 2M_i
        \leq 3M_i^2,
    \end{equation*}
    because only terms in which every sign appears an even number of times have nonzero expectation.
    Consequently, the conditional mean of $Z$ is
    \begin{equation*}
        L = \mathbb{E}[Z \mid N_c,(M_i)]
        = \sum_{i\in\mathcal{I}_M}\frac{1}{M_i}.
    \end{equation*}
    Expanding $Z^2$ and using conditional independence across groups, we obtain
    \begin{equation*}
        \begin{aligned}
            \mathbb{E}[Z^2 \mid N_c,(M_i)]
            &= \sum_{i\in\mathcal{I}_M}\frac{\mathbb{E}[\Delta_i^4\mid N_c,(M_i)]}{M_i^4} + 2\sum_{\substack{i<j\\i,j\in\mathcal{I}_M}}\frac{1}{M_iM_j}\\
            &\leq 3\sum_{i\in\mathcal{I}_M}\frac1{M_i^2} + 2\sum_{\substack{i<j\\i,j\in\mathcal{I}_M}} \frac{1}{M_iM_j}\\
            &\leq 3\left(\sum_{i\in\mathcal{I}_M}\frac1{M_i}\right)^2
            =3L^2.
        \end{aligned}
    \end{equation*}
    Here the equality uses $\mathbb{E}[\Delta_i^2\Delta_j^2 \mid N_c,(M_i)] = M_iM_j$ for $i \neq j$, and the first inequality uses the fourth-moment bound above.
    Moreover, Cauchy--Schwarz gives
    \begin{equation*}
        g^2
        = \left(\sum_{i\in\mathcal{I}_M} \sqrt{M_i}\frac{1}{\sqrt{M_i}}\right)^2
        \leq \left(\sum_{i\in\mathcal{I}_M}M_i\right)L.
    \end{equation*}
    Since $\sum_{i\in\mathcal{I}_M} M_i \leq n_{\mathrm{opt}}$ and $g \geq d_\Pi/4$, we conclude that
    \begin{equation*}
        L \geq \frac{g^2}{\sum_{i\in\mathcal{I}_M}M_i}
        \geq \frac{g^2}{n_{\mathrm{opt}}}
        \geq \frac{d_\Pi^2}{16n_{\mathrm{opt}}}.
    \end{equation*}
    Paley--Zygmund inequality therefore yields
    \begin{equation*}
        \mathbb{P}\{Z \geq L/8 \mid N_c,(M_i)\}
        \geq (1-1/8)^2 \frac{L^2}{3L^2}
        = \frac{49}{192}.
    \end{equation*}
    Consequently, there is an event $\mathcal{G}_{\mathrm{opt}} = \mathcal{B} \cap \{g \geq d_\Pi/4\} \cap \{Z \geq L/8\}$, depending only on the optimization contexts, with
    \begin{equation*}
        \mathbb{P}(\mathcal{G}_{\mathrm{opt}})
        \geq \frac{3}{4} \cdot \frac{1}{2} \cdot \frac{49}{192}
        = \frac{49}{512},
    \end{equation*}
    on which $N_c \geq 1$ and $Z \geq d_\Pi^2/(128n_{\mathrm{opt}})$.
    Combining (\ref{eq:joint_lower_coordinate_displacement}) with (\ref{eq:joint_lower_policy_decomposition}), the statistical contribution on this event is at least
    \begin{equation*}
        \frac{v_{B,\beta}}{8\beta d_\Pi}\sum_{i=1}^{d_\Pi}\widehat{\theta}_i^2
        \geq \frac{v_{B,\beta}}{8\beta d_\Pi}\sum_{i\in\mathcal{I}_M}\widehat{\theta}_i^2
        \geq \frac{v_{B,\beta}a_B^2}{2\beta d_\Pi} \sum_{i\in\mathcal{I}_M} \frac{\Delta_i^2}{M_i^2}
        = \frac{v_{B,\beta}a_B^2}{2\beta d_\Pi} Z
        \geq\frac{v_{B,\beta}a_B^2}{256\beta} \frac{d_\Pi}{n_{\mathrm{opt}}}.
    \end{equation*}
    This statement holds for every calibration realization and every exact empirical optimizer.

    \noindent
    \textbf{Step 6: Combining the contributions.}
    Conditional on $\mathcal{H}_{\mathrm{pil}}$, the events $\mathcal{G}_{\mathrm{cal}}$ and $\mathcal{G}_{\mathrm{opt}}$ are independent.
    Their intersection has conditional probability at least
    \begin{equation*}
        \frac{3}{4} \cdot \frac{49}{512}
        = \frac{147}{2048} 
        > \frac{1}{16}.
    \end{equation*}
    On this intersection, every exact empirical optimizer has $\widehat{t} = c_\gamma$, and the preceding bounds give
    \begin{equation*}
        \begin{aligned}
            \mathrm{SubOpt}_\beta(\widehat\pi_{\mathrm{LCP}},\pi^\star)
            &\geq \frac{v_{B,\beta}a_B^2}{256\beta}\frac{d_\Pi}{n_{\mathrm{opt}}} + wA_\gamma\\
            &\geq \frac{v_{B,\beta}a_B^2}{256\beta}\frac{d_\Pi}{n_{\mathrm{opt}}} + \frac{v_{B,\beta}\min\{B^2,4\}}{16\beta} \left(1-2\varepsilon_{\alpha,\gamma}\right)^2 \left(\gamma+\frac1{n_{\mathrm{cal}}}\right),
        \end{aligned}
    \end{equation*}
    where the last inequality uses (\ref{eq:joint_lower_amplitude_bound}).
    Taking the minimum of the two displayed coefficients proves the theorem with the stated $c_{B,\beta}$.
\end{proof}

\subsection{Additional Lower Bound for Conformal Bias} \label{app:persistent_conformal_bias}

A separate one-dimensional BT construction gives a positive lower bound that holds for every realization of the pilot and calibration samples.

\begin{proposition}[Persistent Conformal Bias with Probability One] \label{prop:persistent_conformal_bias}
    Fix $B, \beta > 0$, $\alpha \in (0, 1)$, and $\gamma \in (0, \alpha/2)$.
    There exist a BT instance and a one-dimensional log-linear policy class $\Pi$, independent of the sample sizes, satisfying Assumptions~\ref{ass:subopt_sampling} and~\ref{ass:bounded_comparison_scores}, with $\pi^\star \in \Pi$, $\mathcal{E}_\Pi = 0$ and $C_{\mathrm{comp}}(\pi^\star; \Pi) \leq 1$. 
    The localized-complexity condition (\ref{eq:subopt_localized_complexity_condition}) also holds with $d_\Pi=1$ and $C_{\mathrm{loc}} = 1$. 
    For any positive integers $n_{\mathrm{pil}}, n_{\mathrm{cal}}, n_{\mathrm{opt}}$, with probability one over all samples, every exact LatentCP-DPO empirical optimizer $\widehat{\pi}_{\mathrm{LCP}}$ satisfies
    \begin{equation*}
        \operatorname{SubOpt}_\beta (\widehat{\pi}_{\mathrm{LCP}}, \pi^\star) \geq \frac{v_{B, \beta} B^2}{32 \beta} > 0, 
        \qquad v_{B,\beta} = \sigma(B / \beta) \sigma(-B / \beta).
    \end{equation*}
\end{proposition}
\begin{proof}
    Recall that $\varepsilon_{\alpha, \gamma} = \gamma / \alpha$ and
    \begin{equation*}
        c_\gamma = \min\left\{B, \log\frac{1 - \varepsilon_{\alpha, \gamma}}{\varepsilon_{\alpha, \gamma}}\right\}.
    \end{equation*}
    Since $\varepsilon_{\alpha, \gamma} < 1 / 2$, we have $c_\gamma \in (0, B]$. Choose
    \begin{equation*}
        \vartheta^\star = 
        \begin{cases} 
            3B/4, & c_\gamma \leq B/2,\\ 
            B/4, & c_\gamma > B/2. 
        \end{cases}
    \end{equation*}
    Then $\vartheta^\star \geq B/4$ and $|\vartheta^\star - c_\gamma| \geq B/4$.

    Consider a single prompt $s$ with two responses, $0$ and $1$, and a uniform reference policy.
    Define
    \begin{equation*}
        \Pi = \{\pi_\vartheta: \vartheta \in [-B, B]\}, 
        \qquad \pi_\vartheta(1 \mid s) = \sigma(\vartheta/\beta), 
        \qquad \pi_\vartheta(0 \mid s) = 1 - \sigma(\vartheta/\beta).
    \end{equation*}
    Every preference context equals $D = (s, 0, 1)$, and
    \begin{equation*}
        x_{\pi_\vartheta}(D) 
        = r_{\pi_\vartheta}(s, 1) - r_{\pi_\vartheta}(s, 0) 
        = \beta \log\frac{\pi_\vartheta(1 \mid s)}{\pi_\vartheta(0 \mid s)} 
        = \vartheta.
    \end{equation*}
    Set $r^\star(s, a) = (2a - 1) \vartheta^\star / 2$ for $a \in \{0, 1\}$.
    Then $x^\star(D) = \vartheta^\star$, and the correctly specified BT labels satisfy $y \mid D \sim \operatorname{Bern}(\sigma(\vartheta^\star))$.
    Draw all pilot and calibration preference units i.i.d. from this distribution. 
    Draw the optimization contexts i.i.d. from the context marginal, independently of the labeled samples.

    It follows from the construction that $\pi^\star = \pi_{\vartheta^\star} \in \Pi$, so $\mathcal{E}_\Pi = 0$. The sampling conditions hold by construction, and
    \begin{equation*}
        \sup_{\pi \in \Pi} |x_\pi(D)| \leq B, 
        \qquad |x^\star(D)| \leq B.
    \end{equation*}
    All policies have full response support. The pilot empirical loss is continuous and strictly convex on $[-B, B]$, so its exact minimizer exists and is unique for every pilot sample.

    Fix any realization of the pilot and calibration samples.
    By (\ref{eq:bt_unit_level_pessimistic_objective}),
    \begin{equation*}
        \mathcal{L}_{\alpha, \gamma}^{\mathrm{pess}}(\pi_\vartheta; D)
        = \begin{cases} 
            \log \sigma(\vartheta) - \varepsilon_{\alpha, \gamma}[\vartheta]_+, & \widehat{\mathcal{C}}_\gamma(D) = \{1\},\\ 
            \log\sigma(-\vartheta) - \varepsilon_{\alpha, \gamma}[-\vartheta]_+, & \widehat{\mathcal{C}}_\gamma(D) = \{0\},\\ 
            \log\sigma(-|\vartheta|), & \widehat{\mathcal{C}}_\gamma(D) \in \{\varnothing, \{0, 1\}\}. 
        \end{cases}
    \end{equation*}
    For $\widehat{\mathcal{C}}_\gamma(D) = \{1\}$, the objective is strictly increasing for $\vartheta < 0$, while its derivative for $\vartheta > 0$ is $\sigma(-\vartheta) - \varepsilon_{\alpha, \gamma}$.
    Its unique maximizer over $[-B, B]$ is therefore $c_\gamma$. 
    By symmetry, the unique maximizer for $\widehat{\mathcal{C}}_\gamma(D) = \{0\}$ is $-c_\gamma$. 
    For an empty or full prediction set, the unique maximizer is $0$.

    Since every optimization context equals $D$, the empirical objective equals this unit-level objective for any $n_{\mathrm{opt}} \geq 1$.
    Writing $\widehat{\pi}_{\mathrm{LCP}} = \pi_{\widehat{\vartheta}}$, we have $\widehat{\vartheta} \in \{-c_\gamma, 0, c_\gamma\}$. 
    The choice of $\vartheta^\star$ therefore gives $|\widehat{\vartheta} - \vartheta^\star| \geq B/4$.
    Recall that
    \begin{equation*}
        L_\beta(x, z) 
        = \beta D_{\mathrm{KL}}(\operatorname{Bern}(\sigma(x/\beta)) \| \operatorname{Bern}(\sigma(z/\beta))).
    \end{equation*}
    For $x, z \in [-B, B]$, its log-partition representation is
    \begin{equation*}
        \frac{L_\beta(x, z)}{\beta} 
        = \log(1 + e^{z/\beta}) - \log(1 + e^{x/\beta}) - \sigma(x/\beta)\frac{z - x}{\beta}.
    \end{equation*}
    Let $F(t) = \log(1 + e^t)$.
    Simple calculation gives $F''(t) = \sigma(t)\sigma(-t)$, which is at least $v_{B,\beta}$ on $[-B/\beta,B/\beta]$. 
    Taylor's theorem therefore gives
    \begin{equation*}
        L_\beta(x, z) 
        = \beta \left[F(z/\beta) - F(x/\beta) - F'(x/\beta) \frac{z - x}{\beta}\right] 
        = \frac{F''(t_{x, z})}{2\beta}(x - z)^2 
        \geq \frac{v_{B, \beta}}{2\beta}(x - z)^2,
    \end{equation*}
    where $t_{x,z}$ lies between $x/\beta$ and $z/\beta$.
    The Gibbs identity now gives
    \begin{equation*}
        \operatorname{SubOpt}_\beta (\widehat{\pi}_{\mathrm{LCP}}, \pi^\star) 
        = L_\beta(\widehat{\vartheta}, \vartheta^\star) 
        \geq \frac{v_{B, \beta}}{2\beta} (\widehat{\vartheta} - \vartheta^\star)^2 
        \geq \frac{v_{B, \beta} B^2}{32\beta}.
    \end{equation*}

    For the comparison coefficient, fix $\vartheta \in [-B, B]$ and draw $a \sim \pi^\star(\cdot \mid s)$ and $a' \sim \pi_\vartheta(\cdot \mid s)$ independently.
    The implicit-reward error difference is $0$ when $a = a'$ and equals $\pm(\vartheta - \vartheta^\star)$ otherwise. 
    Consequently,
    \begin{equation*}
        \begin{aligned}
            \mathbb{E}_{a \sim \pi^\star, a' \sim \pi_\vartheta} \left[\left([r_{\pi_\vartheta}(s, a) - r^\star(s, a)] - [r_{\pi_\vartheta}(s, a') - r^\star(s, a')]\right)^2\right] 
            = \mathbb{P}\{a \neq a'\}(\vartheta - \vartheta^\star)^2\\ 
            \leq (\vartheta - \vartheta^\star)^2 
            = \mathbb{E}_D\left[|x_{\pi_\vartheta}(D) - x^\star(D)|^2\right].
        \end{aligned}
    \end{equation*}
    This proves $C_{\mathrm{comp}}(\pi^\star; \Pi) \leq 1$.

    Finally, conditional on any pilot and calibration samples, every $h \in \mathcal{H}_\zeta$ satisfies $0 \leq h(D)\leq 1$ and $\mathbb{E}_D[h(D)] \leq \zeta$ by the definition of $\mathcal{H}_\zeta$.
    Since $D$ is the unique context, $h(D) = \mathbb{E}_D[h(D)]$, so $h(D) \leq \min\{1, \zeta\}$. 
    With $D_i^{\mathrm{opt}} = D$ for every $i$, Cauchy--Schwarz gives
    \begin{equation*}
        \begin{aligned}
            \mathfrak{R}_{n_{\mathrm{opt}}}(\zeta)
            =& \mathbb{E}\left[\sup_{h \in \mathcal{H}_\zeta} h(D) \left|\frac{1}{n_{\mathrm{opt}}} \sum_{i=1}^{n_{\mathrm{opt}}} \epsilon_i \right|\right]
            \leq \min\{1, \zeta\} \mathbb{E} \left|\frac{1}{n_{\mathrm{opt}}} \sum_{i=1}^{n_{\mathrm{opt}}} \epsilon_i\right|\\
            &\leq \min\{1, \zeta\} \left(\frac{1}{n_{\mathrm{opt}}^2} \sum_{i, j = 1}^{n_{\mathrm{opt}}} \mathbb{E}[\epsilon_i \epsilon_j]\right)^{1/2}
            = \frac{\min\{1, \zeta\}}{\sqrt{n_{\mathrm{opt}}}}
            \leq \sqrt{\frac{\zeta}{n_{\mathrm{opt}}}}.
        \end{aligned}
    \end{equation*}
    Thus (\ref{eq:subopt_localized_complexity_condition}) holds with $d_\Pi = 1$ and $C_{\mathrm{loc}} = 1$, completing the proof.
\end{proof}

\section{Additional Experimental Results} \label{app:exp}

\subsection{Synthetic Experiments} \label{app:exp_synthetic}
We use a synthetic PL construction with known true rewards to examine how policy suboptimality varies with $n_{\mathrm{opt}}$, measure the empirical optimization gap $\Delta_{\mathrm{opt}}$, and assess the effect of the conformal compatibility constraint.

\paragraph{Setup.} We use a correctly specified PL model. 
Let the prompts be $s \in \{s_{j,+}, s_{j,-}\}_{j=1}^{d}$, with the prompt distribution
\begin{equation*}
    \rho(s_{j,+}) = \frac{0.51}{d},
    \qquad \rho(s_{j,-}) = \frac{0.49}{d}.
\end{equation*}
Each prompt has $K$ candidate responses $a^1, \ldots, a^K$, and the reference policy is $\pi_{\mathrm{ref}}(a^k \mid s) = 1/K$. We define a linear reward model with a fixed offset
\begin{equation*}
    r_\theta(s_{j,\pm}, a^1) = g_K\pm\theta_j,
    \qquad r_\theta(s_{j,\pm}, a^k) = 0, 
    \quad k = 2, \ldots, K,
\end{equation*}
where $g_K = \log\{19(K - 1)\}$ and $\theta \in [-1, 1]^d$. 
This model can be expressed as the inner product of the feature vector $\mathbb{I}\{k = 1\}(1, \pm e_j)$ and the coefficients $(g_K, \theta)$, where $e_j$ is the $j$-th standard basis vector. 
The corresponding log-linear policy class is
\begin{equation*}
    \Pi = \{\pi_\theta: \theta \in [-1, 1]^d\},
    \qquad \pi_\theta(a \mid s) = \frac{\pi_{\mathrm{ref}}(a \mid s)\exp\{r_\theta(s, a)/\beta\}} {\sum_b \pi_{\mathrm{ref}}(b \mid s)\exp\{r_\theta(s, b)/\beta\}}.
\end{equation*}
Let $\theta^\star = \mathbf{0}$ and the true reward be $r^\star = r_{\theta^\star}$. 
Therefore, the true KL-regularized optimal policy satisfies $\pi^\star = \pi_{\mathbf{0}} \in \Pi$. 
The complete ranking $y_i$ is generated based on the PL model. 
Under the above settings, the true probability that $a^1$ is ranked first is $0.95$ for all values of $K$.

We set $\alpha = 0.3$, $\gamma = 0.1$, and $\beta = 0.5$. For each $(K, d)$, we fit $\widehat{\pi}_{\mathrm{pil}}$ using $n_{\mathrm{pil}} = 3000$ labeled preference units over the same full policy class $\Pi$ used in subsequent policy optimization, with all $d$ free coordinates optimized within $[-1, 1]^d$. 
We then construct $\widehat{\mathcal{C}}_\gamma(D)$ and $\widehat{\mathcal{U}}_\alpha(D; \gamma)$ using $n_{\mathrm{cal}} = 1500$ independent labeled preference units.
For each fixed $(K, d)$, we keep the fitted pilot predictor $\widehat{\pi}_{\mathrm{pil}}$ and calibration data fixed across all values of $n_{\mathrm{opt}}$, and sample the unlabeled optimization contexts independently of both labeled datasets.

We measure optimization error $\Delta_{\mathrm{opt}}$ using the empirical pessimistic objective gap
\begin{equation*}
    \Delta_{\mathrm{opt}}^{(t)} = \widehat{L}(\pi^{(t)}) - \inf_{\pi\in\Pi}\widehat{L}(\pi),
    \qquad \widehat{L}(\pi) = -\frac{1}{n_{\mathrm{opt}}} \sum_{i=1}^{n_{\mathrm{opt}}} \mathcal{L}_{\alpha, \gamma}^{\mathrm{pess}}(\pi; D_i^{\mathrm{opt}}),
\end{equation*}
which measures the difference between the global maximum of the empirical pessimistic objective and its value at $\pi^{(t)}$, with both values computed on the same optimization sample.
The global reference is computed to high precision using the analytic inner solution and the resulting separable convex outer objective; in this optimization experiment, it is used solely to evaluate the gap.

\paragraph{Additional Results.}
Table~\ref{tab:synthetic_ablation_k} shows that LatentCP-DPO reduces mean policy suboptimality by $95.28\%$--$95.77\%$ relative to full-simplex pessimism across the three tested values of $K$.
This complements Figure~\ref{fig:synthetic}(c), which varies $n_{\mathrm{opt}}$ at fixed $K = 4$.

\begin{table}[htb]
    \centering
    \caption{Conformal compatibility ablation across $K$ at fixed $d = 4$ and $n_{\mathrm{opt}} = 5000$.
    Policy suboptimality is reported as the mean $\pm$ the half-width of a pointwise $95\%$ Monte Carlo confidence interval over $5{,}000$ paired repetitions per $K$, conditional on the fixed pilot and calibration data.}
    \label{tab:synthetic_ablation_k}
    \begin{tabular}{cccc}
        \hline
        $K$ & LatentCP-DPO ($\times 10^{-5}$) & Full simplex ($\times 10^{-5}$) & Relative reduction \\
        \hline
        4 & $4.45 \pm 0.09$ & $94.21 \pm 1.18$ & $95.28\%$ \\
        5 & $3.32 \pm 0.07$ & $75.17 \pm 0.85$ & $95.59\%$ \\
        6 & $2.64 \pm 0.05$ & $62.38 \pm 0.73$ & $95.77\%$ \\
        \hline
    \end{tabular}
\end{table}

\subsection{Full Results on Nectar with Qwen3-8B} \label{app:exp_pl_on_nectar}
Table~\ref{tab:levels50} reports each method's average split-conformal prediction-set size at all 50 nominal coverage levels $\tau \in [0.75, 0.95]$.
Figure~\ref{fig:coverage} shows five representative levels for legibility, whereas the reported average set-size reductions use all $50$.
The gains are not driven by a narrow coverage range: LatentCP-DPO yields the smallest set at $49$ of the $50$ levels, outperforms each individual baseline at no fewer than $49$ levels, and never ranks below second.


All evaluated methods share an identical experimental setup: the same base model, corpus, optimizer budget, $1,500$-context calibration split, and $3,000$-context test split. The pilot policy and the calibrated threshold $\widehat{q}$ construct $\widehat{C}$ over all $8,000$ optimization contexts; each final policy is then trained on a $4,000$-context subsample of that pool (i.e., $n_{\mathrm{opt}} = 4,000$). 
Within a seed, this subsample and the LoRA initialization are common to all methods.
We denote CFA as CFA-style$^\dagger$ because the original method constructs conformal sets over freshly generated answers. 
Since our fixed $K = 6$ ranking corpus involves no generation, we adapt CFA by applying its set-wise aggregation directly over the six provided candidates.

\definecolor{cpwin}{HTML}{D62728}
\newcommand{\cpbar}[1]{\textcolor{cpwin}{\rule[-0.15ex]{#1pt}{1.0ex}}}
\begin{table}[htb]
\centering\footnotesize\setlength{\tabcolsep}{4.5pt}
\caption{Full 50-level prediction-set-size results for the fixed Qwen3-8B setting ($K=6$).
Each row corresponds to a requested coverage level $\tau$; entries give the average split-conformal prediction-set size out of $720$ possible rankings, averaged over matched policy seeds, with the smallest entry in each row shown in bold.
LatentCP-DPO yields the smallest set at \textbf{49 of the 50 levels}.
The final column reports its reduction relative to Pure DPO, with a bar proportional to that reduction.}
\label{tab:levels50}
\begin{tabular}{@{}l rrrrrr r@{~}l@{}}
\toprule
 & \multicolumn{6}{c}{Average prediction-set size} & \multicolumn{2}{c}{} \\
\cmidrule(lr){2-7}
$\tau$ & \textbf{Ours} & Pure DPO & CFA$^\dagger$ & ROPO & DPOP & SimPO & \multicolumn{2}{c}{$\Delta$ vs.\ Pure DPO} \\
\midrule
0.7500 & \textbf{256.4} & \textbf{256.4} & 262.0 & 285.8 & 319.1 & 371.2 & $-0.02\%$ & \cpbar{0.00} \\
0.7541 & \textbf{258.9} & 260.1 & 264.7 & 290.3 & 323.8 & 375.8 & $0.47\%$ & \cpbar{2.45} \\
0.7582 & \textbf{261.7} & 265.7 & 269.8 & 295.1 & 330.4 & 379.8 & $1.54\%$ & \cpbar{7.99} \\
0.7622 & \textbf{266.1} & 269.7 & 272.8 & 298.2 & 333.1 & 383.2 & $1.36\%$ & \cpbar{7.06} \\
0.7663 & \textbf{270.3} & 273.5 & 276.6 & 303.6 & 337.6 & 387.2 & $1.18\%$ & \cpbar{6.16} \\
0.7704 & \textbf{274.5} & 277.8 & 280.3 & 307.5 & 341.8 & 390.1 & $1.17\%$ & \cpbar{6.10} \\
0.7745 & \textbf{276.8} & 284.5 & 286.0 & 311.0 & 345.6 & 393.9 & $2.69\%$ & \cpbar{13.99} \\
0.7786 & \textbf{281.4} & 288.5 & 290.4 & 316.5 & 348.9 & 400.5 & $2.47\%$ & \cpbar{12.85} \\
0.7827 & \textbf{285.2} & 292.1 & 295.7 & 321.0 & 353.9 & 405.3 & $2.38\%$ & \cpbar{12.37} \\
0.7867 & \textbf{288.7} & 297.0 & 300.3 & 326.6 & 357.1 & 409.8 & $2.78\%$ & \cpbar{14.43} \\
\addlinespace[2pt]
0.7908 & \textbf{293.1} & 301.8 & 304.7 & 332.4 & 361.0 & 413.1 & $2.89\%$ & \cpbar{15.02} \\
0.7949 & \textbf{296.1} & 307.3 & 309.7 & 337.8 & 367.3 & 417.0 & $3.65\%$ & \cpbar{19.00} \\
0.7990 & \textbf{301.0} & 311.0 & 312.6 & 342.0 & 372.0 & 422.1 & $3.22\%$ & \cpbar{16.75} \\
0.8031 & \textbf{305.4} & 315.6 & 316.4 & 344.7 & 375.8 & 425.8 & $3.22\%$ & \cpbar{16.75} \\
0.8071 & \textbf{310.0} & 318.8 & 321.2 & 348.2 & 380.3 & 430.6 & $2.79\%$ & \cpbar{14.49} \\
0.8112 & \textbf{315.2} & 324.9 & 326.2 & 352.6 & 383.6 & 435.0 & $3.00\%$ & \cpbar{15.61} \\
0.8153 & \textbf{320.6} & 331.5 & 329.9 & 355.3 & 390.0 & 439.2 & $3.30\%$ & \cpbar{17.18} \\
0.8194 & \textbf{323.9} & 335.4 & 333.8 & 360.8 & 396.3 & 444.4 & $3.45\%$ & \cpbar{17.93} \\
0.8235 & \textbf{330.0} & 339.4 & 339.3 & 366.8 & 401.2 & 449.8 & $2.76\%$ & \cpbar{14.34} \\
0.8276 & \textbf{334.0} & 343.8 & 342.5 & 371.6 & 405.5 & 455.8 & $2.86\%$ & \cpbar{14.87} \\
\addlinespace[2pt]
0.8316 & \textbf{339.7} & 347.5 & 345.6 & 375.4 & 411.8 & 459.5 & $2.24\%$ & \cpbar{11.67} \\
0.8357 & \textbf{344.2} & 355.4 & 350.8 & 381.3 & 417.2 & 463.4 & $3.14\%$ & \cpbar{16.32} \\
0.8398 & \textbf{348.4} & 359.4 & 356.4 & 386.1 & 423.1 & 467.3 & $3.05\%$ & \cpbar{15.86} \\
0.8439 & \textbf{352.5} & 366.0 & 363.8 & 389.7 & 428.9 & 470.5 & $3.67\%$ & \cpbar{19.09} \\
0.8480 & \textbf{356.7} & 372.1 & 368.3 & 394.4 & 435.5 & 475.0 & $4.16\%$ & \cpbar{21.63} \\
0.8520 & \textbf{361.5} & 378.4 & 372.6 & 399.9 & 442.9 & 480.7 & $4.46\%$ & \cpbar{23.21} \\
0.8561 & \textbf{368.0} & 383.6 & 380.0 & 411.1 & 448.1 & 487.8 & $4.07\%$ & \cpbar{21.18} \\
0.8602 & \textbf{370.6} & 388.3 & 386.7 & 414.5 & 453.2 & 494.6 & $4.57\%$ & \cpbar{23.76} \\
0.8643 & \textbf{377.5} & 394.6 & 392.9 & 420.4 & 460.8 & 498.4 & $4.35\%$ & \cpbar{22.60} \\
0.8684 & \textbf{383.6} & 399.2 & 398.4 & 424.9 & 467.2 & 503.2 & $3.91\%$ & \cpbar{20.35} \\
\addlinespace[2pt]
0.8724 & \textbf{389.1} & 406.6 & 404.2 & 433.1 & 473.8 & 508.7 & $4.31\%$ & \cpbar{22.39} \\
0.8765 & \textbf{395.2} & 412.5 & 411.1 & 439.1 & 478.5 & 516.6 & $4.19\%$ & \cpbar{21.81} \\
0.8806 & \textbf{402.3} & 418.3 & 417.3 & 448.4 & 483.8 & 524.1 & $3.82\%$ & \cpbar{19.89} \\
0.8847 & \textbf{409.2} & 423.0 & 423.8 & 457.0 & 490.2 & 528.3 & $3.28\%$ & \cpbar{17.06} \\
0.8888 & \textbf{417.7} & 430.2 & 430.0 & 467.7 & 496.7 & 533.9 & $2.89\%$ & \cpbar{15.03} \\
0.8929 & \textbf{425.3} & 433.6 & 439.2 & 476.3 & 503.6 & 539.9 & $1.90\%$ & \cpbar{9.88} \\
0.8969 & \textbf{432.7} & 441.7 & 445.0 & 487.2 & 508.2 & 545.5 & $2.03\%$ & \cpbar{10.58} \\
0.9010 & \textbf{437.1} & 448.4 & 450.4 & 496.6 & 514.2 & 550.9 & $2.53\%$ & \cpbar{13.16} \\
0.9051 & \textbf{443.1} & 455.1 & 454.6 & 504.0 & 520.5 & 558.7 & $2.64\%$ & \cpbar{13.74} \\
0.9092 & \textbf{454.2} & 461.5 & 463.1 & 509.4 & 528.7 & 564.3 & $1.56\%$ & \cpbar{8.13} \\
\addlinespace[2pt]
0.9133 & \textbf{462.0} & 470.1 & 469.5 & 518.7 & 536.8 & 571.1 & $1.73\%$ & \cpbar{9.00} \\
0.9173 & \textbf{469.0} & 478.7 & 477.7 & 523.3 & 545.7 & 579.6 & $2.02\%$ & \cpbar{10.52} \\
0.9214 & \textbf{475.0} & 490.8 & 484.6 & 529.4 & 558.5 & 586.0 & $3.22\%$ & \cpbar{16.75} \\
0.9255 & \textbf{482.9} & 497.3 & 492.9 & 533.7 & 564.1 & 590.9 & $2.91\%$ & \cpbar{15.11} \\
0.9296 & \textbf{491.2} & 503.1 & 502.3 & 539.3 & 569.3 & 598.1 & $2.37\%$ & \cpbar{12.33} \\
0.9337 & \textbf{498.8} & 511.4 & 513.6 & 542.5 & 575.5 & 606.9 & $2.47\%$ & \cpbar{12.85} \\
0.9378 & \textbf{505.5} & 521.6 & 521.6 & 549.5 & 582.2 & 614.4 & $3.08\%$ & \cpbar{15.99} \\
0.9418 & \textbf{516.2} & 529.8 & 528.4 & 556.7 & 590.4 & 623.1 & $2.56\%$ & \cpbar{13.31} \\
0.9459 & \textbf{527.2} & 538.2 & 539.8 & 563.2 & 597.4 & 627.0 & $2.04\%$ & \cpbar{10.62} \\
0.9500 & \textbf{537.6} & 543.7 & 544.0 & 572.4 & 604.1 & 636.4 & $1.11\%$ & \cpbar{5.79} \\
\midrule
\textbf{Mean} & \textbf{372.46} & 383.10 & 383.27 & 414.26 & 446.70 & 488.69 & $\mathbf{2.78\%}$ & \cpbar{14.44} \\
\bottomrule
\end{tabular}
\end{table}
\providecolor{cpwin}{HTML}{D62728}
\newcommand{\cpbarm}[1]{\textcolor{cpwin}{\rule[-0.15ex]{#1pt}{1.0ex}}}
\begin{table}[htb]
\centering\footnotesize\setlength{\tabcolsep}{4.5pt}
\caption{Set-size matrix indexed by realized rather than requested coverage (locked Qwen3-8B endpoint, $K=6$). Robustness companion to Table~\ref{tab:levels50}. Each arm is interpolated onto a common grid of 50 realized coverages spanning $[0.7517, 0.9382]$, the range all six arms attain, and entries are the average prediction-set size out of 720 at that realized coverage. The smallest entry in each row is in bold; LatentCP-DPO is smallest at \textbf{50 of the 50 points}. The last column is our reduction against Pure DPO.}
\label{tab:covmatched50}
\begin{tabular}{@{}l rrrrrr r@{~}l@{}}
\toprule
 & \multicolumn{6}{c}{Average prediction-set size} & \multicolumn{2}{c}{} \\
\cmidrule(lr){2-7}
Realized cov. & \textbf{Ours} & Pure DPO & CFA$^\dagger$ & ROPO & DPOP & SimPO & \multicolumn{2}{c}{$\Delta$ vs.\ Pure DPO} \\
\midrule
0.7517 & \textbf{256.4} & 267.7 & 267.7 & 300.2 & 333.7 & 390.5 & $4.20\%$ & \cpbarm{18.46} \\
0.7555 & \textbf{259.6} & 271.7 & 271.4 & 303.9 & 339.7 & 394.0 & $4.44\%$ & \cpbarm{19.55} \\
0.7593 & \textbf{264.0} & 275.1 & 275.1 & 308.1 & 345.1 & 397.0 & $4.05\%$ & \cpbarm{17.80} \\
0.7631 & \textbf{267.9} & 279.1 & 278.7 & 312.3 & 350.0 & 400.1 & $4.00\%$ & \cpbarm{17.59} \\
0.7669 & \textbf{271.5} & 283.5 & 282.3 & 317.1 & 355.5 & 403.3 & $4.21\%$ & \cpbarm{18.54} \\
0.7707 & \textbf{275.2} & 287.9 & 285.8 & 320.9 & 359.7 & 406.9 & $4.41\%$ & \cpbarm{19.41} \\
0.7745 & \textbf{278.7} & 292.7 & 289.7 & 325.2 & 364.3 & 410.8 & $4.77\%$ & \cpbarm{20.98} \\
0.7783 & \textbf{282.5} & 296.0 & 294.0 & 330.0 & 369.1 & 414.9 & $4.58\%$ & \cpbarm{20.15} \\
0.7821 & \textbf{287.0} & 300.3 & 298.0 & 334.7 & 373.7 & 418.8 & $4.43\%$ & \cpbarm{19.51} \\
0.7859 & \textbf{292.5} & 305.1 & 302.3 & 339.8 & 378.7 & 422.3 & $4.12\%$ & \cpbarm{18.11} \\
\addlinespace[2pt]
0.7897 & \textbf{297.0} & 309.5 & 306.6 & 344.4 & 383.5 & 427.2 & $4.05\%$ & \cpbarm{17.81} \\
0.7935 & \textbf{300.4} & 313.8 & 310.3 & 348.8 & 389.4 & 431.9 & $4.25\%$ & \cpbarm{18.71} \\
0.7974 & \textbf{304.4} & 317.7 & 316.9 & 353.7 & 394.6 & 436.1 & $4.20\%$ & \cpbarm{18.47} \\
0.8012 & \textbf{308.7} & 322.3 & 320.6 & 357.7 & 399.3 & 439.9 & $4.22\%$ & \cpbarm{18.58} \\
0.8050 & \textbf{313.6} & 326.7 & 324.6 & 362.4 & 404.3 & 445.0 & $4.01\%$ & \cpbarm{17.63} \\
0.8088 & \textbf{318.6} & 330.4 & 330.2 & 367.5 & 409.8 & 448.9 & $3.59\%$ & \cpbarm{15.79} \\
0.8126 & \textbf{323.1} & 334.1 & 333.8 & 372.3 & 415.0 & 453.2 & $3.30\%$ & \cpbarm{14.51} \\
0.8164 & \textbf{327.2} & 339.0 & 339.6 & 377.3 & 420.8 & 457.9 & $3.47\%$ & \cpbarm{15.25} \\
0.8202 & \textbf{332.3} & 343.3 & 344.1 & 382.0 & 426.3 & 462.6 & $3.21\%$ & \cpbarm{14.13} \\
0.8240 & \textbf{336.7} & 347.2 & 348.6 & 386.8 & 431.6 & 468.5 & $3.03\%$ & \cpbarm{13.33} \\
\addlinespace[2pt]
0.8278 & \textbf{341.0} & 351.9 & 353.7 & 392.4 & 437.0 & 472.7 & $3.10\%$ & \cpbarm{13.62} \\
0.8316 & \textbf{347.0} & 356.5 & 358.2 & 397.5 & 442.0 & 476.9 & $2.66\%$ & \cpbarm{11.69} \\
0.8354 & \textbf{352.1} & 360.9 & 362.2 & 402.9 & 448.4 & 481.8 & $2.45\%$ & \cpbarm{10.80} \\
0.8392 & \textbf{357.6} & 365.3 & 366.6 & 408.7 & 453.5 & 485.9 & $2.12\%$ & \cpbarm{9.31} \\
0.8430 & \textbf{363.1} & 370.6 & 371.9 & 414.6 & 458.4 & 491.3 & $2.04\%$ & \cpbarm{8.97} \\
0.8468 & \textbf{368.9} & 376.3 & 376.9 & 419.3 & 463.3 & 497.9 & $1.97\%$ & \cpbarm{8.66} \\
0.8507 & \textbf{374.2} & 381.7 & 382.0 & 424.1 & 468.3 & 503.0 & $1.96\%$ & \cpbarm{8.63} \\
0.8545 & \textbf{379.1} & 387.0 & 387.3 & 428.8 & 473.0 & 507.1 & $2.06\%$ & \cpbarm{9.04} \\
0.8583 & \textbf{382.5} & 391.9 & 391.6 & 433.5 & 477.4 & 511.7 & $2.42\%$ & \cpbarm{10.66} \\
0.8621 & \textbf{386.9} & 397.0 & 396.8 & 439.2 & 482.6 & 516.5 & $2.56\%$ & \cpbarm{11.26} \\
\addlinespace[2pt]
0.8659 & \textbf{392.7} & 403.0 & 402.9 & 443.8 & 487.8 & 521.9 & $2.56\%$ & \cpbarm{11.25} \\
0.8697 & \textbf{399.2} & 408.9 & 408.3 & 448.5 & 493.3 & 527.2 & $2.37\%$ & \cpbarm{10.41} \\
0.8735 & \textbf{406.0} & 414.3 & 413.6 & 453.5 & 499.3 & 532.2 & $2.00\%$ & \cpbarm{8.82} \\
0.8773 & \textbf{411.9} & 419.8 & 419.4 & 458.5 & 505.1 & 538.5 & $1.87\%$ & \cpbarm{8.21} \\
0.8811 & \textbf{417.0} & 426.0 & 425.8 & 463.6 & 510.6 & 544.3 & $2.11\%$ & \cpbarm{9.30} \\
0.8849 & \textbf{422.0} & 432.2 & 431.9 & 469.1 & 517.1 & 549.0 & $2.36\%$ & \cpbarm{10.40} \\
0.8887 & \textbf{427.4} & 438.0 & 438.8 & 475.7 & 522.9 & 556.2 & $2.44\%$ & \cpbarm{10.72} \\
0.8925 & \textbf{433.3} & 444.1 & 445.0 & 481.9 & 527.7 & 563.3 & $2.43\%$ & \cpbarm{10.68} \\
0.8963 & \textbf{438.9} & 451.2 & 452.1 & 487.8 & 532.2 & 569.2 & $2.73\%$ & \cpbarm{12.01} \\
0.9001 & \textbf{444.9} & 458.1 & 458.4 & 492.2 & 536.6 & 574.7 & $2.88\%$ & \cpbarm{12.68} \\
\addlinespace[2pt]
0.9040 & \textbf{451.5} & 464.8 & 463.9 & 496.7 & 542.0 & 580.0 & $2.86\%$ & \cpbarm{12.58} \\
0.9078 & \textbf{457.7} & 471.8 & 470.1 & 503.4 & 547.5 & 585.4 & $3.00\%$ & \cpbarm{13.20} \\
0.9116 & \textbf{463.9} & 478.4 & 476.8 & 510.1 & 552.7 & 590.3 & $3.02\%$ & \cpbarm{13.28} \\
0.9154 & \textbf{471.2} & 484.1 & 484.8 & 515.9 & 557.9 & 597.1 & $2.68\%$ & \cpbarm{11.80} \\
0.9192 & \textbf{477.8} & 489.9 & 490.6 & 523.9 & 562.6 & 602.7 & $2.46\%$ & \cpbarm{10.81} \\
0.9230 & \textbf{484.1} & 496.0 & 498.0 & 530.6 & 567.7 & 608.3 & $2.41\%$ & \cpbarm{10.62} \\
0.9268 & \textbf{490.0} & 503.1 & 505.2 & 538.1 & 573.8 & 614.9 & $2.61\%$ & \cpbarm{11.51} \\
0.9306 & \textbf{497.5} & 512.0 & 511.2 & 544.8 & 579.4 & 622.6 & $2.83\%$ & \cpbarm{12.46} \\
0.9344 & \textbf{504.2} & 520.1 & 521.5 & 552.9 & 585.8 & 630.1 & $3.06\%$ & \cpbarm{13.47} \\
0.9382 & \textbf{511.2} & 526.8 & 528.0 & 560.9 & 593.0 & 636.4 & $2.96\%$ & \cpbarm{13.03} \\
\midrule
\textbf{Mean} & \textbf{371.04} & 382.50 & 382.27 & 419.15 & 460.85 & 500.38 & $\mathbf{3.00\%}$ & \cpbarm{13.19} \\
\bottomrule
\end{tabular}
\end{table}

\subsection{Sensitivity to PL Misspecification on Nectar Contexts} \label{app:exp_misspecification}
We conduct a semi-synthetic experiment using real Nectar prompts with $K = 6$ candidate responses per prompt. 
We assign known synthetic rewards to the candidates and use the induced PL distribution as the unperturbed ranking law. 
We consider two mild departures. 
For adjacent swaps, we draw a ranking from PL and, with probability $p$, swap a uniformly chosen pair
of adjacent responses. 
For stage-dependent choice noise, at stage $j = 0, \ldots, 4$, we select a remaining candidate $a$ with probability proportional to $\exp\{u_a^\star/(1+\delta j/4)\}$, where $u_a^\star$ is its assigned reward and $\delta$ is the maximum relative increase in choice temperature.

For each mechanism, we vary its perturbation parameter over $\{0, 0.02, \ldots, 0.10\}$ and run $50$ paired repetitions. 
Each repetition uses disjoint splits of $2{,}000$ pilot, $1{,}500$ calibration, and $4{,}000$ test contexts. 
We reuse the splits and underlying random draws across perturbation levels. 
At each level, we generate rankings for all three splits under the same ranking law and refit the linear PL pilot.
We fix $\alpha = 0.30$ and $\gamma = 0.10$ throughout.
We measure coverage of the simulated test ranking and of the assigned reward vector, respectively.
\begin{equation*}
    \begin{aligned}
        \text{Ranking coverage}
        &= \mathbb{P}\!\left\{
            y_{n+1} \in \widehat{\mathcal{C}}_{\gamma}(D_{n+1})
        \right\},\\
        \text{Known-reward coverage}
        &= \mathbb{P}\!\left\{
            \boldsymbol{u}^\star_{n+1} \in
            \widehat{\mathcal{U}}_{\alpha}(D_{n+1};\gamma)
        \right\}.
    \end{aligned}
\end{equation*}
Ranking coverage measures whether the prediction set contains the simulated test ranking. 
Known-reward coverage measures whether the latent reward uncertainty set obtained by inverting the ranking prediction set contains the assigned reward vector.

\begin{figure}[ht]
    \centering
    \includegraphics[width=0.85\textwidth]{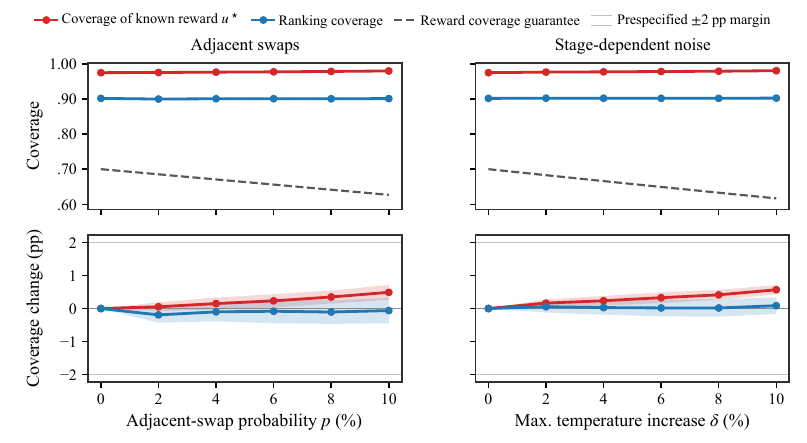}
    \caption{Coverage under mild PL misspecification on Nectar contexts.
    The left and right columns vary the adjacent-swap probability $p$ and the maximum relative temperature increase $\delta$, respectively.
    Blue and red curves show ranking and known-reward coverage in the top row and their paired changes from the common unperturbed PL baseline in the bottom row. 
    Curves are means over $50$ repetitions.
    The gray dashed curves in the top row show the complement of the known-reward miscoverage bound in Theorem~\ref{thm:latent_coverage}, evaluated at the mean total variation distance for each setting.
    In the bottom row, shaded bands show Bonferroni-adjusted $95\%$ confidence intervals, and thin gray lines mark the prespecified $\pm 2$ percentage-point margin.}
    \label{fig:nectar_misspecification_coverage_and_changes}
\end{figure}

As shown in Figure~\ref{fig:nectar_misspecification_coverage_and_changes}, ranking coverage and known-reward coverage under standard PL are $90.17\%$ and $97.50\%$, respectively. 
At an adjacent-swap probability of $10\%$, they are $90.11\%$ and $97.99\%$. 
At a maximum temperature increase of $10\%$, they are $90.26\%$ and $98.07\%$. 
Across the two mechanisms, the changes from the unperturbed baseline lie within the prespecified margin of $\pm 2$ percentage points.
The mean ranking prediction-set size increases from $52.60$ under standard PL to $57.10$ at the adjacent-swap endpoint and $57.50$ at the stage-dependent noise endpoint, out of $720$ possible rankings.

\begin{figure}[ht]
    \centering
    \includegraphics[width=0.8\textwidth]{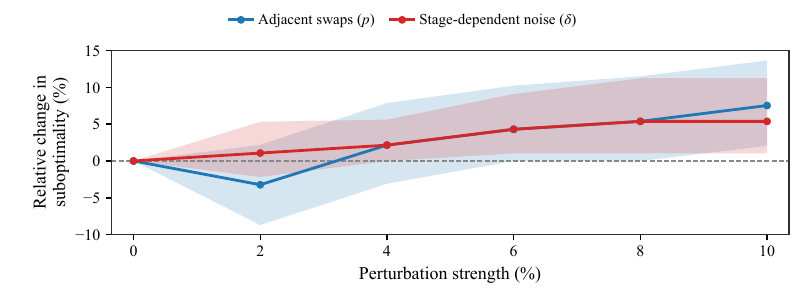}
    \caption{Relative change in mean policy suboptimality under mild PL misspecification on Nectar contexts. 
    At each perturbation level, we express the mean paired difference in $\mathrm{SubOpt}_\beta$ over $100$ repetitions as a percentage of the mean suboptimality under unperturbed PL.
    Shaded bands show pointwise $95\%$ paired bootstrap confidence intervals.}
    \label{fig:nectar_paired_policy_gap}
\end{figure}

We also assess policy sensitivity to the same two perturbations in a separate semi-synthetic experiment. Using the assigned rewards, we evaluate LatentCP-DPO by its policy suboptimality $\mathrm{SubOpt}_\beta(\widehat{\pi}, \pi^\star)$ with $\beta = 1$, where $\pi^\star$ maximizes $J_\beta$ over the finite policy class. 
Mean suboptimality is $0.17921$ under unperturbed PL.
It increases by $0.01349$ ($7.53\%$) at $p = 0.10$ and by $0.00963$ ($5.38\%$) at $\delta = 0.10$, as shown in Figure~\ref{fig:nectar_paired_policy_gap}.

\subsection{Conformal Compatibility Ablation on Semi-Synthetic Nectar Data} \label{app:exp_ablation}
We apply the conformal compatibility ablation from Appendix~\ref{app:exp_synthetic} to a semi-synthetic setting with eight Nectar contexts and six candidate responses per context. 
We assign known rewards, generate rankings under the induced PL model, and optimize over a finite class of $21$ candidate policies. 
In each of $100$ paired repetitions, we draw $1{,}600$ pilot samples, $1{,}000$ calibration samples, and $256$ unlabeled optimization contexts with replacement from the eight contexts. 
We set $\alpha = 0.30$, $\gamma = 0.10$, and $\beta = 1$.
To obtain the full-simplex baseline, we replace the induced ambiguity set in the pessimistic objective (\ref{eq:unit_level_pessimistic_dpo_objective}) with the full simplex over rankings. 
We evaluate both methods by their KL-regularized policy suboptimality under the assigned rewards.

Table~\ref{tab:nectar_calibration_ablation} shows that LatentCP-DPO reduces mean policy suboptimality by $90.79\%$ relative to full-simplex pessimism. 
Pairwise preference accuracy is $81.18\%$ for LatentCP-DPO and $50.00\%$ for full-simplex pessimism. 
The comparison illustrates how restricting worst-case optimization to ranking laws induced by conformally compatible rewards in the finite dictionary can mitigate the conservatism of full-simplex pessimism.

\begin{table}[H]
    \centering
    \caption{Conformal compatibility ablation on semi-synthetic Nectar data ($K = 6$).
    Policy suboptimality is reported as the mean $\pm$ the half-width of a $95\%$ Monte Carlo confidence interval over $100$ paired repetitions.}
    \label{tab:nectar_calibration_ablation}
    \begin{tabular}{@{}ccc@{}}
        \hline
        LatentCP-DPO & Full simplex & Relative reduction \\
        \hline
        $0.17921 \pm 0.00980$ & $1.94621 \pm 0.00000$ & $90.79\%$ \\
        \hline
    \end{tabular}
\end{table}

\subsection{Cross-Architecture Evaluation} \label{app:exp_cross_architecture}
\begin{figure}[ht]
    \centering
    \includegraphics[width=\textwidth]{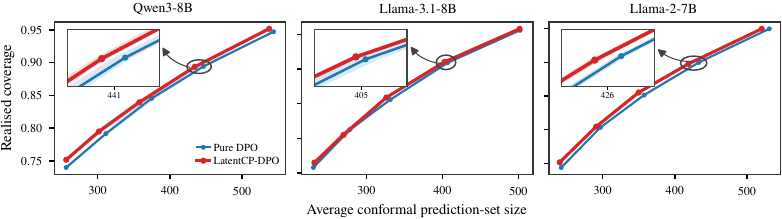}
    \caption{Prediction-set efficiency across base models.
    Realized coverage versus average conformal ranking prediction-set size for LatentCP-DPO and Pure DPO on three base models under the same experimental protocol.
    The Llama runs retain the Qwen3-8B values of $\gamma$.
    Curves show means over three matched policy seeds; shaded bands show their min--max ranges.}
    \label{fig:arch}
\end{figure}

To assess whether the efficiency gains generalize across base models, we evaluate LatentCP-DPO on Llama-3.1-8B-Instruct and Llama-2-7B-Chat.
We use the same data, seed assignments, and optimization budgets as in the Qwen3-8B experiment, including a PL pilot trained on $3,000$ contexts and final policies trained on $4,000$ contexts.
We transfer $\gamma$ unchanged from the Qwen3-8B validation selection,
and no architecture-specific tuning is performed.
As shown in Figure~\ref{fig:arch}, LatentCP-DPO consistently yields smaller conformal prediction sets than Pure DPO at a common requested coverage level across all three models.
The gains thus persist across the tested architectures without architecture-specific hyperparameter search.

\subsection{Experiments with BT Feedback} \label{app:exp_bt}

\paragraph{Synthetic experiment.} We first study how policy suboptimality varies with the number of unlabeled optimization contexts when the fitted pilot and calibration data are held fixed. 
We use a correctly specified BT model with $48$ equally likely contexts and two candidate responses per context. 
Each context has a four-dimensional feature vector $x(D)$, and the true reward difference is $x(D)^\top \theta^\star$, where $\theta^\star = (1.1, -0.7, 0.6, 0.2)^\top$. 
We use a uniform reference policy and set $\alpha = 0.30$, $\gamma = 0.10$, and $\beta = 0.7$.

We fit the BT pilot using $n_{\mathrm{pil}} = 512$ labeled observations and calibrate the prediction sets on an independent sample of size $n_{\mathrm{cal}} = 1{,}024$. 
For policy optimization, we use the one-dimensional affine class $\theta = \theta^\star + t v$. 
This class contains the true optimal policy at $t = 0$.
With the pilot and calibration data fixed, we draw $5{,}000$ independent optimization samples at each $n_{\mathrm{opt}} \in \{2^7, \ldots, 2^{20}\}$, solve the empirical pessimistic objective, and evaluate the resulting policies under the known rewards.

\begin{figure}[ht]
    \centering
    \includegraphics[width=0.8\textwidth]{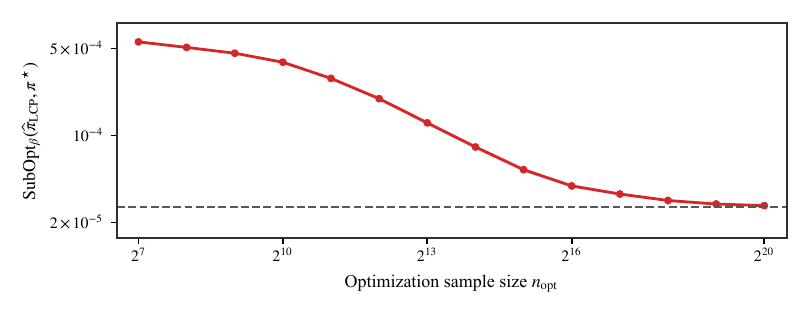}
    \caption{Mean KL-regularized policy suboptimality versus $n_{\mathrm{opt}}$ in the synthetic BT experiment.
    Each point averages $5{,}000$ independent optimization samples, conditional on the same fitted pilot and calibration data.}
    \label{fig:bt_subopt_vs_nopt}
\end{figure}

Figure~\ref{fig:bt_subopt_vs_nopt} shows that mean policy suboptimality decreases from $5.66\times10^{-4}$ at $n_{\mathrm{opt}}=2^7$ to $2.73\times10^{-5}$ at $n_{\mathrm{opt}}=2^{20}$. 
The curve approaches the nonzero suboptimality of the population pessimistic optimizer, $2.66\times10^{-5}$. In this construction, increasing $n_{\mathrm{opt}}$ reduces the effect of optimization-sample variability, while the bias induced by conformal pessimism persists under the fixed conformal uncertainty sets.

\paragraph{Real-data BT evaluation on HelpSteer2.} We evaluate LatentCP-DPO and Pure DPO on human pairwise preferences from \texttt{HelpSteer2} using Qwen2.5-0.5B-Instruct. 
For LatentCP-DPO, we use the fixed configuration $(\alpha, \gamma) = (0.30, 0.04)$.
Both methods use $32$ policy updates. 
We report means over three seeds evaluated on the same $1{,}200$ test pairs. 
For each fitted policy, we calibrate binary preference prediction sets on a separate split of $500$ pairs.

As shown in Table~\ref{tab:helpsteer2_bt_results}, LatentCP-DPO achieves a mean test pairwise accuracy $1.33$ percentage points higher than Pure DPO. 
At the $90\%$ nominal coverage level, realized label coverage is $89.00\%$ for LatentCP-DPO and $88.25\%$ for Pure DPO.

\begin{table}[H]
    \centering
    \small
    \caption{HelpSteer2 test results averaged over three matched seeds.
    Realized label coverage and mean prediction-set size are reported at a common nominal coverage level of $90\%$.}
    \label{tab:helpsteer2_bt_results}
    \begin{tabular}{@{}lccc@{}}
        \hline
        Method & Pairwise accuracy & Realized coverage & Mean set size \\
        \hline
        Pure DPO     & 58.39\% & 88.25\% & 1.6664 \\
        LatentCP-DPO & \textbf{59.72\%} & \textbf{89.00\%} & \textbf{1.6647} \\
        \hline
    \end{tabular}
\end{table}


\subsection{Downstream Response Quality and Overfitting} \label{app:exp_downstream}

\paragraph{Downstream response quality.}
We evaluate responses generated by Qwen3-8B policies trained with LatentCP-DPO and Pure DPO. 
For each of three seeds, we use the final checkpoint of each method to generate responses to the same $128$ Nectar prompts using greedy decoding with a limit of $1{,}024$ new tokens. 
We score the generated responses with ArmoRM \citep{wang2024interpretable}.
As shown in Table~\ref{tab:nectar_response_quality}, LatentCP-DPO receives \emph{slightly higher} scores than Pure DPO on the aggregate score and all three reported attributes. 
The largest difference among these attributes is $0.00549$ for instruction following.

\begin{table}[H]
    \centering
    \small
    \caption{ArmoRM scores for responses generated by Qwen3-8B on $128$  Nectar prompts, averaged over three matched seeds.}
    \label{tab:nectar_response_quality}
    \begin{tabular}{@{}lcc@{}}
        \hline
        Metric & Pure DPO & LatentCP-DPO \\
        \hline
        Aggregate score       & $0.12379$ & \textbf{0.12469} \\
        Helpfulness           & $0.77583$ & \textbf{0.77854} \\
        Correctness           & $0.77675$ & \textbf{0.77931} \\
        Instruction following & $0.81200$ & \textbf{0.81749} \\
        \hline
    \end{tabular}
\end{table}

\paragraph{Late-stage overfitting.} We compare LatentCP-DPO and Pure DPO on HelpSteer2 using Qwen2.5-0.5B-Instruct with BT feedback. 
We independently flip each preference label in the pilot, calibration, optimization, and development splits with probability $0.20$. 
The test labels are left unchanged.
For each of the eight seeds, both methods start from the same fitted pilot and use the same optimization contexts and update schedule.
LatentCP-DPO uses the calibration labels to construct uncertainty sets, while Pure DPO uses the same number of optimization labels. 
We set $\alpha = 0.50$, $\gamma = 0.10$, and $\beta = 0.10$ for LatentCP-DPO and evaluate both methods after $0$, $1$, $2$, $4$, $8$, and $16$ epochs of policy optimization.

\begin{figure}[ht]
    \centering
    \includegraphics[width=\textwidth]{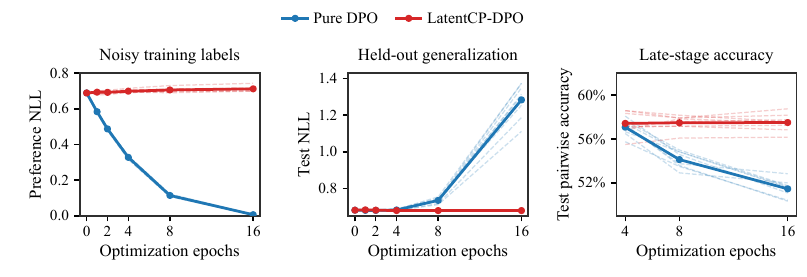}
    \caption{Training trajectories on HelpSteer2 with $20\%$ preference label noise. 
    Panels show NLL on noisy optimization labels, NLL on unflipped test labels, and test pairwise accuracy from epoch $4$ onward.
    Solid lines show means over eight seeds, and faint dashed lines show individual runs.}
    \label{fig:helpsteer2_overfitting}
\end{figure}

Figure~\ref{fig:helpsteer2_overfitting} shows that from epoch $4$ to epoch $8$, Pure DPO's NLL on the noisy optimization labels decreases from $0.32794$ to $0.11432$, while its test NLL increases from $0.68205$ to $0.73442$ and its test accuracy falls from $57.07\%$ to $54.13\%$.
Over the same interval, LatentCP-DPO's test NLL remains nearly unchanged ($0.67898$ to $0.67874$), and its test accuracy stays near $57.4\%$. 
These results indicate that LatentCP-DPO is less susceptible to overfitting under this fixed training schedule.

\subsection{Sensitivity to \texorpdfstring{$\alpha$ and $\gamma$}{alpha and gamma}} \label{app:exp_sensitivity}

\paragraph{BT feedback on HelpSteer2.}
We evaluate LatentCP-DPO with Qwen2.5-0.5B-Instruct, varying one parameter at a time. 
We vary $\alpha \in \{0.10, 0.15, 0.20, 0.25, 0.30\}$ at fixed $\gamma = 0.04$ and $\gamma \in \{0.01, 0.02, 0.04, 0.05, 0.08, 0.10, 0.14\}$ at fixed $\alpha = 0.30$. 
All configurations satisfy $\gamma < \alpha/2$ and use the same three seeds and $32$ policy updates per run.

\begin{figure}[ht]
    \centering
    \includegraphics[width=0.95\textwidth]{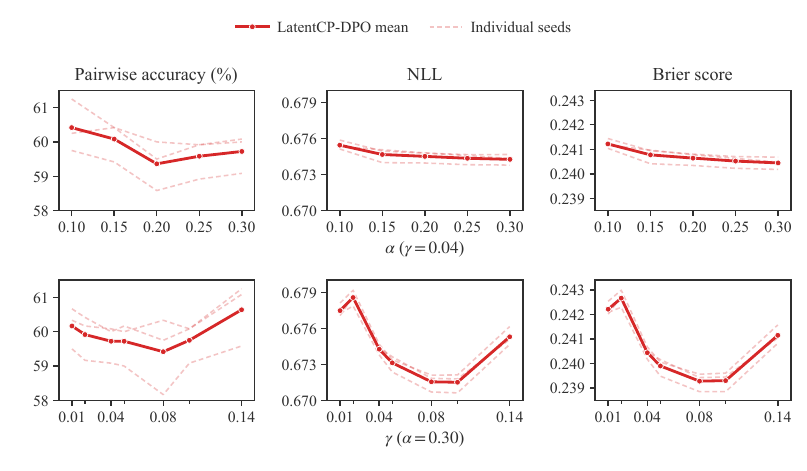}
    \caption{Sensitivity of LatentCP-DPO to $\alpha$ and $\gamma$ under BT feedback on HelpSteer2.
    The top row varies $\alpha$ at fixed $\gamma=0.04$, and the bottom row varies $\gamma$ at fixed $\alpha = 0.30$.
    Columns show test pairwise accuracy, NLL, and Brier score.
    Solid red lines show means over three seeds, and light dashed lines show individual runs.}
    \label{fig:bt_parameter_sensitivity}
\end{figure}

Figure~\ref{fig:bt_parameter_sensitivity} shows that increasing $\alpha$ slightly reduces test NLL and Brier score, with mean pairwise accuracy varying nonmonotonically between $59.36\%$ and $60.42\%$. 
Varying $\gamma$ produces larger, nonmonotonic changes in NLL and Brier score. 
Mean accuracy ranges from $59.42\%$ to $60.64\%$ and is highest at $\gamma = 0.14$ within the $\gamma$ sweep.

\paragraph{PL feedback on Nectar.} We vary $\gamma \in \{0.01, 0.05, 0.10, 0.15, 0.20, 0.25, 0.28\}$ at fixed $\alpha = 0.30$ using Qwen3-8B and $6$ candidate responses ($K = 6$) per prompt. 
We calibrate the ranking prediction sets on $1{,}500$ separate contexts and examine the resulting sets on $8{,}000$ optimization contexts. 
A context retains a pairwise direction if its prediction set is nonempty and at least one candidate pair has the same relative order in every ranking in the set.

\begin{figure}[ht]
    \centering
    \includegraphics[width=0.95\textwidth]{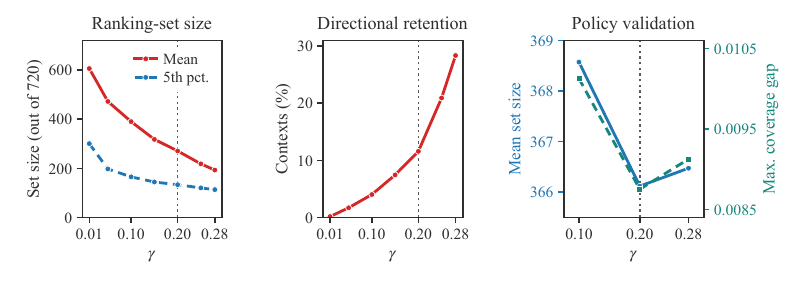}
    \caption{Sensitivity to $\gamma$ under PL feedback on Nectar with Qwen3-8B and $\alpha = 0.30$.
    The left panel shows the mean and fifth percentile of ranking prediction-set sizes on optimization contexts.
    The middle panel shows the percentage of those contexts that retain at least one pairwise direction.
    The right panel shows the mean validation prediction-set size (blue, left axis) and the reported maximum validation coverage gap (teal, right axis) for three values of $\gamma$.
    The vertical dotted line marks the setting selected on validation data, $\gamma = 0.20$.}
    \label{fig:pl_gamma_sensitivity}
\end{figure}

As $\gamma$ increases from $0.01$ to $0.28$, the mean ranking prediction-set size decreases, and the share of contexts retaining a pairwise direction increases from $0.15\%$ to $28.34\%$ (Figure~\ref{fig:pl_gamma_sensitivity}). 
We also evaluate trained policies on validation data at $\gamma \in \{0.10, 0.20, 0.28\}$ using two seeds. 

\end{document}